\documentclass[11pt,a4paper]{article}
\usepackage[top=22mm,bottom=22mm,left=22mm,right=22mm]{geometry}
\usepackage[T1]{fontenc}
\usepackage[utf8]{inputenc}
\usepackage{amsmath,amssymb,amsfonts,bm}
\usepackage{booktabs,longtable,array}
\usepackage{enumitem}
\usepackage{titlesec}
\usepackage{setspace}
\usepackage{graphicx}
\usepackage{url}
\usepackage[hidelinks]{hyperref}
\newcolumntype{L}[1]{>{\raggedright\arraybackslash}p{#1}}
\setlist[itemize]{leftmargin=2em}
\titleformat{\section}{\Large\bfseries}{\thesection}{1em}{}
\titleformat{\subsection}{\large\bfseries}{\thesubsection}{1em}{}
\titleformat{\subsubsection}{\normalsize\bfseries}{\thesubsubsection}{1em}{}

\newcommand{\chiop}{\chi_{\mathrm{op}}}

\newcommand{\Y}{\mathcal{Y}}

\newcommand{\Zphase}{\mathcal{Z}_{\mathrm{phase}}}
\newcommand{\Zref}{\mathcal{Z}_{\mathrm{ref}}}
\newcommand{\Zexo}{\mathcal{Z}_{\mathrm{exo}}}
\newcommand{\Zendo}{\mathcal{Z}_{\mathrm{endo}}}
\newcommand{\Zres}{\mathcal{Z}_{\mathrm{res}}}

\title{Toward AI That Understands Self and Others:\\[4pt]
\large A World-Model Theory of Cognitive Diversity and Alignment}
\author{Toru Takahashi\\[8pt]
Human Informatics and Systems Lab, Doshisha University\\
Linked Open Data Initiative, NPO\\
Keio Research Institute at SFC\\
Stroly Inc.}
\date{Preprint v2, June 2026}

\begin{document}
\maketitle

\begin{abstract}
Modern societies possess more information than ever before, yet they do not converge toward a single shared understanding. The same events, facts, laws, technologies, or risks can be interpreted as evidence of freedom, danger, exclusion, injustice, responsibility, or unrealized possibility. Existing discussions often treat such disagreement as a conflict of values, preferences, or beliefs. This paper argues that disagreement is already a late-stage phenomenon.

The central premise is simple but not trivial: observation is not yet inference. Not every observation becomes inferentially relevant, and not every possible object in an observation sequence becomes an estimation target. A possible target becomes admissible only when a state representation can be constructed--understood here as a compressed internal representation of target-relevant distributions, statistics, or latent variables--that is approximately sufficient for prediction, evaluation, or action with respect to that target. This paper develops a world-model theory of cognitive diversity and alignment by reconstructing recognition as the construction of such approximate sufficient statistics under finite informational, representational, observational, and action constraints.

It formulates this position as the \emph{Multi-Phase Inference Assumption} (MIA) and defines its core internal mechanism as the \emph{Multi-Phase Inference Mechanism} (MIM). The framework then introduces alignment maps and transformation loss to analyze how heterogeneous world models can communicate without being collapsed into a single representation. On this view, cognitive diversity is not a polite name for noise, but a distributed capacity for detecting risks, injuries, constraints, possibilities, and errors that no single inference profile can detect alone.

World-model alignment is therefore processability, not agreement. Rather than seeking a single universally correct representation of the world, the paper proposes a research program for designing AI systems that help heterogeneous forms of intelligence remain mutually processable while preserving their distinct error-detection capacities.
\end{abstract}



\section{Introduction}

\subsection{Why alignment now requires a theory of observation-to-inference}

Modern societies have more information than ever before, yet they do not converge toward a single shared understanding. News, social media, expert reports, policy debates, and everyday experiences of risk, harm, opportunity, and technological change often make people more certain of different dangers, different injustices, different responsibilities, and different futures. This is a familiar but difficult discomfort: if the facts, reasons, or risks seem so clear, why does society not simply move toward the correct conclusion?

This discomfort is no longer only a problem of public discourse. AI systems are entering recommendation, decision support, governance, policy design, and social coordination. They are expected to help classify risks, summarize disputes, support institutional judgment, and mediate human decisions. Yet they enter societies in which what counts as a risk, a value, a loss, a right, a responsibility, or progress is not given in advance. Many theories of intelligence and alignment nevertheless still carry an implicit expectation: given sufficient observations and correct inference, agents should converge toward the same answer.

The alignment problem therefore contains a prior problem. Before AI can be aligned with human values, decisions, or societies, we must ask how anything in an observation sequence becomes a target of inference for humans in the first place. Observation is not yet inference. Not every observation becomes inferentially relevant, and not every possible target becomes available for recognition. A target becomes inferentially admissible only when a state representation can be constructed--a compressed representation of target-relevant distributions, statistics, or latent variables--that preserves enough information for prediction, evaluation, or action. The public event may be shared. The estimation target is not.

This is where the possibility of a different theory of intelligence begins. If cognitive diversity is not noise but distributed error detection, then disagreement is not only a failure of coordination. It is also a clue to errors that no single world model can detect alone. Different people and communities may be sensitive to different distortions in the world: some to structural inconsistency, some to bodily risk, some to existential injury, some to institutional fragility, some to long-term loss, and some to possibilities that existing procedures cannot yet name. The question is not how to erase these differences, but how to make their state representations transformable without losing what each of them can see.

This paper is written from that tension. Modern societies have become better at circulating information and worse, in many cases, at trusting that circulation as understanding. Mass media, institutions, professional systems, social networks, firms, investors, regulators, workers, customers, lawyers, auditors, and citizens all generate signals that are partially visible to one another and partially opaque. The result is not simply pluralism in the comfortable sense. It is a world in which people often experience one another as missing the obvious, while the obvious itself differs across the state representations they inhabit. Treating this as stupidity or bad faith is tempting, and often emotionally efficient, but it stops inquiry exactly where inquiry is most needed.

We may watch the same news reports, encounter the same social-media reactions, and follow the same policy debate; yet we often become more certain of different risks, injuries, responsibilities, and futures. This divergence is not limited to explicit ideological conflict. In one public controversy, one person sees a safety risk, another sees a civil-rights issue, another sees economic opportunity, and another sees institutional failure. Read only as an event, the situation appears to be the same. Inside subjects, however, different worlds may be arising.

The guiding question is therefore prior to the ordinary question of why people disagree: under what condition does anything in an observation sequence become an estimation target at all? An observation sequence does not become recognition as it is. A recognizer must be able to construct, from the observation sequence, a state representation that functions as an approximate sufficient statistic for some target. This is the mathematical hinge of the argument. What becomes available for inference is not whatever has been observed, but whatever can be supported by a target-relevant state representation under finite constraints. Most observations never pass through this chain. They remain unused, unarticulated, embodied, socially suppressed, or inaccessible to conscious and linguistic processing. What counts as the target, which state representation is retained, and which prediction error is treated as a significant occasion for update may differ from person to person, institution to institution, and system to system. These differences are not merely failures of rationality. They are consequences of limited attention and memory, partial observability, action constraints, social constraints, and different histories of world-model update.

Recent AI research has made it natural to describe perception, inference, language, and action in terms of observation sequences, state representations, prediction errors, control, and update. This paper extends that vocabulary to human misunderstanding, value conflict, and disagreement over meaning. It also reframes AI alignment not merely as fitting AI to a human preference model, but as the design problem of making heterogeneous world models mutually processable without forcing them into premature agreement.

\subsection{From SIA to MIA}

Contemporary education, institutional design, social debate, collaborative environments, and AI response design exhibit a recurring implicit expectation: given the same information, anyone reasoning sufficiently rationally should reach roughly the same conclusion. This paper calls this expectation the \emph{Single Intelligence Assumption} (SIA). SIA is not a single proposition explicitly adopted by any specific research field, but a tendency to concentrate intelligence into one privileged mode of reasoning, to normativize it, and to expect convergence from the same input to the same conclusion.

Under finite data, partial observability, representational constraints, action constraints, and social constraints, however, multiple estimation targets, state representations, prediction errors, and update paths can all be consistent with the same observation sequence. This non-identifiability does not imply that humans are merely irrational. Rather, people may form different yet internally coherent inference paths according to their respective world models, available processing resources, histories of representation formation, and error costs.

This paper calls this position the \emph{Multi-Phase Inference Assumption} (MIA). Rather than converging intelligence into a uniform reasoning mode, MIA treats it as an operation in which multiple estimation targets, state representations, and prediction errors may arise in parallel. The paper presents this position as the multi-phase inference framework and calls its core internal mechanism the \emph{Multi-Phase Inference Mechanism} (MIM). In this usage, the framework refers to the overall theoretical proposal of the paper, while MIM denotes the formal device placed at its core. MIM is the formal apparatus that develops MIA in the vocabulary of world models, state representations, foregrounding, and alignment.

A related formulation of inference-profile-based world-model non-identifiability was developed in earlier work~\cite{takahashi2026}. The present paper extends that formulation by shifting the explanatory ground from conclusion divergence itself to the constructive conditions under which estimation targets, state representations, prediction errors, and transformation losses become possible.

\subsection{Question and formal devices of this paper}

The central question of this paper is the following: why do different estimation targets, state representations, prediction errors, and world-model update paths arise from the same observation sequence? To answer this question, the paper begins from the basic structure in which an estimation target $Y$ is endogenously formed from the observation sequence, a state representation $T$ is extracted for that target, and a prediction error $L$ is evaluated.

On this basis, the central devices of this paper can be organized into three. First, the \emph{phase-formation space} $\Zphase$ describes which inferential phases can be formed. Second, the \emph{foregrounding field} $R$ on that space specifies, given an observation sequence and prediction errors, the directions in which processing tends to develop. By ``gradient field'' here we do not mean a gradient field in the strict differential-geometric sense; we mean a direction-field-like structure indicating which directions processing tends to develop within the phase-formation space. Third, the \emph{alignment map} $\Phi$ aligns state representations across different subjects into a processable form. $\Zphase$ carries the field in which diversity is formed; $R$ carries the dynamic operation in that field; and $\Phi$ carries processability across subjects.

Here, $\Zphase$ and $R$ are devices that describe how cognitive differences are generated within a subject, whereas $\Phi$ is the device that makes the resulting state representations processable across subjects. $\Phi$ is therefore not itself the intra-subject generative mechanism, but it is the central device through which this paper connects to social fragmentation and AI alignment.

A subject's operating tendencies are described as a profile state $\Omega$ bundling an operating profile $\theta$, a plasticity/stabilization profile $\lambda$, a firing/awareness-threshold profile $q$, a representation-formation profile $\zeta$, and directional maturity $\sigma$. Here $\theta = (r, e, s)$ is the operating field over the candidate sufficient statistic space that specifies which candidates are foregrounded, which are explored, and which are stabilized. $\lambda$ governs the plasticity and stabilization of this operating field; $q$ controls the thresholds at which errors are routed to reflective processing and the error costs; $\zeta$ controls the formal aspects of estimation-target formation, state-representation formation, compression, and externalization; and $\sigma$ tracks the history of processing precision, fixation, and oversensitivity along each direction. These are auxiliary devices supporting the three pillars. The vocabulary of empirical, ideal, structural, and existential is not part of the axioms of MIM but is introduced from Section~7 onward as auxiliary vocabulary to explain representative coarse-grained regions that appear on $\Zphase$.

\paragraph{Reader's guide: how to enter this paper}
The shortest statement of the paper is this: observation is not yet inference, correct inference excludes errors, and alignment is not agreement. The positive form of the same statement is more generative: cognitive diversity is distributed error detection, and alignment is the design of processability among heterogeneous worlds. The point is not to romanticize every difference, nor to treat all conclusions as equally valid. The point is sharper: what becomes an estimation target is not given in advance. A possible target becomes inferentially available only when a state representation can be constructed that is approximately sufficient for prediction, evaluation, or action with respect to it. Between observation and inference there is therefore a constructive condition: observations must become relevant, relevance must be organized around an admissible target, the target must be supported by a state representation, differences must become prediction errors, errors must be assigned costs, and only then can update, action, disagreement, or alignment begin. This sequence is the common thread connecting the formal theory, the philosophical reconstruction, the account of social fragmentation, and the implications for AI alignment.

This paper is intentionally written as a foundation paper rather than as a narrow technical contribution. Different readers can enter it through different paths. Readers interested in world models, active inference, and representation learning may begin with Sections~2--6, where the paper derives MIM from the construction of approximate sufficient statistics. Readers interested primarily in AI alignment may focus on Sections~3, 4, 9, 11, and 12, which develop state representations, transformation loss, processability, and multi-phase design. Readers from philosophy and social theory may begin with Sections~1, 6, 7, and 10, where the theory is connected to recognition, error, institutions, tradition, and social fragmentation. Readers interested in cognitive typology may treat Section~8 as a stress test for the framework rather than as an axiom of the theory. Readers concerned with empirical and technical development may focus on Sections~13 and 14.

Because the paper introduces many symbols, it is useful to distinguish the core trajectory from the auxiliary machinery. On a first reading, it is enough to follow six symbols: $Y,T,L,\Zphase,R,\Phi$. $Y$ is the estimation target, $T$ the state representation, $L$ the prediction error, $\Zphase$ the phase-formation space, $R$ the foregrounding field, and $\Phi$ the alignment map for transforming state representations across subjects. The remaining symbols refine this trajectory into an implementable research program.

\begin{center}
\footnotesize
\renewcommand{\arraystretch}{1.15}
\begin{tabular}{p{0.20\textwidth}p{0.72\textwidth}}
\toprule
Reading path & Sections to prioritize \\
\midrule
World-model and active-inference readers & Sections~2--6, then Sections~9 and 13. These sections reconstruct recognition through state representations, prediction errors, and world-model update. \\
AI alignment and AI governance readers & Sections~3, 4, 9, 11, and 12. These sections define transformation loss, processability, AI as mediator rather than sovereign optimizer, and the multi-phasing of design itself. \\
Philosophy, social theory, and political thought readers & Sections~1, 6, 7, 10, and 12. These sections redescribe recognition, disagreement, liberalism, tradition, identity, and institutions as problems of distributed error detection. \\
Cognitive diversity and typology readers & Section~8, read together with Sections~3--6. Typology is used as a coarse-grained stress test for the theory, not as its foundation. \\
Technical development readers & Sections~13 and 14, after the core definitions in Sections~3--5. These sections outline empirical tractability and future implementation paths. \\
\bottomrule
\end{tabular}
\end{center}

MIM is not a static classification. It is a dynamic loop in which prediction errors arise from observation sequences, those errors foreground estimation targets and state representations, plan candidates such as thought, speech, action, suspension, and coordination are generated, and their results return to the next observation and world-model update. The central message can be stated simply: world-model alignment is processability, not agreement; cognitive diversity is distributed error detection; and future AI alignment requires making the design process itself multi-phase. The point is not to celebrate every difference as equally correct, but to prevent any single inference regime from becoming so complete that the errors it cannot see disappear from collective attention.

\subsection{Contributions and organization}

The contributions of this paper can be organized into five points. First, the paper clarifies SIA as a composite assumption of centralization, normativization, and exchangeability, and shows its limits. Second, it formulates MIA as a multi-phase inference assumption grounded in the non-identifiability of world-model estimation. Third, it introduces the candidate sufficient statistic space, the phase-formation space, the foregrounding gradient field, the operating profile, and the profile state, and gives a constructive description of the cognitive and inferential mechanism of humans. Fourth, it introduces the alignment map $\Phi$ and presents a design principle by which AI does not identify different world models with one another but rather bridges them into a mutually processable form. Fifth, through the concepts of distributed error detection, transformation loss, and the multi-phase character of design itself, it redefines AI alignment as the problem of extending the endogenous error-detection, error-selection, and error-adjustment capacities of human society.

In this paper, therefore, ``understanding'' does not mean sharing an identical representation. It means that a state representation formed in one subject is processed without breakdown within another subject's world model, and is reconfigured into a form that can enter, as needed, into exploratory updates. In this sense, an AI that understands self and others is not a device that eliminates differences but a device that renders differences processable, makes transformation loss explicit, and makes plural error-selection candidates comparable.

The paper consists of fifteen sections. Section~1 (the present section) introduces the opposition between SIA and MIA and presents the central devices of the paper. Section~2 organizes existing models and the gap this paper fills. Section~3 shows that an observation sequence is not by itself recognition, and that a state representation $T$ constructed as an approximate sufficient statistic for an estimation target $Y$ is required under finite constraints. Section~4 formalizes the candidate sufficient statistic space and the phase-formation space $\Zphase$ from these constraints. Section~5 treats the operating profile $\theta = (r, e, s)$, the foregrounding gradient field $R$, the profile state $\Omega$, and the generative loop on the candidate sufficient statistic space. Section~6 provides an interim summary of the epistemological turn induced by the formalization developed up to that point. Section~7 reads the structure of error in the history of philosophy to the extent necessary for an MIA-based theory of AI alignment. Section~8 positions cognitive typology as a coarse-grained explanatory map. Section~9 formally defines the alignment map $\Phi$ and world-model alignment. Section~10 redescribes society, culture, and civilization as distributed error detection. Section~11 redefines AI alignment, Section~12 develops the principle of multi-phasing design itself, Section~13 presents empirical tractability and technical research, Section~14 the research program, and Section~15 the conclusion.

Depending on the reader's interest, readers concerned with computer science and AI research may focus on Sections~2--6 together with Sections~9, 11, and 13. Readers concerned with philosophy and social theory may first read Sections~1, 6, 7, 10, and 12 to grasp the theoretical scope of the paper. Readers interested in cognitive typology may read Section~8 as an independent bridging section.

\section{Existing Models and the Gap This Paper Fills}

This section is a map, not a detour. The same difficulty has been named in several languages: world models, active inference, prediction error, alignment, recognition, disagreement, and social fragmentation. The point is to show why these vocabularies need a common layer of state representations and transformation loss.

Many existing models have addressed the process by which state representations are formed from observations and prediction errors are reduced. However, they have often not adequately addressed what is foregrounded as the estimation target, why distinct state representations arise as legitimate error-processing paths for distinct subjects, and how those differences can be made mutually processable across subjects. These are the three points this paper supplements.

In this section, to clarify what the Multi-Phase Inference Mechanism newly supplements, we organize existing research in two stages. First, we examine how world-model research, the free-energy principle, theory of mind, AI alignment, multi-agent systems, and cognitive typology relate to this paper. Second, we sketch the formal skeleton common to these and extract the gap that existing models have not adequately handled.

\subsection{Relation to existing research areas}

The related research within which our framework should be situated spans six areas: world-model research, the free-energy principle and active inference, theory of mind (ToM) research, AI alignment research, multi-agent system research, and cognitive typology. This section is not an exhaustive review but a map indicating where the formal devices of this paper become necessary. The insights accumulated by world-model research, the free-energy principle, theory of mind, AI alignment, multi-agent systems, and cognitive typology, each from a different angle, are rearranged from the standpoint of differences in state representation across subjects. The three central devices to be introduced in later sections---the phase-formation space $\Zphase$ (Section~4), the foregrounding field $R$ (Section~5.3), and the alignment map $\Phi$ (Section~9)---do not replace the achievements of these six areas but serve as devices for connecting the problems they have addressed to a comparable formal vocabulary. The three-layer alignment hierarchy is formally introduced in Section~9, and its social and technical implications are developed in Sections~10--12.

\subsubsection{World-model research and its relation to this paper}

World-model research has presented architectures in which an intelligent subject extracts latent states from an observation sequence, learns their temporal evolution as an internal representation, and uses it for action policy \cite{ha2018,lecun2022}. For example, the world model of Ha and Schmidhuber consists of an encoder compressing observations into a latent vector space, an RNN that learns the temporal evolution of latent states, and a controller whose action policy operates on the latent state as input. LeCun's Joint Embedding Predictive Architecture presents a design that secures predictability while avoiding generative detail by learning predictive relations among observations directly in a latent space.

The contribution of these studies is to show that a world model can be a general computational architecture that does not depend on a specific sensory modality or a specific embodiment. Our framework inherits this generalization and describes the world model as the internal model $W^\alpha$ of subject $\alpha$. The additional contribution of this paper is to give a formal description of inter-subject differences in estimation-target selection and state-representation extraction within the world model. While the Ha--Schmidhuber and LeCun frameworks address ``how a given world model is learned,'' this paper addresses ``different subjects learn different world models; how can that difference be described?'' The two are complementary.

\subsubsection{The free-energy principle and active inference}

The free-energy principle (FEP) and active inference describe intelligence as a unified process of prediction-error minimization, perceptual inference, and active inference \cite{friston2010,friston2017,parr2022}. While referring to the basic perspective of FEP---that the subject updates its world model and generates action through prediction-error minimization---this paper does not extend FEP but shifts the focus to describing inter-subject differences of state representation. In particular, concepts such as prior preference, precision-weighted prediction error, active inference, and epistemic value connect to the directional-fit mechanism $\mu$ of this paper \cite{schmidhuber1991,schwartenbeck2015}.

There are three points at which this paper connects to the vocabulary of FEP. First, instead of describing the process as a unique minimization of variational free energy, it is described as a non-unique process unfolding under a subject-specific prior operating profile $\theta=(r,e,s)$. Second, the generation of the estimation target $Y_K$ depends not only on the observation sequence but also on the encodability available within the world model (the estimation-target space $\Y^{\alpha}$); this dependence is made explicit. Third, social fragmentation and AI alignment are described within the scope of FEP as inter-world-model operations mediated by $\Phi$.

This paper is not a replacement or extension of FEP but provides an independent formal apparatus for describing ``subject-specific differing operation'' within FEP's vocabulary.

\subsubsection{Theory of mind (ToM) and other-agent modeling}

ToM research treats the capacity of one subject to infer the mental states (beliefs, desires, attention, goals) of another \cite{rabinowitz2018}. For example, the Machine Theory of Mind presents a neural-network architecture that estimates an agent model of others from observed behavior, implementing observation-based other-agent modeling in a computable form. The Thinking Through Other Minds framework \cite{veissiere2020} extends active inference to cultural and social transmission processes and describes world-model updating through the generative models of others as variational inference.

The reference-target preference toward the endo-phenomenal space $\Zendo$ (the degree to which one includes self/other internal-world variables in the state representation) and the alignment map $\Phi$ introduced in this paper connect directly with these studies. In an operation that strongly foregrounds $\Zendo$, the state representation itself includes self/other internal-world variables, and $\Phi$ aligns that state representation into another subject's world model. This can be read as a redescription of what Machine Theory of Mind tacitly performs as other-agent modeling, recast as a more general framework of state-representation alignment. The contribution of this paper is to position ToM as a special case in which endo-phenomenal reference is strongly foregrounded, and to show that the same form of alignment operation applies to operations differing in exo-phenomenal content or resolution.

\subsubsection{AI alignment research (RLHF / DPO / Constitutional AI)}

Current AI alignment research has developed as a one-way optimization problem in which human preferences and safety criteria are reflected in AI outputs. RLHF \cite{ouyang2022} learns a reward model from pairwise comparisons by human evaluators and adjusts a language-model policy via reinforcement learning. DPO \cite{rafailov2023} optimizes the policy directly from preference data without explicitly learning a reward model. Constitutional AI \cite{bai2022} introduces a loop of self-evaluation and revision using constitutional principles set by humans.

In our framework, these studies are repositioned as state-representation alignment under specific conditions. That is, they are an instance of one-directional, phase-specific state-representation alignment in which the sender is human, the receiver is an AI, and the aligned content is a human preference signal (mainly a mixture of existential and structural regions). The contribution of this paper is to hierarchize these specific implementations as part of a single formal device $\Phi$, and to show that the same form of operation can be applied not only between AI and humans, but also among humans, between AIs, and even among different inference phases within a single self.

Under the three-layer hierarchy of this paper (value, state representation, world model), current RLHF/DPO/CAI is located at the value-alignment layer. By placing the state-representation alignment layer above it and the world-model alignment layer above that, this paper reorganizes current research as a delimited sub-problem within a broader hierarchical structure.

\subsubsection{Multi-agent system research and distributed representation}

Multi-agent system research, especially coordination among large language models and distributed representation learning, has been developing rapidly in recent years. Distributed representation learning via the Transformer \cite{vaswani2017}, stepwise reasoning via Chain-of-Thought \cite{wei2022}, action coupling via ReAct / tool use \cite{yao2023}, and bodily action generation via Vision-Language-Action models \cite{brohan2023} can all be read as attempts to extend state-representation processing and action coupling across different phases.

Our framework offers two contributions to these implementations. First, by classifying which coarse-grained region the state representations foregrounded by these implementations belong to, it provides a tool for describing the capability profiles of AI systems as continuous quantities. Second, by implementing $\Phi$ as a communication protocol among multi-agent AIs, it presents a path through which AI agents with mutually different internal representations may coordinate via state-representation alignment.

\subsubsection{Connection with cognitive typology}

Cognitive typology---Jungian and post-Jungian psychological typology \cite{jung1921}, MBTI \cite{myers1985}, Kepinski's information-metabolism theory \cite{kepinski1972}, and Socionics \cite{augustinaviciute1980,pietrak2018}---is a research program that has empirically described which kinds of information human subjects foreground and along which directions they tend to update their world models. Because these theories lacked a modern computer-science vocabulary, they have long been treated as personality classifications, and their connection with computational theory has been severed.

Our framework redescribes these typological observations as differences in estimation-target region, directional bias, directional maturity, awareness threshold, and world-model alignment capacity. In particular, the four coarse-grained regions can be read as a structural counterpart of Jung's cognitive functions (sensation, intuition, thinking, feeling), and the two directional biases recast the opposition between extraversion and introversion as the distinction between the expansion direction and the compression direction of the state-representation distribution within the same phase. Details are developed in Section~8.

\subsection{Position of this paper}

On the basis of the above, the distinct contributions of this paper can be organized as follows.

First, this paper connects, through a common formal apparatus---estimation-target space, state representation, coarse-grained region, and state-representation alignment---the research areas of world-model research, FEP, theory of mind (ToM), AI alignment, multi-agent systems, and cognitive typology, which have hitherto developed independently. These areas all address processing inside the world model, but because the phenomena each targets are different, mutual reference has been sparse. Our framework describes them in a single formal language.

Second, this paper expands AI alignment from a one-directional problem of ``aligning AI to humans'' into a bidirectional, multi-agent, and multi-phase problem of constructing convertibility between different world models. The three-layer alignment hierarchy (value, state representation, world model) is the hierarchical structure used to organize this expansion.

Third, this paper provides devices for reconstructing the empirical observations of cognitive typology as hypotheses verifiable within computer science. As a result, the typological knowledge that has been isolated as personality classification acquires the possibility of dialogue with AI research, FEP research, and theory-of-mind (ToM) research.

The six areas above have each developed their own vocabularies, but the common structure underlying them has not yet been named. What this paper sketches from the next section onward is the Multi-Phase Inference Mechanism as that common structure.

\subsection{Formal sketches of the major learning and inference models}

\paragraph{Strategy of the formal sketches}
This section places side by side, in minimal formal form, the major computer-scientific learning and inference models referenced by this paper---world-model theory, the free-energy principle (FEP) and active inference, predictive coding, the information bottleneck, reinforcement learning, and theory of mind (ToM). It is not a substitute for specialized expositions of each model; its aim is to provide the reader with a common formal vocabulary for positioning our framework.

In the previous section, we organized the relation to related research areas at the level of prose. In this section we move that to a formal level. Each model taken up here shares the skeleton from observation to internal representation, prediction, error, and update, while each has focused on a particular facet. In the next section, with these formal sketches as background, we introduce the central device of this paper, the \emph{Multi-Phase Inference Mechanism}, and make explicit its commonalities with and differences from existing models.

\subsubsection{Formal sketch of world-model theory}

In the standard form of world-model research, a subject with observation space $\mathcal{O}$, latent space $\mathcal{Z}$, and action space $\mathcal{A}$ is constructed as follows. Here we have in mind Ha--Schmidhuber \cite{ha2018}, Dreamer \cite{hafner2020,hafner2021}, and the JEPA family \cite{lecun2022}.
\begin{equation}
z_t = \mathrm{enc}(o_t), \quad z_{t+1} \sim p(z_{t+1} \mid z_t, a_t), \quad \hat{o}_t = \mathrm{dec}(z_t).
\end{equation}
The loss can be written as the sum of reconstruction, transition, and reward prediction:
\begin{equation}
\mathcal{L}_{\mathrm{WM}} = \mathbb{E}\!\left[\, \|o_t - \hat{o}_t\|^2 + D_{\mathrm{KL}}\!\left[q(z_t) \,\big\|\, p(z_t \mid z_{t-1}, a_{t-1})\right] + \|r_t - \hat{r}_t\|^2 \,\right].
\end{equation}

The design choice here is to commit to a single latent space $\mathcal{Z}$, a single encoder $\mathrm{enc}$, and a single inference path. The same world model supports prediction and action for multiple purposes. The three-layer structure of compressing observations into latent representations, learning the temporal evolution of latent states, and selecting actions through a policy constitutes the basic skeleton of world-model research.

\subsubsection{Formal sketch of the free-energy principle and active inference}

The free-energy principle (FEP) describes perception and learning uniformly as the minimization of variational free energy \cite{friston2010,parr2022}. For a generative model $p(o, s)$ and a recognition distribution $q(s)$, the variational free energy is defined as
\begin{equation}
F_t \;=\; \mathbb{E}_{q(s)}\!\left[\ln q(s) - \ln p(s, o_t)\right] \;=\; D_{\mathrm{KL}}\!\left[q(s) \,\big\|\, p(s \mid o_t)\right] - \ln p(o_t).
\end{equation}
Perceptual inference is written as the minimization of $F_t$:
\begin{equation}
q_t \;=\; \arg\min_{q}\; F_t[q].
\end{equation}

Active inference extends this to action selection. Defining the expected free energy $G(\pi)$ for policy $\pi$,
\begin{equation}
G(\pi) \;=\; \sum_{\tau > t}\, \mathbb{E}_{q(s_\tau, o_\tau \mid \pi)}\!\left[\ln q(s_\tau \mid \pi) - \ln p(s_\tau, o_\tau)\right],
\end{equation}
action selection becomes $a_t^{\ast} = \arg\min_{a} G(\pi \ni a)$ \cite{friston2017,schwartenbeck2015}. The observation sequence is not given passively; rather, $a_t$ actively selects the next observation $o_{t+1} \sim p(o \mid s_{t+1}, a_t)$.

FEP writes perception, learning, and action uniformly with a single free-energy function $F$ and a single recognition distribution $q(s)$. Active inference plays a central role as the reference basis of this paper precisely because it incorporates action while preserving this unity.

\subsubsection{Formal sketch of predictive coding}

Predictive coding \cite{rao1999,clark2013} describes the cortical hierarchy as a loop of prediction--error--update. For a hierarchical level $\ell = 1, 2, \ldots, L$, each level holds a state $s_\ell$ and is updated by predictions from the higher level and error signals from the lower level:
\begin{align}
\hat{x}_\ell &= g_\ell(s_\ell), & \text{(prediction from higher to lower level)}\\
e_\ell &= x_\ell - \hat{x}_\ell, & \text{(lower-level error)}\\
s_\ell &\leftarrow s_\ell + \alpha \cdot W_\ell^{\top} e_\ell. & \text{(higher-level update)}
\end{align}
Here $\alpha$ is the learning rate and $W_\ell$ is the linear mapping for prediction. Lower-level errors drive higher levels and higher-level predictions suppress lower levels; a pair of top-down predictions and bottom-up errors runs through the hierarchy.

Implicit in this scheme is the assumption that a single local structure (prediction--error--update) is stacked hierarchically. From low-level sensory layers near perception up to high-level abstract representations, all are processed by an isomorphic local rule. The framework discussed in later sections of this paper, while taking this local structure of ``prediction--error--update'' into account, introduces a parallelism along a different axis orthogonal to the hierarchical dimension.

\subsubsection{Formal sketch of the information bottleneck}

The information bottleneck (IB) theory \cite{tishby1999} formalizes, in terms of mutual information, the compression--prediction trade-off for input $X$, target variable $Y$, and representation $T$:
\begin{equation}
\min_{p(T \mid X)}\; I(X; T) - \beta \cdot I(T; Y).
\end{equation}
$T$ is optimized as a sufficient statistic for predicting $Y$. $I(X; T)$ represents the complexity of the representation, $I(T; Y)$ represents predictive accuracy, and $\beta$ is the trade-off coefficient.

An important implication is that the representation $T$ is determined relative to the prediction target $Y$. There is no universally ``good'' representation; what one wants to predict selects the representation. This position is one of the information-theoretic foundations for the claim, introduced in later sections, that ``$T$ differs by subject and by inference phase.''

\subsubsection{Formal sketch of reinforcement learning}

Reinforcement learning (RL) \cite{sutton2018} learns a policy that maximizes cumulative reward over a Markov decision process with states $s \in \mathcal{S}$, actions $a \in \mathcal{A}$, and rewards $r$. Via the value function $V^{\pi}(s) = \mathbb{E}_{\pi}[\sum_{t} \gamma^t r_t \mid s_0=s]$ and the action-value function $Q^{\pi}$, the optimal policy is obtained as the solution of the Bellman equation. Model-based RL also learns a world model $p(s' \mid s, a)$ \cite{hafner2020,hafner2021}; under a reinterpretation of reward as a negative log-prior, $V^{\pi}$ and the expected free energy $G$ of active inference become formally connected.

\subsubsection{Formal sketch of theory of mind}

Theory of mind (ToM) treats the capacity of one subject to infer the mental states---beliefs, desires, attention, goals---of another \cite{premack1978,rabinowitz2018}. The form most consistent with the argument of this paper is close to a nested generative model or to interactive POMDPs \cite{gmytrasiewicz2005}. For subject $\alpha$ to predict the behavior of subject $\beta$, $\alpha$ must infer the internal state $\mathbf{m}^{\beta}$ of $\beta$ within $\alpha$'s own generative model, which can be written as $p_\alpha(a_\beta \mid o_\alpha) = \int p_\alpha(a_\beta \mid \mathbf{m}^{\beta}) \, p_\alpha(\mathbf{m}^{\beta} \mid o_\alpha) \, d\mathbf{m}^{\beta}$. The structure at the core is the recursively nested operation in which the subject applies its own inferential machinery to the other. The alignment map $\Phi$ introduced later in this paper can be positioned as a generalization of the operational side of this other-agent inference, recast as a state-representation translation between self and other.

\subsection{Common skeleton and limitation of existing models}

Surveying the formal sketches above, these models share the skeleton of observation $\rightarrow$ internal representation $\rightarrow$ prediction $\rightarrow$ error $\rightarrow$ update while each focuses on a particular aspect. World-model theory uses a single latent space and a single encoder; FEP and active inference use a single free-energy function and a single recognition distribution; predictive coding uses a vertical-hierarchical prediction--error loop; the information bottleneck uses a single sufficient statistic relative to a single prediction target; reinforcement learning uses a single policy and a single value function; theory of mind handles a self-other binary nested inference.

There is one blind spot common to all of these. Namely, ``intra-subject horizontal parallelism''---the possibility that within the same subject, multiple inference paths run simultaneously and each foregrounds a different prediction target---is not structurally addressed by any of these models.

This blind spot appears in three forms. First, \emph{singularity}. Many models assume ``one internal representation'' and ``one inference path'' per subject. It is difficult for them to accommodate the possibility that the same subject elevates multiple different prediction targets in parallel from the same observation. Second, \emph{vertical hierarchy}. Predictive coding has a hierarchical structure, but this is a vertical hierarchy of ``different abstraction levels of the same world,'' not a horizontal structure that ``elevates different prediction targets in parallel.'' Third, \emph{non-separation of time scales}. Many models treat per-step updates and long-term structure formation under the same objective function. They do not explicitly separate the time scale of per-step prediction-error minimization from that of long-term processes in which learning changes the structure of tendencies themselves.

Singularity hides horizontal parallelism along the same dimension; vertical hierarchy hides parallelism along a different direction; non-separation of time scales hides how parallelism becomes fixed on a long time scale. These three forms are different appearances of the same blind spot. In the next section, we introduce the Multi-Phase Inference Mechanism, the central device of this paper, as an apparatus that opens this blind spot head-on.

\section{Conditions for the Emergence of State Representations under Constraints}

\paragraph{Methodological note on ``subjects.''}
From this section onward, this paper uses the term ``subject'' as a methodological abstraction. This abstraction should not be read as an ontological identification of humans and AI systems. Humans have bodies, histories, emotions, relationships, cultures, religious commitments, and existential meanings in ways that current AI systems do not. The claim is narrower: insofar as humans, AI systems, or institutional decision procedures construct internal states from observation sequences and use them for prediction, evaluation, or action, they face a shared class of information-theoretic constraints. They must compress observations, preserve some information, tolerate some distortions, and decide which discrepancies count as errors. In this restricted sense, they can be compared as different implementations of constrained state-representation formation. The purpose of this comparison is not to reduce humans to machines, but to ask how human meaning, value, embodiment, culture, social loss, and existential concern can become processable by AI systems without being flattened into a single representation.

\paragraph{Position of this section.}
A constrained recognizer cannot keep everything it observes. Recognition begins with selection and loss: most of the observation sequence must be discarded, compressed, ignored, deferred, or made irrelevant before anything can become inferable at all. This is not a defect of cognition. It is the condition under which cognition becomes possible under finite attention, memory, representation, observation, and action.

From this section through Section~6, this paper derives its theoretical core in a constructive manner. The introductory discussion approached human limits in terms of attention, memory, experience, and action. From here on, these limits are abstracted as representational/computational, observational, and action constraints. The starting point is the constraint that an observation sequence is not, by itself, recognition. A subject, in the methodological sense just stated, cannot retain, compare, predict from, or act upon an observation sequence in raw form. Therefore, for recognition to be established, the observation sequence must be compressed into an internal state that can be used for prediction, error evaluation, and the generation of action. This paper calls such an internal state a \emph{state representation} $T$, understood here as a compressed representation of target-relevant distributions, statistics, or latent variables.

In what follows, this paper writes the observation sequence received by subject $\alpha$ up to time $t$ as
\begin{equation}
O_t^\alpha=o_{1:t}^\alpha .
\end{equation}
The observation sequence $O_t^\alpha$ is not the world itself; it is an input given to the subject under partial, temporal, embodied, and institutional constraints. Observation is always incomplete $(C_{\mathrm{obs}})$; the subject's computational resources, attention, memory, and representational capacity are finite $(C_{\mathrm{comp}})$; and the subject must select among acts such as thinking, suspending judgment, speaking, coordinating, and bodily action within finite time $(C_{\mathrm{act}})$. Under these three constraints, the observation sequence cannot be treated as recognition in its raw form.

\paragraph{Outline of the derivation.}
First, by $C_{\mathrm{comp}}$, the subject cannot retain, compare, or use the entire observation sequence $O_t^\alpha$ for action generation at a uniform resolution. Second, by $C_{\mathrm{obs}}$, the same observation sequence is compatible with multiple latent structures and does not uniquely determine the state of the world. Third, by $C_{\mathrm{act}}$, the subject cannot continue exploration indefinitely and must, at some point, choose among action, utterance, suspension, or reinterpretation. Therefore the subject must construct, from the observation sequence, an internal state $T$ usable for prediction, error evaluation, and action generation. The state representation $T$ is a necessary condition for recognition to be established under these three constraints.

\subsection{Observation sequences are not, by themselves, recognition}

The observation sequence $O_t^\alpha$ is high-dimensional, extends in time, contains noise, and includes information that the subject cannot directly compare or retain. Moreover, the same observation sequence is compatible with multiple latent structures. Therefore, under incomplete observation, the state or meaning of the world is not uniquely determined by the observation sequence. For a recognizing subject to be said to have recognized something, the observation sequence must be transformed into a state representation usable within the subject's world model.

What matters here is that the state representation $T$ is not a mere record. A state representation is an internal representation that the subject uses for future prediction, error evaluation, explanation, action selection, and sharing with others. It is therefore not simply a contraction of the observation sequence: it must preserve predictive information about some latent variable or latent relation.

For a candidate latent variable $Y$, to say that the state representation $T$ is usable means that at least the following approximation holds:
\begin{equation}
 p^\alpha(Y\mid O_t^\alpha) \approx p^\alpha(Y\mid T^\alpha).
\end{equation}
Ideally,
\begin{equation}
 p^\alpha(Y\mid O_t^\alpha)=p^\alpha(Y\mid T^\alpha) .
\end{equation}
In this case $T^\alpha$ is a sufficient statistic of the observation sequence $O_t^\alpha$ with respect to $Y$. For real recognizing subjects, full sufficiency need not be assumed. Below, we call $T^\alpha$ an \emph{approximate sufficient statistic} with respect to $Y$ when
\begin{equation}
D_{\mathrm{KL}}\!\left(p^\alpha(Y\mid O_t^\alpha)\,\Vert\,p^\alpha(Y\mid T^\alpha)\right)\leq \varepsilon .
\end{equation}

In what follows, unless otherwise noted, this paper uses the term ``sufficient statistic'' in a broad sense that includes not only strict sufficient statistics in classical statistics but also approximate, task-relative \emph{state representations in the broad sense} that preserve information about the estimation target $Y$ in a form that a finite subject can use for prediction, error evaluation, and action generation.

\subsection{Estimation targets are constrained by sufficient statistics}

The estimation target $Y$ in this paper is not an object that exists, as given, within the observation sequence. An estimation target is a latent variable or latent relation in terms of which prediction error can be defined, when the subject uses the observation sequence for compression, prediction, error evaluation, and action generation. More fundamentally, an estimation target is established within the subject's world model only when an approximate sufficient statistic can be constructed for it.

\paragraph{Constructibility condition.}
The world model $W_t^\alpha$ of subject $\alpha$ does not generate arbitrary latent variables from the observation sequence without restriction. At time $t$, the subject has a class $\mathcal{H}_t^\alpha$ of candidate latent variables and candidate latent relations, and a class $\mathcal{T}_t^\alpha$ of internal states realizable under its computational resources, memory, attention, body, and institutional scaffolds. For some $Y\in\mathcal{H}_t^\alpha$ to be established as an estimation target, there must exist an internal state $T\in\mathcal{T}_t^\alpha$ that preserves sufficient predictive information about $Y$ without retaining the entire observation sequence $O_t^\alpha$. Accordingly, this paper says that $Y$ belongs to the set $\mathcal{Y}_t^\alpha$ of estimation targets for subject $\alpha$ at time $t$ if and only if the following condition holds:

\begin{equation}
Y\in\mathcal{Y}_t^\alpha
\quad\Longleftrightarrow\quad
Y\in\mathcal{H}_t^\alpha\;\;\text{and}\;\;
\exists T\in\mathcal{T}_t^\alpha\;\;\text{s.t.}\;\;
D_{\mathrm{KL}}\!\left(p^\alpha(Y\mid O_t^\alpha)\,\Vert\,p^\alpha(Y\mid T)\right)\leq \varepsilon .
\end{equation}

This condition expresses the constraint that, for a finite subject, a latent variable can become a stable target of prediction, error evaluation, and action selection only if an approximate sufficient statistic for it is available. Therefore the subject cannot treat an arbitrary latent variable as an object of recognition. Any $Y$ for which an approximate sufficient statistic $T$ cannot be constructed will not become a stable target of prediction, error evaluation, or action selection for the subject.

This condition has weak implications for consciousness as well. The present paper does not aim to provide a theory of consciousness as such. Nevertheless, for a subject to use some content stably for reporting, attention allocation, reflection, and action generation, that content must at least be formed as some internal state $T$. Therefore information that cannot be constructed as an approximate sufficient statistic is unlikely to become a target of recognition in the sense of this paper, nor a stable object of reflective consciousness. The approximate sufficient statistic referred to here does not denote only fully explicit and stable representations. Unease, presentiments, bodily reactions, and undifferentiated biases of attention are also treated as immature state representations, insofar as they can be repeatedly used in subsequent prediction-error processing and action selection.

\subsection{Recognition as an information bottleneck}

A mind that kept everything would not thereby understand everything. It would be unable to decide what kind of difference matters. Recognition therefore requires a bargain: preserve what is relevant for a target, and let the rest become loss. A state representation $T$ does not faithfully replicate the entire observation sequence $O$. What a finite subject requires is to compress the observation sequence while preserving information about the estimation target $Y$. This requirement can be expressed in the form of information bottleneck theory~\cite{tishby1999}:
\begin{equation}
\mathcal{L}_{\mathrm{IB}}
=
I(O;T)-\beta I(T;Y).
\end{equation}

The first term $I(O;T)$ expresses how much information the state representation retains from the observation sequence---that is, the representational cost. The second term $I(T;Y)$ expresses how much information the state representation retains about the estimation target. The coefficient $\beta$ controls the trade-off between compression and predictive information. The information bottleneck thus expresses the constraint that a recognizing subject does not preserve arbitrary information: it retains information useful for the estimation target and discards the rest.

What matters here is that the information bottleneck is not, in this paper, a mere metaphor borrowed from an external theory. If a finite subject cannot retain the full observation sequence and yet requires a state representation usable for estimation, prediction, error evaluation, and action generation, then the subject necessarily faces a trade-off between compression of input information and preservation of information about the estimation target. This trade-off is precisely the condition for the formation of the state representation $T$.

\paragraph{Scope of the derivation used here.}
The claim made here is not the strong neurocomputational hypothesis that the human brain explicitly optimizes the information-bottleneck objective. The claim is weaker and more foundational. Any subject that must, by means of a finite-capacity internal state $T$, preserve predictive information about an estimation target $Y$ from an incomplete observation $O$ faces, at least formally, a trade-off between $I(O;T)$ and $I(T;Y)$. The information bottleneck is therefore used in this paper not as an implementation algorithm but as a constraint condition on the possibility of recognition in finite subjects.

\subsection{Rate-distortion and the form of recognition}

Compression is never innocent. To compress is to decide which distortions are tolerable and which would destroy the point of the representation. The approximate sufficient statistic $T$ is not unique. Even for the same estimation target $Y$, the subject must choose which differences to preserve and which to discard. The relevant quantity here is the distortion function $d$. The distortion function specifies how heavily each kind of error is weighted. In rate--distortion theory, the information required under an allowed average distortion $D$ is expressed as~\cite{cover2006}
\begin{equation}
R(D)
=
\min_{p(t\mid o):\,\mathbb{E}[d(Y,\hat{Y}(T))]\leq D}
I(O;T).
\end{equation}

This expression shows that the form of recognition is not determined by information quantity alone, but by which distortions are tolerated. Whether the subject weights short-term error more heavily or long-term error more heavily; whether it weights local differences more heavily or wide-area structure more heavily; whether it weights concrete, embodied differences more heavily or abstract, institutional relations more heavily; whether it tolerates loss through coarse-graining or computational cost through fine-graining---all of these can be understood as differences in the distortion function and representational capacity of the state representation $T$.

This paper collectively calls the granularity, temporal range, spatial range, level of abstraction, and operability of a state representation its \emph{resolution} $\rho$. Resolution is not mere fineness: it is the form of a state representation determined by which distortions are preserved and which are discarded. Accordingly, the resolution-foregrounding tendency $r_{\mathrm{res}}$ introduced in later sections is understood as an operating tendency that expresses which rate--distortion trade-off the subject is likely to adopt.

At this point, coarse-graining and fine-graining cease to be merely explanatory words. Coarse-graining is the operation of gathering wide-range variation at low representational cost while discarding local differences as distortion. Fine-graining is the operation of preserving local differences while increasing representational and exploratory cost. The same rate--distortion trade-off appears, depending on the structure of the target, as several apparent oppositions.

\begin{center}
\small
\begin{tabular}{p{0.25\linewidth}p{0.62\linewidth}}
\toprule
Manifestation & Meaning in rate--distortion \\
\midrule
Coarse vs.\ fine & Lower representational cost while gathering variation, vs.\ paying higher representational cost to preserve local differences. \\
Abstract vs.\ concrete & In hierarchical representation, discarding differences between individual instances to preserve relational structure, vs.\ preserving the sensory and operational differences of individual instances. \\
Long vs.\ short range & In the temporal distortion function, preserving distant consequences, vs.\ preserving immediate errors at high resolution. \\
Global vs.\ local & Preserving wide-area spatial, institutional, or relational structure, vs.\ preserving concrete variation in the immediate neighborhood. \\
\bottomrule
\end{tabular}
\end{center}

Abstraction, concretization, short-range and long-range processing, and localization and globalization are therefore not psychological axes added independently. They are local manifestations of the resolution $\rho$ as it acts on temporal, spatial, and hierarchical structures. However, operability and ideality are not determined by $\rho$ alone; they are determined by its coupling with the action constraint $C_{\mathrm{act}}$, the exploration field $e$, and the stabilization field $s$. The coarse-graining domains and typological vocabulary used in later sections are introduced as human-readable low-dimensional maps of this rate--distortion structure.

\subsection{Standard form of finite-dimensional sufficient statistics}
\label{sec:finite-sufficient-statistics}

The discussion so far has shown that the state representation $T$ must be approximately sufficient with respect to the estimation target $Y$. The term ``sufficient statistic,'' however, is not a mere metaphor. In statistical theory, a sufficient statistic is a compressed representation that preserves the information necessary for some estimation target or parameter without retaining the entire observation $O$. In regular statistical models, the Pitman--Koopman--Darmois theorem---which states that families of distributions admitting finite-dimensional sufficient statistics independent of sample size are precisely the exponential families---illustrates this point in canonical form~\cite{pitman1936,darmois1935,koopman1936}.

This paper does not claim that all human recognition is strictly described by exponential families. Rather, exponential families and information geometry are used as the \emph{standard form} and reference model for treating finite-dimensional sufficient statistics~\cite{wainwright2008,amari2016}. In an exponential family, a distribution can be written in terms of a natural parameter $\eta$ and a sufficient statistic $T(O)$ as
\begin{equation}
 p_{\eta}(O)=h(O)\exp\!\left(\eta^{\top}T(O)-A(\eta)\right) ,
\end{equation}
where $A(\eta)$ is the log-partition function and its gradient $\nabla A(\eta)$ yields the expectation of the sufficient statistic. In this structure, information about the observation is compressed into the space of a finite set of estimation-relevant statistics $T_i(O)$ and the natural parameters $\eta_i$ that weight them.

The ``form'' of a state representation is therefore not chosen arbitrarily. As long as a finite subject operates with finite-dimensional sufficient statistics, the state representation possesses a geometry: which statistics to retain, and along which natural-parameter directions to evaluate error. In the vocabulary of information geometry, the natural and expectation parameters form dual coordinates, and the KL divergence can be treated as a distance or distortion on this statistical manifold. From this point of view, the resolution, conditioning basis, and foregrounding field introduced in later sections are understood not as psychological categories but as operating structures concerning the directions in which finite-dimensional sufficient statistics are formed, compared, and updated.

\subsection{Multi-phase inference as a constructive consequence}
\label{sec:constructive-consequence-of-mia}

From the discussion so far, multi-phase inference appears not as a mere empirical fact that people hold different values but as a constructive consequence expected from the recognition conditions of finite subjects. A finite subject cannot retain the full information of the observation sequence and must form an approximate sufficient statistic with respect to an estimation target. Moreover, that sufficient statistic is constrained by the trade-off between compression and preservation of relevant information through the information bottleneck, by the allocation of distortion through rate--distortion, and by the coordinate structure imposed by the standard form of finite-dimensional sufficient statistics.

Under these conditions, when subjects differ in available capacity, learning history, embodied and institutional scaffolds, allowed distortions, errors that are easily triggered, and possibilities of action, the space of constructible approximate sufficient statistics also differs across subjects. Hence, that the same observation sequence gives rise to different estimation targets, different state representations, and different prediction errors is naturally to be expected from the fact that finite subjects have different conditions for sufficient-statistic formation, without assuming any additional psychological typology. The Multi-Phase Inference Assumption (MIA) of this paper is the name that makes this constructive consequence explicit as a basic assumption governing the updating of world models.

\section{From Sufficient Statistics to the Phase-Formation Space}

Once recognition is understood as selective compression, the question changes. We no longer ask only what the subject observed. We ask what kinds of targets and representations could have been formed at all. This section introduces the spaces in which that question can be stated.

\subsection{Co-Constitution of Estimation Targets and State Representations}

In classical statistical theory, the estimation target and its parameters are given in advance, and sufficient statistics are defined relative to them. In human cognition, however, what counts as an estimation target is itself contingent on the observation sequence and on the subject's capacity for representation formation. We therefore begin by defining, for a subject $\alpha$, the set of $(Y,T)$ pairs that can be constructed from the observation sequence as admissible estimation-target--state-representation pairs.

\begin{equation}
\mathcal{A}_t^\alpha
=
\left\{(Y_i^\alpha,T_i^\alpha)\mid
D_{\mathrm{KL}}\!\left(p^\alpha(Y_i^\alpha\mid O_t^\alpha)\,\Vert\,p^\alpha(Y_i^\alpha\mid T_i^\alpha)\right)\leq \varepsilon
\right\}_{i}.
\end{equation}

Here $\mathcal{A}_t^\alpha$ is the space of admissible pairs that subject $\alpha$ can construct at time $t$. An estimation target $Y_i$ is not something discovered as an entity within the observation sequence. Rather, $Y_i$ comes to exist as an estimation target precisely when an approximate sufficient statistic $T_i$ for it can be constructed. Hence $Y$ and $T$ are not determined unidirectionally; they are co-constituted under the finite subject's constraints of compression, prediction, and distortion minimization.

The space $\mathcal{A}_t^\alpha$, however, is an abstract space that records only the conditions under which $Y$ and $T$ may stand. In the next subsection we decompose how $T$ is constructed, through what conditioning basis and at what resolution, and we introduce the candidate sufficient statistic space $\mathcal{C}_t^\alpha$ that incorporates this decomposition.

\subsection[Conditioning Basis psi and Resolution rho]{Conditioning Basis $\psi$ and Resolution $\rho$}

Even for a fixed estimation target $Y$, the question of which information the subject retains in its state representation does not have a unique answer. We therefore decompose the degrees of freedom of state-representation formation into two components. First, the subject constructs how the observation sequence is articulated into latent structures, feature sets, or relational structures that serve as a basis for conditioning. We call this constructive variable the conditioning basis $\psi$. Second, the subject selects the grain, time scale, spatial scale, abstraction level, and operationality at which this basis is compressed. We call this representational form the resolution $\rho$.

It is essential that $\psi$ is not an ``object'' that exists ready-made in the observation sequence. Rather, $\psi$ is a constructive variable that expresses how, in the course of forming a state representation $T$ for $Y$, the subject articulates the observation sequence into a conditioning basis. Likewise, $\rho$ is not mere coarseness of display: it is a representational form determined by which distortions are preserved and which are tolerated.

The state representation can then be written as
\begin{equation}
T_t^\alpha
=
\tau_\rho^\alpha(O_t^\alpha;Y^\alpha,\psi^\alpha).
\end{equation}
This expression does not mean that $T$ is mechanically extracted from the observation sequence. Rather, it expresses that the subject, with respect to an estimation target $Y$, constructs an approximate sufficient statistic $T$ via a conditioning basis $\psi$ and a resolution $\rho$.

\subsection{The Phase-Formation Space}

Taken together, the candidates for sufficient-statistic formation can be represented as four-tuples consisting of the estimation target $Y$, the conditioning basis $\psi$, the resolution $\rho$, and the state representation $T$:
\begin{equation}
\mathcal{C}_t^\alpha
=
\left\{(Y,\psi,\rho,T)\mid
T=\tau_\rho^\alpha(O_t^\alpha;Y,\psi),\;
D_{\mathrm{KL}}(p^\alpha(Y\mid O_t^\alpha)\Vert p^\alpha(Y\mid T))\leq \varepsilon
\right\}.
\end{equation}

Here $\mathcal{C}_t^\alpha$ is the candidate sufficient statistic space that subject $\alpha$ can construct at time $t$. An element $c=(Y,\psi,\rho,T)$ of this space simultaneously specifies what stands as an estimation target, through which conditioning basis and at which resolution, and which state representation is thereby constructed as an approximate sufficient statistic.

The phase-formation space ${\Zphase}_t^\alpha$ can then be defined as the space obtained from the candidate space by forgetting the state representation itself and retaining only the constructive degrees of freedom for the estimation target, the conditioning basis, and the resolution. In its minimal form,
\begin{equation}
{\Zphase}_t^\alpha
=
\left\{(Y,\psi,\rho)\mid
\exists T:\;(Y,\psi,\rho,T)\in\mathcal{C}_t^\alpha
\right\}.
\end{equation}

Here $Y$ is the latent variable that defines error, $\psi$ is the conditioning basis, and $\rho$ is the resolution of the state representation. This triple determines which state representations are constructible for the subject. The phase-formation space $\Zphase$ is therefore not merely a psychological classification space; it is a space that expresses the constructive degrees of freedom of the candidate space within which a finite subject can form approximate sufficient statistics.

For expository purposes, later sections will sometimes distinguish a reference component from a resolution component, writing the space of conditioning bases as $\Zref$ and the space of resolutions as $\Zres(\psi)$. These, however, are not target regions given ready-made in the observation sequence; they are articulations of the conditioning basis and resolution that the subject constructs in the course of sufficient-statistic formation.

\subsection{Coarse-Graining Domains}

The phase-formation space $\Zphase$ is in general high-dimensional. Each individual triple $(Y,\psi,\rho)$ represents a very fine-grained candidate for sufficient-statistic construction. Human cognition and social vocabulary, however, do not operate directly on this high-dimensional space. Regions that are formed repeatedly, transfer easily among one another, and tend to share similar distortion functions and state representations coalesce into stable coarse-graining domains.

We call such stable regions coarse-graining domains $K$. The vocabulary of empirical, ideational, structural, and existential domains, used from Section~7 onward, provides a human-readable coarse-graining map; it does not exhaust the phase-formation space itself. These labels are coarse-grained representations, projected into a vocabulary that humans can handle, of which kinds of sufficient statistics tend to be formed recurrently.

\section{Operating Profile and the Multi-Phase Inference Mechanism}

A candidate space does not yet think. It only tells us what could, in principle, become thinkable for a subject. A finite subject still has to move through that space: to foreground some candidates, explore others, and stabilize a few. This movement is where a world model begins to acquire its recognizable style.

In Section~4 we derived, from the admissible pair $\mathcal A_t^\alpha$, the candidate sufficient statistic space $\mathcal C_t^\alpha$ together with its conditioning bases and resolutions, and then the phase-formation space ${\Zphase}_t^\alpha$ as the space of its constitutive degrees of freedom. The present section formalizes how a finite subject actually operates over this candidate space. The question is not whether candidates exist, but which of them are raised to processing, which are explored, and which are stabilized.

\subsection[Derivation of the Operating Profile theta]{Derivation of the Operating Profile $\theta$}

A finite subject cannot simultaneously explore the entire candidate sufficient statistic space $\mathcal{C}_t^\alpha$ that is in principle constructible. Moreover, under the action constraint $C_{\mathrm{act}}$, the subject must select which candidates to foreground, which to explore, and which to stabilize. From this requirement we derive the operating profile $\theta$.

In this paper, the operating profile is defined as a triple of operating fields on the candidate sufficient statistic space,
\begin{equation}
\theta_t^\alpha=(r_t^\alpha,e_t^\alpha,s_t^\alpha),
\end{equation}
where
\begin{align}
r_t^\alpha &: \mathcal{C}_t^\alpha \to \mathbb{R},\\
e_t^\alpha &: \mathcal{C}_t^\alpha \to \mathbb{R}_{\geq 0},\\
s_t^\alpha &: \mathcal{C}_t^\alpha \to \mathbb{R}_{\geq 0}.
\end{align}

The foregrounding field $r_t^\alpha(c)$ expresses the degree to which a candidate $c=(Y,\psi,\rho,T)$ tends to be raised to the subject's processing. The exploration field $e_t^\alpha(c)$ expresses the degree to which processing tends to extend from that candidate to neighboring estimation targets, conditioning bases, resolutions, and state representations. The stabilization field $s_t^\alpha(c)$ expresses the degree to which the candidate tends to be maintained and consolidated as an existing state representation.

Under this definition, the operating profile $\theta$ is not an independently posited psychological trait parameter. It is derived as the control structure required for a finite subject to operate over the candidate sufficient statistic space. Subjective differences appear as differences in which $(Y,\psi,\rho,T)$ can be constructed at all, and in which candidates are foregrounded, explored, and stabilized.

\subsection{Conditioning-Basis Direction and Resolution Direction}

When needed, the foregrounding field $r_t^\alpha$ can be decomposed into a conditioning-basis direction and a resolution direction. This decomposition does not introduce additional operating profiles; it makes explicit which constitutive component of the candidate sufficient statistic $c=(Y,\psi,\rho,T)$ carries the foregrounding tendency. For instance, one may write
\begin{equation}
r_t^\alpha(c)
=
r_{\mathrm{ref},t}^\alpha(\psi\mid Y,O_t^\alpha)
+
r_{\mathrm{res},t}^\alpha(\rho\mid Y,\psi,O_t^\alpha)
+
r_{T,t}^\alpha(T\mid Y,\psi,\rho,O_t^\alpha).
\end{equation}
This decomposition is not mandatory, but it is expositorily useful. Here $r_{\mathrm{ref}}$ represents which conditioning basis the subject tends to employ, and $r_{\mathrm{res}}$ represents which rate--distortion trade-off, that is, which resolution, the subject tends to use in forming state representations.

For the exploration field $e$ and the stabilization field $s$, there is no need to introduce separate symbols for each reference or resolution direction, since they are already defined as functions on the candidate $c=(Y,\psi,\rho,T)$. Thus, a subject who is exploratory with respect to internal conditioning bases but stabilizing with respect to institutional conditioning bases can be expressed by differences in the values of
\begin{equation}
e_t^\alpha(Y,\psi_{\mathrm{endo}},\rho,T)
\quad\text{and}\quad
s_t^\alpha(Y,\psi_{\mathrm{inst}},\rho,T).
\end{equation}

\subsection[Profile State Omega]{Profile State $\Omega$}

The operating profile $\theta$ is the core of foregrounding, exploration, and stabilization. It does not, however, exhaust the subject's full profile state. State representation formation also involves the mode of representation formation, error costs, thresholds for consciousness and action, plasticity, and directional maturity. In this paper we therefore bundle the profile state as
\begin{equation}
\Omega_t^\alpha=(\theta_t^\alpha,\lambda_t^\alpha,q_t^\alpha,\zeta_t^\alpha,\sigma_t^\alpha).
\end{equation}
Here $\lambda$ is the plasticity/stabilization parameter, which specifies the degree to which the operating fields are updated by experience or held fixed. The parameter $q$ is the firing parameter, which bundles error costs, the threshold for entering consciousness, and the threshold for action. The parameter $\zeta$ is the representation-formation profile, which governs the mode of state-representation formation and the possibilities of compression, symbolization, and externalization. Finally, $\sigma$ is directional maturity, expressing how stably and precisely a given configuration of candidate sufficient statistics can be operated.

The directional compatibility $\mu$ used in subsequent sections is not an independent component of $\Omega$. It is a derived evaluative quantity that measures how well a candidate sufficient statistic $c$ aligns with a receiver's operating fields, representation-formation mode, and maturity. That is, $\mu(c;\Omega)$ is used as an index of whether the candidate $c$ tends to be processed exploratorily and receptively within the receiver's world model, or whether it tends to elicit rejection and defensive reactions.

When needed, the representation-formation profile can be decomposed as $\zeta=(\chiop,\tau,\kappa)$. Here $\chiop$ is the control of how estimation-target candidates are formed; $\tau$ is the map that, given a conditioning basis and resolution, forms the state representation $T$; and $\kappa$ is the control of compressibility, symbolization, and externalization. The resolution $\rho$ is not treated as the first component of $\zeta$; it is part of the candidate $(Y,\psi,\rho,T)$. This convention prevents conflation of the estimation-target formation control $\chiop$ with the resolution $\rho$.

Through this organization, $\theta$ is kept compact as the core that specifies operating directions on the candidate sufficient statistic space, while $\lambda,q,\zeta,\sigma$ are gathered into $\Omega$ as peripheral controls. In this paper, therefore, we do not pack operating direction, plasticity, firing threshold, representation formation, and maturity into a single $\theta$; we separate their roles as distinct internal components of $\Omega$.

\subsection{Prediction Error and Post-Operating State}

When the subject makes predictions on the basis of a candidate $c=(Y,\psi,\rho,T)$, the prediction error can be defined as
\begin{equation}
L_t^\alpha(c)
=
\ell\!\left(Y_t^\alpha,\hat{Y}_t^\alpha(T_t^\alpha)\right),
\end{equation}
or instantiated as a negative log-likelihood, a reconstruction error, a KL divergence, a free energy, or a similar quantity.

After observations are processed, which candidates were actually foregrounded, explored, and stabilized is recorded as the post-operating state
\begin{equation}
\tilde{\theta}_t^\alpha=(\tilde r_t^\alpha,\tilde e_t^\alpha,\tilde s_t^\alpha).
\end{equation}
Here $\tilde r,\tilde e,\tilde s$ are operating fields of the same type as those in $\theta$; they describe the foregrounding, exploration, and stabilization that were actually fired after observation processing. Through repetition of post-operating states, the long-term operating profile $\theta$ and the profile state $\Omega$ become progressively stabilized.

\subsection{The Generative Loop of the Multi-Phase Inference Mechanism}

On the basis of the derivations above, the Multi-Phase Inference Mechanism can be organized as the following generative loop:
\begin{equation}
O_t^\alpha
\rightarrow
\mathcal{C}_t^\alpha
\rightarrow
(Y_t^\alpha,T_t^\alpha,L_t^\alpha)
\rightarrow
\theta_t^\alpha
\rightarrow
P_t^\alpha
\rightarrow
y_t^\alpha
\rightarrow
O_{t+1}^\alpha
\rightarrow
\Delta_t^\alpha
\rightarrow
W_{t+1}^\alpha .
\end{equation}

In this loop, estimation targets do not arise directly from the observation sequence. The candidate sufficient statistic space $\mathcal C_t^\alpha$ that is constructible by the finite subject is first formed. Within it, those candidates that are foregrounded by the operating profile $\theta$ are activated as the estimation target $Y$, the state representation $T$, and the prediction error $L$. From this activation, plan candidates $P$ are generated -- including thought, suspension, reinterpretation, explanation, utterance, bodily action, and coordination -- and are returned to the environment as the output $y$. The resulting feedback $\Delta$ updates not only the generative model $M$ in the narrow sense, but the world-model state in the broad sense $W$, which comprises the phase-formation space $\Zphase$, the operating profile $\theta$, and the profile state $\Omega$:
\begin{equation}
W_t^\alpha=
\left(M_t^\alpha,{\Zphase}_t^\alpha,R_t^\alpha,\Omega_t^\alpha\right).
\end{equation}

Here $R_t^\alpha$ is the foregrounding gradient field, which, given the operating profile $\theta_t^\alpha$ and the prediction error $L_t^\alpha$, returns the direction in which processing tends to unfold on the candidate sufficient statistic space. More precisely,
\begin{equation}
R_t^\alpha:
\mathcal{C}_t^\alpha\times \mathcal{L}_t^\alpha
\longrightarrow
T\mathcal{C}_t^\alpha,
\end{equation}
where $T\mathcal{C}_t^\alpha$ denotes the tangent space over the candidate space and is used here as a direction field indicating which candidates processing tends to move toward.

\section{The Cognitive Shift Opened by the Multi-Phase Inference Mechanism}

In Section~3 we formulated the conditions for the emergence of state representations, in Section~4 the candidate sufficient statistic space and the phase-formation space, and in Section~5 the operating profile defined on that candidate space. Together, these position the Multi-Phase Inference Mechanism as a structure that arises when finite subjects construct approximate sufficient statistics. This section organizes the implications.

\subsection{The question itself shifts --- endogeneity of estimation targets}

The differences between subjects appear, prior to differences in conclusions, as differences in what is constituted as an estimation target. If a subject cannot construct an approximate sufficient statistic $T$ for a given $Y$, then $Y$ does not become a stable object of recognition. Disagreement over the same observation sequence is therefore not only a difference in answers to the same question, but a difference in what questions can be constructed in the first place.

\subsection{From error to non-identifiability --- relocating epistemic diversity}

Under SIA, subjects who reach different conclusions from the same observations are easily framed as exhibiting information deficits, misunderstanding, irrationality, or bias. Under MIA, however, the space of approximate sufficient statistics that can be constructed differs across subjects. Disagreement therefore does not immediately signify error. It points to a non-identifiability concerning which $T$ could be constructed for which $Y$.

\subsection{From disorder to inference phases --- structuring diversity}

Of course, not all differences are equally valid. MIM is not a theory that legitimates any subjective view whatsoever. Rather, it asks which approximate sufficient statistic a subject has constructed, under which distortion function the subject preserved which errors, and which errors were discarded. Epistemic diversity is thereby treated not as disorderly relativism, but as diversity structured by information bottleneck and rate--distortion constraints.

\subsection{Bridge to state-representation alignment}

If different subjects construct different sufficient statistics, mutual understanding cannot consist in sharing identical representations. What is required, rather, is that one subject's state representation be transformed into a form processable within another subject's world model. The alignment map $\Phi$ introduced in Section~9 addresses this problem. $\Phi$ is not a map that guarantees agreement; it is a device that makes explicit what is preserved, what is coarse-grained, and what is lost as transformation loss between different approximate sufficient statistics.

\subsection{Bridge to the history of philosophy, typology, social fragmentation, and AI alignment}

The history of philosophy, cognitive typology, social fragmentation, and AI alignment, all taken up in the following sections, are applications of the same structure built through this section. Philosophical traditions can be read as long-term records of which errors were preserved and which distortions were rendered impermissible. Cognitive typology can be re-described as repeatedly stabilized coarse-graining patterns in sufficient-statistic formation. Social fragmentation arises when different sufficient-statistic formations collide without alignment maps. AI alignment, accordingly, is redefined not as convergence to a single value but as a design problem of displaying the transformation loss between heterogeneous state representations and supporting multiple error-selection processes.

\section{Reading Error Structures in the History of Philosophy}

This section does not extend the core formalization of MIM. It offers an error-structure reading of several philosophical traditions: a way of asking which estimation targets, state representations, prediction errors, error costs, and update paths each tradition tried to preserve. The purpose is not to rewrite the history of philosophy as an appendix to MIM. The purpose is to show that long-standing philosophical oppositions can be reread as conflicts among different error-detection structures that resist reduction to a single regime of correctness.

Why did philosophical disputes fail to end over two millennia? Not because each position was simply mistaken. Rather, each tried to preserve prediction errors that other positions would erase. The aim of this section is not to ``classify'' the history of philosophy in the vocabulary of the present paper, but to recover what each tradition tried to preserve. The formal apparatus of the paper is used not to reduce these traditions, but to non-reductively preserve them. The history of philosophy can be read, in this limited sense, as a long record of the errors that human world models have tried not to lose.

\subsection{Method of this section --- reading error structures rather than surveying the history of philosophy}

The question of this section is not to exhaustively determine ``what each philosopher really said.'' The question is: for an AI alignment theory based on MIA, which error sensitivity has each tradition preserved? For example, the opposition between empiricism and idealism is not merely the slogan-level opposition between ``experience'' and ``idea.'' Empiricism tried to preserve the possibility that claims are corrected by observation and external resistance. Idealism tried to preserve the error that empirical fragments alone cannot explain the unity of the world, meaning, normativity, and subjectivity.

Accordingly, this section reads each tradition from the standpoint of (i) the foregrounded estimation target, (ii) the preserved state representations, (iii) the prediction errors deemed inadmissible, (iv) the coarse-grained domain in which they are processed, and (v) the transformation loss under the alignment map $\Phi$. This method does not reduce the history of philosophy to MIM. Rather, it is an attempt to connect the non-commutative error structures that the history of philosophy has preserved to a vocabulary that is processable within an AI alignment theory.

Empiricism, idealism, structuralism, existentialism, semiotics, phenomenology, philosophy of language, philosophy of science, linguistic relativity, pragmatism, and dual-process theory have each, in different ways, preserved the points where the world fails to pass through cleanly for a subject. The formal apparatus of this paper describes the prediction errors preserved by these traditions as differences in the estimation target $Y$, the state representation $T$, the prediction error $L$, the phase-formation space $\Zphase$, the foregrounding field $R$, and the profile state $\Omega$. Through this, philosophical opposition is rearranged not as a debate that should converge on a single correct answer, but as a problem that requires an alignment map $\Phi$ between different world-model update paths.

The terms ``empirical,'' ``ideational,'' ``structural,'' and ``existential'' used here are not axioms of MIM. They are auxiliary vocabulary for explaining the representative coarse-grained domains that appear on $\Zphase$. The empirical domain foregrounds observable events, bodily feedback, and local operability. The ideational domain foregrounds meaning, ideas, possible worlds, and long-range conceptions. The structural domain foregrounds institutions, forms, causality, and reproducible order. The existential domain foregrounds self, other, responsibility, injury, recognition, and trust. These are not discrete entities but coarse-grained maps that allow humans to handle a continuous phase-formation space.

\subsection{Platonic pole and Aristotelian pole}

Prior to empiricism and idealism, one can consider a deeper bifurcation: the Platonic pole and the Aristotelian pole. The Platonic pole \cite{plato380} tries to preserve the ideas, forms, and universal structures behind variable phenomena. The Aristotelian pole \cite{aristotle350} tries to preserve particulars, practice, generation and change, observation of nature, and teleological arrangement.

In the vocabulary of MIM, this bifurcation can be read as a deep philosophical-historical expression of the resolution direction $r_{\mathrm{res}}$. The pole of micro-resolution, operation, and practical engagement preserves how particular phenomena change and how they are grasped within the space of available actions. The pole of coarse-graining, ideation, and universalization, on the other hand, asks what is preserved beyond particular phenomena as identity, form, norm, or meaning. In modern terms, the former can be called empiricist and the latter idealist, but at the substrate there is already the tension between the Platonic and Aristotelian poles.

This deep bifurcation locates the modern oppositions treated in the subsequent sections within a longer time span. Empiricism/idealism, structuralism/existentialism, and pragmatism/phenomenology each rehearse, in their own vocabularies, the question of where the center of gravity of state representation should be placed between these two poles.

\subsection{Empiricism and idealism --- preserving external resistance and internal form}

What empiricism tried to preserve is the fact that the world offers resistance to the subject from outside \cite{locke1690,hume1748}. No matter how refined ideas or speculations may be, the responsiveness given through sense experience, the updating of observation, and local operability cannot be replaced by internal coherence alone. Empiricism strongly preserved this inviolability as a foundation of knowledge.

What idealism tried to preserve is the fact that, for resistance to be constituted as cognition, the subject's own forms are required \cite{kant1781,hegel1807}. Without the intuitive forms of space and time, the categories of understanding, the conceptual system, and ideas, sensation is a flow of unsorted stimuli, not cognition. Idealism strongly preserved this constitutivity as a foundation of knowledge.

In the framework of this paper, empiricism can be redescribed as the position that pursues the high-precision refinement of state representations extracted by the empirical domain --- observational pathways, available actions, and local feedback. Idealism can be redescribed as the position that pursues the high-precision refinement of state representations extracted by the ideational domain --- long-range representations, concepts, possible worlds, and ideas. Their opposition arose not because one of them is correct, but because each tried to ground different sufficient statistics on its own as the sole foundation.

If a world model minimizes only the prediction error of the empirical domain, the accumulation of experience is obtained, but the long-range map of meaning and possibility is lost. Conversely, if it minimizes only the prediction error of the ideational domain, a system of ideas is obtained, but the point of contact that absorbs resistance from the world is lost. The two must coexist non-reductively while remaining mutually translatable.

\subsection{Structuralism and existentialism --- non-reductive coexistence of relational system and irreplaceability}

What structuralism tried to preserve is the discovery that meaning arises not from the inherent properties of particular events, but from a system of differences \cite{saussure1916,levi1958,foucault1966}. Saussure's linguistics, L\'{e}vi-Strauss's analysis of kinship structures, and Foucault's theory of the episteme each extracted structural relations that exceed individual phenomena and showed that those structures make the meaning of individual events possible. Structuralism established this methodological rigor as one of the most powerful tools acquired by the humanities.

What existentialism tried to preserve is the irreplaceability in the relations of self and other \cite{kierkegaard1849,heidegger1927,sartre1943,levinas1961,buber1923}. Death, responsibility, trust, response, and injury are, from a third-person systemic view, one position among many, but for the involved party they are events that cannot be substituted. Kierkegaard's single individual, Heidegger's being-in-the-world, Sartre's responsibility, Levinas's face of the other, and Buber's I-Thou relation are all forms of resistance to being processed as exchangeable positions within a structural relational system.

The framework of this paper does not reduce this opposition to either side. The state representations extracted in the structural domain --- relational form, systems of difference, and reproducible order --- are indispensable for handling institutions, law, procedure, logic, and convention. The state representations extracted in the existential domain --- the inner-world variables of self and other, relational prediction errors, the burden of responsibility, and the situation of being involved --- equally cannot be reduced to the prediction errors of institutions and rules.

The existential appeal is not, as structuralists may say, the residue of feeling that has no essence. Relational prediction errors --- who was injured, who was made responsible, who was silenced in this setting --- cannot be reduced to structural prediction errors such as which rule was broken or whose system was made incoherent. Conversely, structural order cannot be dismissed, as existentialists may say, as cold formalism. Relations without structure end in momentary resonance; structures without relations end in an order that is for no one.

What is required between structuralism and existentialism is therefore not reduction in either direction. It is to align state representations assembled in the structural domain into a form processable in the existential domain, and to establish the reverse alignment as well. The alignment map $\Phi$ of this paper is a device for constituting this non-reductive coexistence as processability without erasure.

\subsection{Social contract and political philosophy --- prehistory of world-model alignment}

This section is not a survey of political philosophy. In constructing an AI alignment theory based on MIA, it extracts only those elements of the social contract tradition that can be translated into one of state-representation alignment, the alignment map $\Phi$, or preference aggregation. Hobbes, Locke, and Rousseau can be redescribed as positions that, from different angles, describe what happens when subjects with mutually heterogeneous world models share the same social space.

What Hobbes tried to preserve is the question of how a state in which the alignment map $\Phi$ has not been established is constituted \cite{hobbes1651}. The state of nature is not a gathering of morally degenerate individuals; it can be described as an informational state in which the map $\Phi$ that predicts the inner states of others is not functioning. When it is mutually invisible what others raise as their estimation targets and what they treat as high-cost prediction errors, others are constituted, for each subject, as unpredictable sources of risk. The war of all against all that Hobbes described is not a naturalization of violent tendencies; it is an equilibrium reached by rational subjects in a state where state-representation alignment has collapsed. The transfer of right to the sovereign can be read as an attempt to dissolve this informational state by compressing it into a single authority. In the framework of MIM, this corresponds to a solution that short-circuits the problem of generating an alignment map by collapsing into a single profile rather than generating the map itself.

What Locke tried to preserve is the observation that, even starting from the state of nature, a network of state representations shareable across multiple subjects can be formed through the mediation of property, labor, right, and consent \cite{locke1689}. The mixing of labor into land, the mutual recognition of property rights, and explicit or tacit consent can be redescribed as the work of erecting collective state representations that cut across the individual world models of each subject. What is important in Locke's argument is that order is constituted not by a single sovereign, but by the very process in which partial $\Phi$s are built up among horizontally arranged subjects. In MIM vocabulary, this is positioned as the formation of a collective $\Phi$ network. The problem that Hobbes tried to solve by compression into a single authority is solved by Locke through the accumulation of distributed alignment maps.

What Rousseau tried to preserve is the non-identity, arising at the level of preference aggregation, between the sum of individual wills and the general will \cite{rousseau1762}. Even if the local preferences of each subject are aggregated by majority rule, a coherent world-model update path for the collective is not guaranteed. In the framework of MIM, this corresponds to the fact that, among the alignment maps, $\Phi_U$ --- the one concerning preference aggregation --- has its own peculiar difficulty. The operation of mapping value functions defined on each subject's world model back onto a single value function on a shared estimation target space is not a special case of state-representation alignment $\Phi$; it is a different kind of alignment problem. Rousseau's general will can be read as a rule for this different kind of alignment problem.

Placing the three side by side, the social contract tradition is rearranged as a prehistory of AI alignment. Hobbes extracted the state in which $\Phi$ does not hold; Locke extracted the formation of a distributed $\Phi$ network; Rousseau extracted the specificity of the preference-aggregation map $\Phi_U$. If the design problem of AI alignment is to support processability among heterogeneous world models without compressing them into a single authority, a single value function, or a single progressivist historical view, then the three social-contract thinkers are the version in which that design problem was first written down in the language of political philosophy.

\subsection{Traditionalist conservative thought and long-term error sensitivity}

This section positions traditionalist conservative thought as an inferential phase that preserves cultural, institutional, bodily, and communal error sensitivities accumulated over long periods. The figures treated here are Burke and Oakeshott. Hayek's theory of distributed knowledge concerns a different axis --- the social distribution of knowledge and the alignment map mediated by the price mechanism --- and is treated in Section 10's discussion of social implementation.

What Burke tried to preserve is a kind of knowledge whose error-correction history, accumulated across generations, is held in institutions, customs, and prejudices in forms that cannot be explicitly articulated \cite{burke1790}. When Burke wrote that ``prejudice is accumulated wisdom,'' what he tried to preserve was not the defense of individual biases, but the observation that a kind of state representation that cannot easily be derived from explicit, rule-based world-model updating is accumulated on the side of custom, institution, and communal memory. Custom is the product of the long-term fixation of the estimation-target-formation map $\rho$ and the state-representation extractor $\tau$ from observation sequences, and that fixation is inherited across generations beyond the lifespan of an individual subject.

Burke's famous idea that ``society is a partnership between those who are living, those who are dead, and those who are to be born'' can be redescribed in MIM vocabulary: the state representation and the profile state $\Omega=(\theta,\lambda,q,\zeta,\sigma)$ are inherited across generations as an error-correction history that accumulates beyond the lifespan of individual subjects. Burke's caution toward the French Revolution was directed at the fact that revolutionary design adopts only the explicit world models that can be constructed within a short time span, while making invisible those kinds of prediction errors --- collapse of institutions, dissolution of communities, loss of intergenerational trust --- that are observable only over longer time spans. In the vocabulary of this paper, this can be read as resistance to having long-term and coarse-grained error-absorption paths overwritten by short-term and explicit alignment.

What Oakeshott tried to preserve is the non-reductive distinction between the kind of knowledge that can be explicitly and rule-followingly externalized within a world model (technical knowledge) and the kind that is harder to externalize and is held in bodily, context-dependent forms (practical knowledge) \cite{oakeshott1962}. Oakeshott repeatedly referred to the fact that a cook cannot cook from a recipe alone, that a politician cannot govern from a manifesto alone. In MIM vocabulary, technical knowledge corresponds to state representations with high externalizability and low context-dependence; practical knowledge corresponds to state representations with low externalizability and high context-dependence. The latter accumulates as the long-term maturation of $\tau$ --- which carves out state representations from observation sequences --- and the implicit maturation of the directional maturity $\sigma$.

What is important is that practical knowledge is not ``an unrefined precursor of technical knowledge.'' The two are different kinds of state representation, and although they are mutually translatable, they cannot be fully reduced to one another. What Oakeshott showed in his critique of rationalist politics is the observation that when institutions are redesigned on the assumption that practical knowledge can be fully transcribed into technical knowledge, the long-term error sensitivity preserved in the residue of transcription is lost. In the framework of this paper, this corresponds to the description that the directional maturities $\sigma$ along the externalizability and context-dependence directions are formed under different constraints.

That said, what must be most articulately stated in the MIM-style redescription of traditionalist conservative thought is its two-sidedness. State representations accumulated over the long term preserve the ability to detect a certain kind of error, while at the same time making another kind of error invisible. Custom inherited across generations is a map of errors detected by past generations up to that point, and it does not initially register as errors those that are newly generated thereafter --- new subject populations, new relational forms, new bodily realities, new forms of labor. Indeed, the very fact that tradition has functioned stably is also the result of having structurally excluded certain errors from observation. Cases in which exclusion and inequality have been inherited as a ``natural order'' over long periods show that tradition is simultaneously a device that preserves errors and a device that renders errors invisible.

Therefore, the question to ask is not whether to defend or destroy tradition. It is which errors tradition preserves and which errors it renders invisible. Burke and Oakeshott can be redescribed as positions that strongly preserve the long-term error sensitivities overlooked by explicit, short-term rationalization, while the framework of MIM equally describes the two-sidedness whereby those same positions may also structurally render new errors invisible. The alignment map $\Phi$ of this paper does not give priority to either tradition or innovation. It is a device for aligning both --- as collections of state representations preserved under different time spans, different externalizabilities, and different context-dependences --- into mutually processable forms.

\subsection{Semiotics --- the differentiality of meaning and the formation of the estimation target}

What semiotics tried to preserve is the discovery that meaning arises not from a natural correspondence between sign and object, but from the differences and interpretive processes internal to a sign system. Saussure's arbitrary coupling of signifier and signified \cite{saussure1916} and Peirce's triadic relation of sign, object, and interpretant \cite{peirce1958} both extracted the differential and relational structures required for meaning to be established.

In the framework of this paper, the structure described by semiotics is redescribed in terms of the linkage among the estimation-target-formation map $\rho$, the state-representation extractor $\tau$, and the estimation target $Y$. The signifier corresponds to the action of $\tau$ that carves out a state representation from an observation sequence; the signified corresponds to the estimation target $Y$ predicted by that state representation; the interpretant corresponds to the reapplication of $\rho, \tau$ in another subject or another phase. The Peircean chain of interpretants can be described as a process in which a state representation enters the observation sequence of the next subject, and an estimation target is again raised by that subject's $\rho, \tau$.

\subsection{Language, meaning, translation, and incommensurability}

Whereas the early Wittgenstein investigated the conditions under which formal logic can picture the world \cite{wittgenstein1922}, what the later Wittgenstein tried to preserve is the insight that the meaning of a word depends not on definition or correspondence, but on its use within language games and forms of life \cite{wittgenstein1953}. What Quine tried to preserve is the indeterminacy of translation: the choice of which predicates to raise from an observation sequence is not uniquely fixed \cite{quine1960}. What Kuhn tried to preserve is the incommensurability whereby subjects belonging to different scientific paradigms cannot fully map each other's claims into the same state representation \cite{kuhn1962}.

What these three positions preserved is the discovery that meaning, understanding, and translation are structural processes that cannot be reduced to the correspondence relations of formal logic. The meaning of an utterance is constituted not only through the literal sentence but through implicature, cooperative principles, and contextual expectation \cite{grice1975}. Conceptual categories, moreover, are formed through embodied experience and metaphorical structure \cite{lakoff1987}. Even when a word has the same dictionary meaning, it may be processed in the structural domain as a rule violation, in the existential domain as an insult, in the empirical domain as an on-site constraint, and in the ideational domain as a betrayal of an ideal. In this case, the identity of the word does not guarantee the identity of the state representation.

In the framework of this paper, the use-dependence of meaning is described by the fact that the estimation-target-formation map $\rho$ and the state-representation extractor $\tau$ depend on the subject's profile state $\Omega$ and observational context. The indeterminacy of translation is the non-identifiability whereby the same observation sequence can give rise to different estimation targets $Y$ and state representations $T$ under different combinations of $(\rho,\tau)$. Incommensurability can be described as a state in which $\Phi$ acts only partially between the state-representation spaces of two subjects.

What is important is that $\Phi$ is not an operation that ``makes meanings the same.'' $\Phi$ is an operation that, under the formation rules of different $(\rho,\tau)$, aligns a state representation into a sufficient statistic processable on the receiver's side. The difficulties pointed out by Quine and Kuhn do not disappear, but by making the structure of those difficulties explicit, one can ask which mapping at which layer is functioning, and which is not.

\subsection{Linguistic relativity and social profile formation}

What Sapir and Whorf tried to preserve is the observation that speakers of different native languages categorize the world differently, and that this categorization affects cognition, inference, and action \cite{sapir1929,whorf1956}. Differences in color vocabularies, the presence or absence of tense, the structure of grammatical gender, the development of honorifics, and systems of nominal definiteness affect the cognitive segmentation of what is distinguished, what is grouped together, and what is pushed into the background.

In the framework of this paper, the lexical and grammatical systems of a native language impose long-term constraints on foregrounding biases over the reference-target space, on preferences for grain and range in the resolution space, and on the representation-formation profile $\zeta$. A linguistic community is one social device that inherits and fixes the profile state $\Omega$ across generations. This does not deny cognitive universality; it describes the fact that, while the basic structure of $\Zphase$ is shared, the foregrounding distribution over it is biased by linguistic community.

\subsection[Phenomenology, epoche, and eidetic reduction: conscious redesign of rho and tau]{Phenomenology, epoche, and eidetic reduction --- conscious redesign of $\rho,\tau$}

What phenomenology tried to preserve is attention to the very experiential structure by which an object appears to a subject \cite{husserl1913,heidegger1927,merleau1945}. Against the direction in which the natural sciences treat objects of observation as detached from the subject, and in which psychology processes subjective experience as introspective report, phenomenology took the very interface between subject and object --- ``how'' an object appears to a subject --- as its object of description. Husserl's epoch\'e and eidetic reduction are methodological operations for reflectively extracting the processes of object constitution that we ordinarily carry out without awareness.

In the framework of this paper, the epoch\'e is the operation of temporarily bracketing automatic routing and reflectively asking how an observation sequence is being constituted as a particular estimation target. The eidetic reduction further advances this inquiry, positioning itself as the work of discerning the sufficient statistic $T$ that must be preserved in order for an object to be established as that object. That is, the phenomenological method is not a method for changing the observation sequence itself; it is a method for consciously redesigning the extraction rules $\rho,\tau$ by which estimation targets and state representations are raised from an observation sequence.

The phenomenological method can also be read as a methodological refinement of coarse-graining-oriented strategies for resolving prediction error. The epoch\'e temporarily suspends the immediate processing of particular observations and reflectively resets the granularity of object constitution itself. The eidetic reduction re-identifies, under higher-order coherence, the sufficient statistic that must be essentially preserved. This is not short-term error absorption through action, but long-term error absorption through abstraction, integration, and extension of range.

\subsection{Pragmatism and the intentional stance --- action tests the world model}

What pragmatism tried to preserve is the inseparability of cognition and action: the truth or falsity of cognition should be judged by its practical consequences \cite{james1907,peirce1958,dewey1925}. Peirce, James, and Dewey depicted cognition as a process tested not only by observational verification from the world, but also by the consequences that action generates. Dennett's intentional stance preserved the functional usefulness of attributing beliefs and desires to others \cite{dennett1991}.

In the framework of this paper, pragmatism's practical verification is described as the next observation sequence $o_{t+1}$ generated by action through the environment, and as the world-model update driven by its prediction error. The intentional stance is redescribed as an operation that raises the inner state of the other as an estimation target within one's own world model and connects it, through the alignment map $\Phi$, into a processable state representation. Cognition is not a representation closed within the head; it is a process of world-model updating tested by action and re-observation.

\subsection{Action and reflection --- prediction-error absorption by resolution and time span}

Action is the kind of knowledge that absorbs prediction error through intervention in the environment. The subject acts on the environment and, through the local feedback obtained as a result, short-term aligns local incoherences. This point also connects to the discussion of affordance --- that the environment offers possibilities for action to the subject \cite{gibson1979}. Even in situations where, at a coarse-grained level, plans appear to have broken down, fine on-site adjustment and bodily operation can in fact restore coherence with prediction.

In contrast, there is also a process that retains and reconstructs, over longer time spans, the incoherences that cannot be immediately resolved through action. Experiences in which contradictions or unresolved issues remain when only the details are examined may still cohere within the longer range of coarse-grained narratives, forms, responsibilities, and meanings. Reflection here is not the absence of action. It is another active form of world-model updating that absorbs prediction error through long-term state-representation reconstruction.

In the vocabulary of this paper, the former is short-term, local alignment; the latter is long-term, coarse-grained alignment. Neither is the matured form of the other. The philosophical significance of the Multi-Phase Inference Mechanism lies in the fact that, as differences in foregrounding tendency over the resolution space, alignment through action and alignment through reflection can be treated within the same framework.

\subsection{Bridge to cognitive science --- Kahneman's dual process}

What Kahneman's dual-process theory tried to preserve is the empirical observation that multiple cognitive processing modes operate in parallel within the same individual, and that they sometimes lead to mutually contradictory conclusions \cite{kahneman2011}. The distinction between immediate, intuitive System 1 and deliberative, analytical System 2 has played an important role in cognitive science, judgment and decision-making research, and behavioral economics.

In the framework of this paper, the dual process can be redescribed as a low-dimensional approximation of intra-individual multi-phase inference. MIM addresses not only differences in processing speed and deliberativeness, but also differences in what is raised as the estimation target, which state representations are preserved, and which prediction errors are treated as high-cost. Dual-process theory is therefore an important bridge by which MIM connects to cognitive science, but in this section it is positioned as an auxiliary bridge.

\subsection{From the history of philosophy to state-representation alignment}

As the above redescriptions make clear, the traditions in the history of philosophy can be re-read not as a collection of mutually exclusive positions, but as a collection of world-model update strategies that have preserved different prediction errors. Empiricism preserved resistance from the world; idealism preserved cognitive forms; structuralism preserved systems of difference; existentialism preserved irreplaceable involvement; semiotics preserved the differentiality of meaning; phenomenology preserved object constitution; pragmatism preserved verification by action; and dual-process theory preserved differences in intra-individual processing modes.

MIM does not declare any one of these traditions the winner. Rather, it formalizes the prediction errors that each has preserved as differences in world model, state representation, error cost, and update path. Through this, philosophical opposition is rearranged not as a clash between correct and incorrect theories, but as a state in which different state representations have not yet been made mutually processable. What becomes necessary here is state-representation alignment, and the alignment map $\Phi$.

In this sense, this paper does not erase philosophical oppositions. It formalizes the conditions under which oppositions are generated. That is, it provides a foothold on which oppositions expressed in different vocabularies can be compared on the same formal space.

\medskip
\noindent This chapter should therefore be read as an error-structure interpretation rather than as a history of philosophy. Its claim is limited but important: philosophical traditions endure because each preserves errors that a rival form of correct inference would tend to exclude. This prepares the later discussion of society and AI alignment, where the problem becomes institutional and technical rather than merely interpretive: how can heterogeneous error-detection structures remain mutually processable without being collapsed into one another?

\section{Cognitive Typologies as Low-Dimensional Error-Structure Approximations}

\subsection{Typology as a coarse-grained map of inferential diversity}

This chapter does not treat typological systems as foundations of the present theory. Rather, it examines them as historically successful attempts to compress recurrent patterns of inferential diversity into tractable representations. In other words, typologies are treated here as low-dimensional, error-structure approximations of a much richer operating space.

This section therefore examines how far the cognitive differences that cognitive typology has empirically observed can be redescribed by the formal apparatus of MIM. Humans have long given names to differences such as: some people respond strongly to concrete experience, some pursue possibility and ideas, some see through structure and institution, and some treat damage to relations or dignity as a high-cost error. These vocabularies are not precise survey maps. But if people have repeatedly given names to the same ridges and river lines, there must be something in the terrain. The aim of this section is not to adopt that map as it is, but to redraw it under a more testable coordinate system.

Here, what is to be preserved and what is to be updated should be separated. What is to be preserved is the phenomenon, observed by Jung, Kepinski, Augustinaviciute, and Myers, that ``even given the same stimulus, the way experience, judgment, attention, and relation are organized differs systematically by subject.'' What is to be updated is the form that handles such differences as fixed, discrete types or personality labels. MIM preserves the observation and updates the form.

This arrangement does not slight the observations of practitioners and readers who have valued typology. At the same time, it does not slight the measurability, reproducibility, and caution against excessive labeling that researchers and readers wary of fixed classifications have tried to preserve. Rather, it aims to relocate what each side has tried to preserve, in their different ways, into the relation between a continuous world model and a coarse-grained explanatory map.

In this section, Jungian S/N differences are redescribed primarily as weighting directions of resolution, range, and abstraction within $\Zres(\psi)$; T/F differences primarily as reference-target preferences toward $\Zexo$ and $\Zendo$; and extraversion/introversion as the explorative/expansive direction and the stabilizing/compressive direction over distributions of state representation. In this redescription, the sixteen types, the eight functions, and Model A are positioned as empirical approximations that coarse-grain a continuous, high-dimensional operating structure on $\Zphase$ into a low-dimensional map that humans can handle.

Therefore, the issue of this section is not ``whether humans can be classified into sixteen types.'' The issue is: which low-dimensional projection of the high-dimensional operating-profile space has how much explanatory power in which observational domain. Typology is repositioned not as a tool closed to the use of simply judging people, but as a provisional visualization that explains epistemic diversity to humans and makes it processable for AI.

\subsection[Four coarse-grained domains and two-direction eight cognitive elements]{Four coarse-grained domains $\times$ two-direction eight cognitive elements and inter-directional conflict}

Combining the four coarse-grained domains with two update directions yields eight cognitive elements.

\begin{longtable}{lll}
\toprule
Coarse-grained domain & Explorative/expansive direction & Stabilizing/compressive direction \\
\midrule
Empirical domain   & Se & Si \\
Ideational domain  & Ne & Ni \\
Structural domain  & Te & Ti \\
Existential domain & Fe & Fi \\
\bottomrule
\end{longtable}

This correspondence does not presuppose typology; it constructively redescribes the eight elements that post-Jungian typology has observed, starting from the four coarse-grained domains and the two update directions \cite{jung1921,myers1985,pietrak2018}.

\paragraph{Why the extraverted and introverted elements of the same phase conflict}

The extraverted and introverted elements belonging to the same coarse-grained domain share the region of state representation they handle. However, the update direction they prefer is opposite.

Ne and Ni both belong to the ideational domain. Ne treats the expansion of possibility in the ideational domain as pleasure. Ni treats the convergence of meaning in the ideational domain as pleasure. Therefore, to Ne, Ni appears to close down possibility; to Ni, Ne appears to scatter meaning.

Te and Ti both belong to the structural domain. Te opens structure through external data, implementation, measurement, and results. Ti compresses structure into definition, system, and internal consistency. Therefore, to Te, Ti appears too closed and unimplemented; to Ti, Te appears coarse and driven by external outcomes.

Fe and Fi both belong to the existential domain. Fe opens to the response of the setting, the propagation of emotion, and feedback from others. Fi stabilizes trust, boundaries, injury, and responsibility internally. Therefore, to Fe, Fi appears not open to the setting; to Fi, Fe appears to blur internal boundaries.

Se and Si both belong to the empirical domain. Se opens to new stimuli, kinetic possibility, and on-site change. Si compresses toward bodily state, comfort, and familiar environment. Therefore, to Se, Si appears as stagnation; to Si, Se appears to disrupt bodily stability.

Thus, the extraverted and introverted elements of the same phase share the object of directional maturity, while the subject's directional orientation points to opposite update directions; consequently the psychological evaluation reverses.

\subsection{Connection to Jungian functions}

The vocabulary of sensing, intuition, thinking, and feeling in Jungian typology can be redescribed in MIM not as basic phases of the world, but as coarse-grained domains of estimation target and state representation. Sensing corresponds to the empirical domain --- oriented toward observable events, bodily reality, and local feedback. Intuition corresponds to the ideational domain --- oriented toward possible worlds, ideas, and long-range scope. Thinking corresponds to the structural domain --- oriented toward structure, institution, causality, and formal relation. Feeling corresponds to the existential domain --- oriented toward self, other, relation, responsibility, dignity, and injury. However, this is not a one-to-one identification; it is a coarse-graining correspondence that maps a historical vocabulary onto the continuous space of MIM.

Furthermore, each domain has an explorative direction and a stabilizing direction. The explorative direction broadens state representation toward possibilities not yet fixed and unobserved structures. The stabilizing direction compresses state representation into existing models, known categories, and habitual explanations. Through this reading of ``four coarse-grained domains $\times$ two directions,'' the eight cognitive-function terms are reconstructed not as a fixed personality classification, but as directions of state-representation formation.

\subsection{Connection to constraint sequences in typology}

Jungian typology, MBTI-style function stacks, and the Model A of Socionics each have different formalizations, but they share a common structure: an ordering that describes ``which cognitive elements tend to come to the fore, and which tend to function in auxiliary, vulnerable, or background roles.'' This section calls this common structure the \textbf{constraint sequence in typology}, and, without adopting any particular typological system, shows how these can be redescribed on the continuous formal apparatus of MIM.

In the framework of this paper, the function positions and function stacks in typology can be read as combinations of directional compatibility $\mu$, directional maturity $\sigma$, the threshold of conscious awareness, and the allocation of cognitive resources for each direction. For example, the dominant function position in a given system can be read as a direction with high $\mu$, high directional maturity, and a low threshold of conscious awareness. The auxiliary position can be described as a direction that has high maturity but is less fixed than the dominant one, and is used for alignment with others or for support. The vulnerable position corresponds to a direction with low maturity in which, nevertheless, prediction errors are easily made conscious and low $\mu$ tends to surface. The receptive or developing position can be organized in terms of differences in self-maturation, receptivity to external input, and how directional compatibility emerges.

Formally, the constraint sequence of each subject can be introduced as
\begin{equation}
\Gamma^{\alpha}=(\gamma_1^{\alpha},\gamma_2^{\alpha},\ldots),
\quad
\gamma_i^{\alpha}\in\mathcal{K}_{\mathrm{coarse}}\times\{e,s\}
\end{equation}
Here, $\gamma_i$ is a rank-order description over the continuous quantity of directional strength, and the corresponding strength distribution is left open as an empirical question. $\gamma_i$ expresses which coarse-grained domain $K$ tends to take a relative first or second position with which directionality $e/s$. $\Gamma^\alpha$ is an auxiliary description --- an ordering structure derived from the central apparatus $(\theta^\alpha, \zeta^\alpha, \mu^\alpha, \sigma^\alpha)$ of this paper --- that does not presuppose any particular typological system. Once the first and second elements stabilize, the entropy of the remaining constraint-sequence distribution decreases, and a low-entropy attractor can be formed through finite cognitive resources, the trade-off between exploration and stabilization, inter-axis complementarity, and inter-phase interference. This structure provides an MIM-based interpretive route in cases where, within a given subject population, configurations such as MBTI-style function stacks or Socionics' Model A are observed.

What is important here is not to adopt Model A or the MBTI function stack itself as a final classification. What matters is that the vocabulary of function arrangement can be redescribed as a problem of constraint sequences --- which directions are foregrounded, and which are made auxiliary, vulnerable, or background. The unique derivation of each function position, the explanatory power of the sixteen-pattern approximation, and the quantitative evaluation of the extent to which the constraint sequences of multiple typological systems capture the same latent structure are tasks for subsequent research.

\medskip

Through this redescription, the long-standing opposition surrounding cognitive typology --- between those who defend typology and those who dismiss personality classification as pseudoscience --- becomes visible in a different way. The object of the dispute is not whether humans can be classified into fixed boxes. It is: which low-dimensional projection of the high-dimensional structure on $\Zphase$ has how much explanatory power in which observational domain. The systematic differences that Jung, Kepinski, Augustinaviciute, and Myers observed are repositioned, within the framework of this paper, not as personality labels, but as empirical samples of representative stable regions over the inferential-phase space. The problem of discrete fixed classification that critics have pointed out is resolved by the shift to a continuous formal apparatus. The two are not opposed positions; what becomes explicit is that they were questions at different levels --- the content of observation, and the precision of its formalization.

In this sense, the aim of this section is not to adopt existing typologies as final classificatory systems. Rather, it is, by redescribing the differences observed by existing typologies as low-dimensional projections of $\Zphase$, $R$, and $\Omega=(\theta,\lambda,q,\zeta,\sigma)$, to convert typological vocabulary into a continuous operating structure that AI can handle.

\subsection{Relation to Big Five and psychometrics}

Psychometric models such as the Big Five are important in that they provide relatively stable trait descriptions at the population level. MIM, on the other hand, does not measure traits but describes which estimation targets, state representations, prediction errors, and action candidates are raised in a given context. The two therefore do not compete. Trait models are strong on distributional, reproducible description; MIM describes the generative process of context-dependent inferential differences.

For example, the trait description that a given subject is ``extraverted'' does not mean that the subject always operates in the same inferential phase. Rather, it remains necessary to ask, additionally, which conditioning basis tends to be externalized in a particular context, and which state representations are foregrounded as action possibilities or social responses. MIM handles this question in the vocabulary of $\Zphase$, $R$, and $\Omega$.

\subsection{From typological vocabulary to AI-tractable structure}

MIM converts typological vocabulary not into a set of labels, but into a set of questions. That is, instead of asking ``which type is this person?'' it asks: ``What is being raised as an estimation target for this observation?'' ``Which state representations are being preserved as sufficient statistics?'' ``Which prediction errors are being evaluated as high-cost?'' ``In which direction is the foregrounding field pushing processing?''

Through this conversion, typology is transformed into a structure that AI can handle. AI need not believe in type names. But it can estimate the observations that typological vocabulary has pointed to --- the distribution of estimation targets, the distribution of state representations, error costs, foregrounding directions, and directional maturity. Through this, typological observations are made reusable as continuous operating structures for explainability, dialogue support, state-representation alignment, and team design.

\subsection{From typology to the alignment map --- what to preserve under transformation}

What is important in the connection to typology is that, when one translates the other into one's own vocabulary, something is preserved, something is coarse-grained, and something is lost. The alignment map $\Phi$ is a local map that translates the state representation of a subject into a form processable within another subject's world model. However, in general there is no mapping that preserves all features. Therefore, alignment is not identification; it is a design problem of choosing which estimation targets, state representations, and error costs to preserve under transformation.

\section[The Alignment Map Phi and World-Model Alignment]{The Alignment Map $\Phi$ and World-Model Alignment}

The formal apparatus developed in Sections~3--6 now enters its inter-subject form. Once recognition is understood as the construction of estimation targets $Y$, state representations $T$, candidate sufficient-statistic spaces $\mathcal C$, phase-formation spaces $\Zphase$, and operating profiles $\theta$, the next problem is not whether subjects agree, but whether a state representation formed in one subject can become processable in another. The alignment map $\Phi$ is introduced for this purpose.

The alignment map $\Phi$ is not a map that guarantees agreement. A subject can, without agreeing with another subject's conclusion, reconstruct the other subject's state representation into a form processable within its own world model. Conversely, even when subjects appear to agree on the same words, processability may not hold if those words connect to different estimation targets or prediction errors. Therefore, the first goal of world-model alignment is not agreement, but processability. Further details are provided in Appendix B.2.

When a subject voices an institutional incoherence, the problem for that subject is a prediction error in the structural domain. However, if the receiver is foregrounding the existential domain, that appeal may be processed as ``an attack on my responsibility or sincerity.'' A third party may further treat it as merely a matter of phrasing. What fails here is not mere translation. The sender's state representation has not been mapped, under the receiver's inferential phase, operating profile, and representation-formation rules, into a processable sufficient statistic.

The alignment map $\Phi$ defined in this section is a local operation for handling this kind of failure. $\Phi$ is not the central principle of the Multi-Phase Inference Mechanism itself. The central principle lies in the fact that subjects constitute the world under different estimation targets, state representations, prediction errors, and operating profiles. $\Phi$ is the operation that, in order to handle this multi-phase character across subjects, maps a state representation into a form processable in another subject or another phase.

What is important is that the goal of $\Phi$ is not ``to make meaning the same.'' Understanding is not the sender and receiver having identical representations. It is that, within the receiver's world model, the sender's state representation is processed without breakdown and, where needed, can be entered into explorative update. Therefore, the alignment map is not a map that erases differences between subjects; it is a map that constitutes processability while preserving difference.

This point also applies to the alignment of meaning and value. $\Phi$ is not a map that transplants one subject's meaning or value directly into another. It is the operation that maps the generative conditions of an utterance or event --- which estimation target it connects to within the sender's world model, which state representations and prediction errors it generates, which error costs it carries --- into a sufficient statistic processable within the receiver's world model. State-representation alignment is therefore not the identification of meaning and value; it is a local alignment that makes their generative conditions mutually tractable.

Building on the formal apparatus of the Multi-Phase Inference Mechanism constructed in Sections~3--6 --- estimation targets, state representations, candidate sufficient-statistic spaces, phase-formation spaces, and operating profiles --- this section presents the formal map \textbf{alignment map} $\Phi$, which is responsible for \emph{aligning state representations across different subjects and phases}. $\Phi$ is the reference apparatus by which the philosophical traditions, cognitive typology, social fragmentation, and AI alignment treated in the latter half of this paper are described in a common vocabulary. The incommensurability and typological differences treated in Sections 7 and 8 can be re-read as the problem of which state representations are alignable at which layer and where they are severed. In Section 10, the full operation, partial operation, and severance of $\Phi$ are treated as problems of social fragmentation and distributed error detection; in Section 11, this is recast as the design problem of AI alignment.

The processability referred to in this paper is the following: a state representation $T^\alpha$ formed by a subject $\alpha$ can, within the world model of another subject $\beta$, be reconstructed --- rather than discarded as a mere rejection response or fatal prediction error --- into some estimation target $Y^\beta$, state representation $T^\beta$, and updatable prediction error $L^\beta$. The alignment map $\Phi$ is not a universal cooperation operator that generates this processability; it is a local map that translates a state representation into the receiver's processable region. Value alignment, world-model alignment, and the mitigation of social fragmentation are achieved not by $\Phi$ alone, but through the estimation of operating profiles, the evaluation of error costs, institutional constraints, social context, and chains of multiple $\Phi$s.

\subsection[Definition of the alignment map Phi]{Definition of the alignment map $\Phi$}

In this paper, the operation that translates a subject's state representation into a form processable within another subject's world model is called \textbf{state-representation alignment}. The formal map that realizes this operation is written as the \textbf{alignment map} $\Phi$. Further, the broader design principle by which individual alignment maps combine to enable world-model collaboration among subjects is called \textbf{world-model alignment}. This section formally defines the alignment map $\Phi$ and also organizes the three-layer hierarchy of world-model alignment (value, state representation, world model).

We write $\Phi$ for the map that aligns a state representation or prediction error arising in phase $K$ into a form processable in another subject or another phase. Let $T_{K,t}^{\alpha}$ be the state representation that sender $\alpha$ holds in the coarse-grained domain $K$. Let $\tilde T_{K',t}^{\beta}$ denote the state representation aligned to be processable for receiver $\beta$ in the coarse-grained domain $K'$. Then,
\begin{equation}
\tilde{T}_{K',t}^{\beta}
=
\Phi_{\alpha\to\beta}^{K\to K'}
\left(
T_{K,t}^{\alpha}
\mid
\theta_t^{\beta},\zeta_t^{\beta}
\right).
\end{equation}
Here, $\theta_t^{\beta}$ is the receiver's prior operating profile, and $\zeta_t^{\beta}=(\rho_t^{\beta},\tau_t^{\beta},\kappa_t^{\beta})$ is the receiver's representation-formation/processing profile.

State-representation alignment is not the identification of state representations across subjects. It is the operation that aligns sender $\alpha$'s state representation $T_K^\alpha$ into a sufficient statistic $\tilde{T}_{K'}^\beta$ that is processable within receiver $\beta$'s world model. $\Phi$ formally acts on the state representation $T$, but its purpose is collaboration between the world models $W^\alpha, W^\beta$. In this paper, the broader goal that the collection of $\Phi$s aims to achieve is called world-model alignment, and individual operations of $\Phi$ are called state-representation alignment, organized hierarchically. Details are developed in this section.

What this equation shows is that what to say and how to say it for communication to succeed depends not only on the content itself, but also on which phases the receiver tends to foreground, with which extractors the receiver carves up the world, what level of directional maturity the receiver has, and at which threshold the receiver enters reflective processing.

However, the goal of $\Phi$ is not merely to translate state representations. More importantly, $\Phi$ enables the \emph{meta-recognition} that subjects constitute the world under mutually different inferential phases, estimation targets, and operating profiles. State-representation alignment is the local operation that supports this meta-recognition; world-model alignment is the design principle that, through the organization of such local operations, ``constitutes processability while preserving difference.''

Alignment has two levels. First, intra-subject alignment. This is the process within a subject by which prediction errors or state representations arising in one phase are aligned into a form processable in another phase. For example, bringing a structurally understood policy down into empirical action, or aligning an existential sense of unease with a structural problem setting. Second, inter-subject alignment. This is the process by which, when one subject's output becomes another subject's observation sequence, that output is aligned so that it becomes processable in the receiver's estimation-target region.

\bigskip
\begin{quote}
\emph{Cooperation is not changing the other. It is aligning the prediction error that arose in one's own phase into a form processable in the other's world model.}
\end{quote}
\bigskip

\subsection[Three-layer structure of Phi]{Three-layer structure of $\Phi$}

In Sections 3 to 5, the Multi-Phase Inference Mechanism and the inference profile were formalized under the vocabulary of statistics (sufficient statistics \cite{lehmann1998}, rate--distortion theory, variational free energy, exponential statistical families \cite{wainwright2008}). Developing each layer of $\Phi$ statistically with the same precision is a research program that exceeds the scope of this paper. Here, for connection to subsequent research, two hierarchies are distinguished. First, as a hierarchy of implementation targets, we distinguish the state-representation layer $\Phi_T$, the world-model layer $\Phi_W$, and the value/preference layer $\Phi_U$. Second, as a hierarchy of purposes, value alignment is positioned as a restricted region, state-representation alignment as the formal operation, and world-model alignment as the higher-level goal.

The following $\Phi_T, \Phi_W, \Phi_U$ are obtained by decomposing the inter-phase, inter-subject alignment map $\Phi_{\alpha\to\beta}^{K\to K'}$ defined in the immediately preceding section into three implementation levels: state representation, generative model, and value/preference distribution. The three layers do not act independently; as described below, they form a constraint structure $\Phi_U \to \Phi_T \to \Phi_W$.

\textbf{First layer --- state-representation alignment} $\Phi_T$: $T_K^\alpha \to \tilde{T}_{K'}^\beta$ is the map that aligns the sender's sufficient statistic $T_K^\alpha$ into a sufficient statistic $\tilde{T}_{K'}^\beta$ that is processable within the receiver's world model. When sufficient statistics are treated as natural-parameter representations of an exponential statistical family \cite{wainwright2008}, $\Phi_T$ can be handled as a \emph{map on the natural parameter space} within the framework of information geometry \cite{amari2016}. Information-conservation conditions (information injectivity) and the optimal transformation under rate--distortion constraints are objects for subsequent technical research.
\begin{equation}
\Phi_T:\ \nu_K^\alpha \;\mapsto\; \tilde\nu_{K'}^\beta\ \mid\ \theta^\beta, \zeta^\beta,
\end{equation}
where $\nu_K^\alpha$ is the natural-parameter representation of $T_K^\alpha$ (a different symbol is used to distinguish it from the reflective threshold $\eta_{K,t}^{\alpha}$ in Section 5.2).

\textbf{Second layer --- world-model alignment} $\Phi_W$: $p^\alpha(o, s; W^\alpha) \to p^\beta(o, s; W^\beta)$ is the alignment map between generative models. This may be handled as Bayesian model averaging \cite{hoeting1999}, structural isomorphism of probabilistic models (DAG isomorphism, alignment of causal structures), or alignment of the surrogate posterior in variational inference. In the context of information geometry, it is described as a projection on the manifold of generative models.
\begin{equation}
\Phi_W:\ p^\alpha(o, s) \;\mapsto\; p^\beta(o, s).
\end{equation}

\textbf{Third layer --- value alignment} $\Phi_U$: this is the map that aligns the prior preference \cite{parr2022} (preferred outcome distribution) $\pi^\alpha(o)$ of active inference with the other subject's preference distribution $\tilde\pi^\beta(o)$. This is not equivalent to the alignment of utility or reward functions; more generally, it is described as the alignment of the prior preferences that a subject holds as a ``desired outcome distribution.''
\begin{equation}
\Phi_U:\ \pi^\alpha(o) \;\mapsto\; \tilde\pi^\beta(o).
\end{equation}

\textbf{Hierarchical relation among the three layers}: the three layers do not act independently; they form a hierarchical structure that mutually constrains one another. Two kinds of priority must be distinguished. Normatively and motivationally, the value layer $\Phi_U$ may guide what is worth observing, predicting, or protecting. Operationally and constructionally, however, value alignment becomes processable only when some state-representation alignment $\Phi_T$ has already been achieved. State-representation alignment then supports the world-model layer $\Phi_W$, while the reachable range of world-model alignment constrains which state-representation and value alignments can be stabilized. Thus the apparent directions $\Phi_U \to \Phi_T \to \Phi_W$ and $\Phi_T$ before $\Phi_U$ do not contradict one another; they refer to different orders, one normative and one processual. In this paper, this hierarchy is presented conceptually; the precise structure of the statistical maps in each layer, their information-geometric description, and the analysis of composite operations are objects for subsequent research (see the research horizons in Sections 13 and 14).

Under this hierarchical structure, the action patterns of $\Phi$ can be classified as (a) \emph{full operation} --- a state in which alignment holds in all three layers; (b) \emph{partial operation} --- a state in which alignment holds only in some layers (for example, alignment holds in the state-representation layer but the value layer is misaligned); and (c) \emph{severance} --- a state in which alignment fails in all three layers. In Section 10, these three categories are used to describe social fragmentation and distributed error detection; in Section 11, AI alignment is redefined. The incommensurability among philosophical traditions and the relation to cognitive typology become auxiliary lines for re-reading the range discussed in Sections 7 and 8 in the vocabulary of $\Phi$.

\paragraph{Action patterns of $\Phi$ organized through concrete examples}

The three-layer hierarchy of $\Phi$ and its three categories form a shared vocabulary used in subsequent sections. This section presents only representative examples in compressed form, rather than long case analyses. Detailed empirical verification is a task for subsequent research.

\begin{itemize}
\item \textbf{Expert collaboration}: even when a machine-learning engineer and a clinical psychologist address the same ``long-term well-being,'' the former organizes the state representation as quantifiable behavioral data, while the latter organizes it as a relational state of trust and responsibility. In this case, even if the value layer partially agrees, the state-representation layer $\Phi_T$ remains in partial operation.
\item \textbf{Political polarization}: this can be described as the process by which a $\Phi_T$ that initially operated partially in the state-representation layer increases its alignment cost through selective exposure and iterative fixation, and proceeds to severance in the world-model layer $\Phi_W$ and the value layer $\Phi_U$.
\item \textbf{Current AI alignment}: RLHF, DPO, and Constitutional AI can be repositioned mainly as one-directional optimization in the value layer $\Phi_U$. The state-representation layer $\Phi_T$ and the world-model layer $\Phi_W$ are, in many cases, not explicitly addressed.
\end{itemize}

Applications to philosophical incommensurability and to the inter-type relations of cognitive typology are developed in Sections 7 and 8, respectively.

As the above examples show, $\Phi$ is not a single operation. (i) It acts on the three layers of value, state representation, and world model; (ii) it takes the three categories of full operation, partial operation, and severance; and (iii) at the level of subject populations, the collection of individual $\Phi$s forms a $\Phi$ network. In Section 10, under this shared vocabulary, social fragmentation, culture, and civilization are redescribed as distributed error detection; in Section 11, AI alignment is redefined.

\paragraph{Bridge to Transformation Loss}

The partial operation of $\Phi$ entails a structural cost: when the sender's state representation $T_K^\alpha$ is aligned into the receiver-side processable $\tilde T_{K'}^\beta$, some information must necessarily be lost. Conceptually, this cost can be viewed as the residual between the information about the sender's estimation target preserved by the original state representation and the information preserved after transformation:
\begin{equation}
\Delta I_{\Phi}=I(T_K^{\alpha}; Y_K^{\alpha})-I(\tilde T_{K'}^{\beta}; Y_K^{\alpha}).
\end{equation}
At this point the equation functions only as an orientation marker for the alignment map. Its connection to rate--distortion theory, information geometry, and AI design is developed in the section ``AI that displays transformation loss'' of Section~11, ``Redefining AI Alignment.'' There, transformation loss is recovered not as a ``failure of translation,'' but as a design principle: it is treated as an informational layer that AI, as a mediating apparatus, should present to human society.

\section{Society, Culture, Civilization: Diversity as Distributed Error Detection}

In this section, we extend MIM to a theory of society, culture, and civilization. This is not, however, a reduction of society to a single function. Rather, we read human society as a distributed world-model updating structure, in which different inference subjects detect different prediction errors, and in which those errors have been preserved, translated, and coordinated as institutions, language, education, custom, science, art, and politics.

The aim of this section is not to advocate a particular political position. The multi-phase pluralism introduced here is a constructive scaffold for rendering comparable, under a common vocabulary of estimation targets, state representations, prediction errors, error costs, and alignment maps, the issues that liberalism, traditionalist conservatism, communitarianism, and deliberative democracy have each preserved separately. The question this section poses is therefore not which political position is correct, but rather which errors each position picks up, which it preserves, and which it tends to render invisible.

This is not an external application added after the formal theory. It is the social-scale reading of the devices introduced earlier: $\mathcal C$ specifies which candidate sufficient statistics can be formed, $\theta$ specifies which candidates are foregrounded, $c_{\mathrm{err}}$ specifies which prediction errors become costly, and $\Phi$ specifies whether the resulting state representations can be made mutually processable.

\subsection{Social Fragmentation and Processability}

\paragraph{What is fragmentation?}

Social fragmentation is not difference of opinion as such. It is the process whereby a state representation that was sufficient within one subject's world model is acted upon without being aligned to a sufficient statistic that is processable within another subject's world model, thereby mutually amplifying prediction errors. This rupture appears with particular force in conflicts over meaning and value. The reason is that the same words ``justice'' or ``evil'' circulate as shared public vocabulary while carrying different error costs in different phases.

The action $a_{t+1}^{\alpha}$ of subject $\alpha$ becomes an observation $o_{t+1}^{\beta}$ for subject $\beta$. If that observation raises $\beta$'s prediction error above threshold in the coarse-graining domain $K'$ and the directional compatibility is negative, defense, rejection, anger, or avoidance are likely to follow. When this behavior then enters yet another subject's observation stream and triggers the same processing, fragmentation spreads socially.

Consider a typical scene. Subject $\alpha$, strongly foregrounding the existential domain, posts a short message expressing anger at an event and empathy with those affected. The same post appears to subject $\beta$, who strongly foregrounds the structural domain, as ``emotional argument with sloppy logic.'' As a rebuttal, $\beta$ posts a proposition about institutional-design consistency. To $\alpha$, who then reads it, this appears as ``cold logic that crushes the pain of those involved under numbers.'' Although the two are talking about the same event, their words circulate without passing through $\Phi$, mutually amplifying prediction errors with each observation. SNS functions as an apparatus that diffuses such unaligned prediction errors directly, bypassing the alignment network (education, editors, face-to-face discussion).

\paragraph{Understanding as processability}

The goal of world-model alignment is not for sender and receiver to share the same state representation. The goal is that, after alignment, the state representation becomes processable within the receiver's world model.

Processability can be expressed, for example, by the following condition. For the post-alignment state representation $\tilde T_{K',t}^{\beta}$, if the receiver's prediction error is below threshold, processing does not break down.
\begin{equation}
L_{K',t}^{\beta}(\tilde T_{K',t}^{\beta})\leq \eta_{K',t}^{\beta}.
\end{equation}
And even when the prediction error exceeds threshold, if the directional compatibility is positive, the receiver can enter receptive, exploratory processing rather than vigilance or rejection.
\begin{equation}
L_{K',t}^{\beta}(\tilde T_{K',t}^{\beta})>\eta_{K',t}^{\beta}
\quad\mathrm{and}\quad
\mu_{K',t}^{\beta}(\tilde T_{K',t}^{\beta})>0.
\end{equation}
\bigskip
\begin{quote}
\emph{Understanding is not the sharing of identical representations. It is that, within the receiver's world model, a state representation becomes processable.}
\end{quote}
\bigskip

\paragraph{Arguments with overly large subjects}

When anxiety, anger, or alienation arising in the existential domain are converted into the structural domain without adequate structural state representation, coarse universal propositions are produced. Examples are arguments framed with large subjects such as ``society,'' ``men,'' ``women,'' ``STEM types,'' ``liberals,'' or ``conservatives.''

This is an attempt to convert existential prediction error into structural propositions, but because the required state representations are insufficient, the argument appears to readers in the structural domain as logically coarse. Then the prediction error of subjects who foreground the structural domain rises, and their rebuttal in turn raises the prediction error of subjects who foreground the existential domain.

\subsection{Society as a Network of World-Model Alignment}

Society can be understood as a network that constitutes world-model alignment among different inference phases and different subjects. Education, institutions, ritual, law, professions, media, narrative, translators, and mediators are all devices that align the state representation of one phase or subject into a form processable by another phase or subject.

SNS tends to function as a device that bypasses this alignment network and circulates unaligned prediction errors directly. As a result, unaligned prediction errors are easily spread as citation, rebuttal, mockery, condemnation, or expressions of solidarity~\cite{bail2021,sunstein2017,pariser2011}.

\paragraph{The dissipative-structural role of exploration and stabilization}

The mechanism of exploratory and stabilizing directions introduced in Section~5 is not merely a within-subject psychological tendency; it carries a computational role of mediating the trade-off between short-term and long-term predictability. In the short term, the stabilizing direction compresses the existing distribution of state representations and increases the certainty of prediction. This is indispensable for immediate decision-making and action generation. Stabilization alone, however, drives the distribution of state representations to converge to the existing model and loses the capacity to detect new differences. The exploratory direction temporarily expands the distribution of state representations, incorporating new hypotheses, possibilities, and external information. In the short term this increases prediction error and consumes more cognitive resources, but in the long term it opens new regions of predictability. Exploration is thus not contrary to free-energy minimization; rather, by allowing a short-term increase of free energy, it extends the range over which long-term free-energy minimization is possible.

This within-individual structure also corresponds to a social distributional structure. Within an individual, stabilization-oriented phases maintain low-entropy structured assets (institutions, procedures, formalized knowledge), thereby reducing the risk of exploration in another phase. Within society, new prediction errors detected by exploration-oriented subjects are shared as institutions, technologies, language, concepts, and bodily skills, becoming stabilizing assets for other subjects. That is, the phase-wise allocation of exploration and stabilization is at once an allocation of cognitive resources within individuals and a distributed updating mechanism of the social world model. As a whole, society maintains a dissipative structure of long-term world-model updating by dividing exploratory updating and stabilizing preservation among a population of subjects with diverse prior operating profiles $\{\theta^{\alpha_i}\}$. This is isomorphic to the structure shown by Prigogine's dissipative-structure theory~\cite{prigogine1984}: ``order formation in non-equilibrium systems is maintained by the inflow and dissipation of energy or information.''

From this perspective, diversity of intelligence is both an ethical value and a functional condition for maintaining the generalization performance of the social world model. If all subjects foreground the same phase in the same direction, prediction errors in some domain are shared socially while prediction errors in other domains accumulate undetected. The fact that diverse prior operating profiles are distributed across society broadens the domain of detectable prediction errors and supports the long-term stability of the social world model.

\subsection{Working Definitions of Meaning, Value, Justice, Evil, Culture, and Civilization}

In the vocabulary of MIM, meaning is the connectability of a given observation, sign, event, or utterance: to which estimation target it can connect within a subject's world model, which state representation it can form, and which prediction error it can produce. Value is the allocation of error costs and update priorities: the degree to which a given state representation, prediction error, or update path is treated, within a subject's or group's world model, as something to be preserved, protected, updated, sacrificed, or exchanged.

Justice is the social alignment rule that determines, when the meanings and values of multiple subjects collide, which estimation targets, state representations, and prediction errors are to be protected, which transformations are to be permitted, and which sacrifices are to be justified. Evil is constituted, within the world model of a given community, as that which destroys the meanings, values, and state representations that ought to be preserved. This is not an argument that exonerates evil. It is an argument for objectifying what is constituted as evil, and on the basis of which prediction errors and error costs it is rendered impermissible.

Culture is the scaffolding of meaning and value, collectively shared and environmentalized as language, custom, institution, art, technology, and education. Civilization is a network of collective world models in which such scaffolding is organized at large scale and over long temporal scales. From this perspective, AI alignment cannot be exhausted by fitting AI to human preferences. It must address the conditions under which meaning, value, and justice arise, and the transformations that render them processable.

\subsection{The Necessity Argument for Diversity --- The High-Dimensional World and the Incompleteness of a Single Inference Profile}

The world is high-dimensional. From the side of observation streams, from the side of estimation targets, from the side of state representations, and from the side of error costs, independent variations far exceeding the granularity that any single human can grasp at once are constantly in motion. A single subject extends a foregrounding gradient field $R$ over a limited region of the inference-phase formation space $\Zphase$, processing observation streams under a limited combination of $\theta, \zeta, \sigma$. In phases that are not foregrounded, in directions handled with low maturity, and in domains with high reflective-processing thresholds $\eta$, prediction errors $L_K$ may occur yet remain unconscious, or accumulate without being acted upon. The prediction errors that a single subject can detect at one time are thus a proper subset of all prediction errors the world is generating.

This asymmetry is not a problem of insufficient capacity. It is a structural consequence under the three constraints---computational $C_{\mathrm{comp}}$, observational $C_{\mathrm{obs}}$, and action constraint $C_{\mathrm{act}}$---of the fact that $\Zphase$ is high-dimensional and $\Omega=(\theta,\lambda,q,\zeta,\sigma)$ can only be a low-dimensional projection. Hence, in order to detect prediction errors of the world as a society as a whole, subjects with different $\Omega$ must exist in parallel. Diversity is neither ornament nor surplus to society. It is a structural necessity for the collective world model to maintain its generalization performance.

\paragraph{Connection to Hayek's distributed-knowledge thesis}

This consequence is structurally isomorphic to the distributed-knowledge thesis Hayek developed in ``The Use of Knowledge in Society''~\cite{hayek1945} and ``The Constitution of Liberty''~\cite{hayek1960}. Hayek's knowledge problem---that economically relevant knowledge is scattered in time-, place-, and situation-specific forms and cannot be aggregated by any single subject or central planner---can be translated into MIM's vocabulary as follows. Each economic subject $\alpha_i$ detects, under its coarse-graining domain $K_i$, a prediction error $L_{K_i,t}^{\alpha_i}$ from the observation stream $o_{1:t}^{\alpha_i}$. That error is first extracted as a sufficient statistic $T_{K_i,t}^{\alpha_i}$ only under that subject's prior operating profile $\theta^{\alpha_i}$ and representation-formation-and-processing profile $\zeta^{\alpha_i}$. To transfer it directly to the sufficient-statistic space of another subject would require the existence of $\Phi_{\alpha\to\beta}^{K_i\to K_j}$, and that operation is inevitably accompanied by information loss and error amplification. The central planner is, by definition, an operation that attempts to integrate the prediction errors of all subjects under its single $\Omega$, and this faces the impossibility of compressing the high-dimensional $\Zphase$ into a single low-dimensional projection.

Hayek's argument about the price mechanism can be reread, in the vocabulary of MIM, as the spontaneous formation of a collective $\Phi$ network. Prices are a low-bandwidth channel that aligns each subject's local prediction error and error cost into a sufficient statistic that is processable within other subjects' world models. The price itself is a coarse projection that loses most of the original state representation, yet it carries enough information to establish processability $L_{K',t}^{\beta}(\tilde T_{K',t}^{\beta}) \le \eta_{K',t}^{\beta}$. What Hayek called ``coordination unachievable by central planning'' is, in this paper's framework, ``world-model alignment that is established by the parallel operation of many local $\Phi$ without being compressed into a single $\Omega$.'' The present paper does not restrict this argument to the market mechanism. Language, custom, law, ritual, scholarly communities, and SNS are each positioned, however imperfectly, as devices that can constitute $\Phi$ networks.

\subsection{The Need Argument for Diversity --- Environmental Change and the Ecology of Error-Detecting Subjects}

The necessity argument addresses the fact that diversity arises. The need argument addresses what happens when diversity is lost. The two are different claims.

The environment changes. The generative distribution $p(o_{1:t})$ of observation streams follows a long-term non-stationary trajectory under ecosystems, climate, technology, infectious disease, demographic structure, energy use, language contact, and external political shocks. The distribution of prior operating profiles $\{\theta^{\alpha_i}\}$ that was socially sufficient at one moment will, at another moment, leave prediction errors in certain domains undetected. For example, a society dominated by stabilization-oriented subjects optimized for institutional design in a stable period relatively lacks exploration-oriented subjects to detect new prediction errors when ecosystemic change or technological disruption occurs. Conversely, a society dominated only by exploration-oriented subjects cannot accumulate the stabilizing assets that carry long-term predictability. A structure isomorphic to the way species diversity in an ecosystem sustains resilience to disturbance also holds for the distribution of inference profiles.

Translated into evolutionary terms, this can be put as follows. Long-term updating of the social world model depends on a balance between two directions: the preservation of error-detection capacity and the generation of new error-detection capacity. The former is preserved through tradition, craft, ritual, professions, and generational transmission. The latter is generated through migration, intercultural contact, generational turnover, education, the adoption of new technologies, voices from the periphery, and the experiences of minorities. When the ratio tilts to one side, prediction errors in certain domains accumulate undetected by society and, once a threshold is crossed, manifest themselves destructively. This is the same structure as the vulnerability of monoculture in an ecosystem, or the synchronized collapse of homogeneous strategies in financial markets.

Diversity is therefore not first of all a virtue or an ethical demand; it is a functional requirement that is ecological and evolutionary. For a society to retain self-correcting capacity in the face of a changing environment, the domain of detectable prediction errors must extend broadly enough to anticipate the direction of environmental change. Who will detect which error first cannot be decided in advance. Those who detect an error first are not always members of the majority. This is the core of the need argument: ``against unpredictable environmental change, preserve the diversity of error-detecting subjects.''

This perspective connects to the research tradition of epistemic democracy. Landemore's~\cite{landemore2013} theory of ``democratic reason'' redescribed the legitimacy of democracy not as the aggregation of preferences by majority vote, but as the collective utilization of cognitive capacities distributed throughout society. Ober~\cite{ober2008} showed, through studies of ancient Athenian democracy, historical cases in which distributed knowledge and diverse error detection constituted a collective capacity for problem-solving. Sunstein~\cite{sunstein2006} empirically organized the conditions and limits of collective intelligence. In the present framework, these are rearranged as attempts to extract, through different empirical, historical, and normative entry points, the same isomorphic structure: ``democracy = a distributed error-detection mechanism of society.'' The need argument for diversity rebases this tradition of epistemic democracy upon the constructive fact of differences of state representation among subjects.

\subsection{The Defense Argument for Diversity --- Functional Defense and Defense from Intrinsic Value}

Both the necessity argument and the need argument are functional arguments. When dealing with prejudice, discrimination, and exclusion, another kind of argument must be raised. In this section, we deliberately distinguish these two strands. The dignity of subjects who suffer discrimination must not be reduced to function.

The defense argument in this paper partially overlaps with the value pluralism Berlin~\cite{berlin1969} formulated---the position that multiple values cannot be fully reduced to one another and that priorities must be continually reconstituted according to context. The present paper differs from Berlin in two respects, however. First, the present paper treats not value itself but the state representations and error costs out of which value arises. Second, whereas value pluralism is a normative demand ``to respect multiple values equally,'' the present paper addresses the constructive demand of ``design conditions under which multiple state representations become processable.'' The two are not opposed; the latter is positioned as a vocabulary that translates what the former demands into an implementable form.

\paragraph{Functional defense --- discrimination as the production of an undetectable domain}

In MIM's vocabulary, prejudice, discrimination, and exclusion are ``operations that structurally exclude prediction errors detected by a given subject-group from the collective processing of the society as a whole.'' The discriminated subject processes the observation stream $o_{1:t}$ within their coarse-graining domain $K$, forms a state representation $T_K$, and detects a prediction error $L_K$. That prediction error, however, is placed in a state where $\Phi$ does not operate at the entry of the social alignment network (education, media, professions, law, political representation). Because the alignment map is refused from the outset, the state representation is acted upon without being translated into a processable sufficient statistic. The consequence is that society as a whole becomes structurally unable to detect prediction errors of a particular domain. Discrimination is at once a damage to those who are discriminated against and a damage to the self-correcting capacity of the world model of those who discriminate.

This functional argument has the power to explain why discrimination also generates costs for society as a whole. It cannot, however, serve as the sole ground for criticizing discrimination. To do so would theoretically permit the consequence that subjects deemed functionally less useful need not be defended.

\paragraph{Defense from intrinsic value --- identity as the historical condition of error sensitivity}

The dignity of discriminated subjects precedes any functional-reductive argument. The present paper does not treat identity as a fixed group essence. At the same time, categories such as sex, race, class, region, religion, disability, occupation, and generation decisively pertain to historical processes of formation: which subject repeatedly experiences which prediction errors, which state representations they come to form, and which error costs they come to evaluate highly. Even for the same observation stream, who, by virtue of belonging to a particular category, finds their prediction error rising above $\eta$ and becoming conscious, and whose prediction error remains undelivered to reflective processing and accumulates, differs systematically. Identity is treated not as essence but as the historical condition for the formation of error sensitivity.

From this perspective, defense from intrinsic value can be articulated as follows. Each subject has an error sensitivity proper to its historical process of formation. That error sensitivity cannot be substituted for by another subject. Non-substitutability holds independently of whether it is functionally useful. Even if a given subject's error sensitivity is not currently detecting socially useful prediction errors, the state of not treating that subject as a being who possesses a unique error sensitivity is described, in the framework of the present paper, as an operation that, in advance, excludes a subject from the processability network of society as a whole. This is, independently of functional loss, an operation that breaks the basic premise of world-model alignment---that each subject is treated as a being who possesses a unique $\Omega$.

Functional defense and defense from intrinsic value are compatible. Functional defense speaks to society's capacity for self-correction; defense from intrinsic value speaks to the very existence of a single subject with its own error sensitivity. The present paper keeps these two strands separate and does not reduce the latter to the former.

\subsection{Identity and Error Sensitivity}

This paper does not treat identity as a fixed group essence. However, social positions such as sex, race, class, region, religion, disability, occupation, and generation influence which prediction errors a subject repeatedly experiences, which state representations they form, and which error costs they evaluate highly. Identity is therefore treated not as essence but as the historical condition for the formation of error sensitivity. Taylor's theory of recognition showed that self-understanding depends on social recognition and narrative formation~\cite{taylor1989,taylor1992}. In MIM, this point is redescribed as the problem that the state representations and error sensitivities of subjects are formed within social relations.

This stance does not reduce individuals to groups. At the same time, it does not look only at individual differences and ignore historically formed error sensitivities. Identity is the condition of formation that enables one to understand what a subject has experienced as errors that cannot be overlooked; it is not a final essence.

\subsection{The Design Argument for Diversity --- The Simultaneous Demand of Preservation and Suppression}

Diversity is not something to be celebrated unconditionally. The parallel existence of diverse error-detecting subjects is the precondition of self-correcting capacity, but at the same time, when their prediction errors circulate without passing through $\Phi$, mutual prediction errors are destructively amplified. The social fragmentation described in Sections~10.1--10.2 is not a problem of diversity itself, but a problem of diversity lacking a processability network.

The task of the design argument is therefore to design institutions, cultures, AI, and governance structures that simultaneously preserve diverse error detection and suppress destructive error amplification. The two stand in tension. Operations that try to maximize error detection often permit local polarization. Operations that try to suppress error amplification often dull error-detection capacity itself. The three-layer hierarchy and three typologies of $\Phi$ discussed in Sections~10.1--10.3 are a common vocabulary for handling this tension.

Concretely, four design targets become issues simultaneously. First, preservation devices that do not erase the unique $\Omega$ of each subject or group (education, professions, mother tongue, custom, regional culture, minority representation). Second, translation devices that organize $\Phi$ among different $\Omega$ (editors, mediators, translators, venues of face-to-face discussion, modes of inter-professional collaboration). Third, circulation-control devices that do not directly diffuse mis-aligned prediction errors (SNS algorithmic design, media ethics, AI response design, the conduct of public discussion). Fourth, reversibility-guarantee devices that ensure destructive error amplification does not acquire institutional irreversibility (courts, retrials, elections, correction procedures, correctable records). The research horizon toward the collective scale to be opened in Section~14 is positioned as the task of systematically developing these four design targets in MIM's vocabulary.

\subsection{Tradition and Reform --- Preserved Errors and Newly Picked-Up Errors}

The argument for diversity concerns not only synchronic parallelism but also diachronic preservation and updating. In this section, we distinguish traditionalist conservatism and reformist thought by the character of the prediction errors each addresses.

Traditionalist conservatism was positioned in Section~7.5 as the preservation of long-term error sensitivity. Observations repeated across generations, prediction errors that first become manifest only on long time scales, and error costs that short-term optimization does not detect---the structures of $\zeta$ and $\sigma$ that carry these are sedimented in ritual, custom, transmission, craft, religious practice, and family structures. To compress these as ``low rationality'' or ``inefficient'' is an operation that loses from society those error sensitivities that can only be detected on long time scales.

Reformist thought is a movement that seeks to bring newly detected prediction errors---errors that until then accumulated below $\eta$ or were detected only by particular peripheral subjects---into the social processability network. Articulated in this paper's framework, reformist thought becomes ``resistance to society's structurally continuing to ignore newly detected prediction errors.'' Reformist thought is an active demand for state-representation alignment, seeking to connect detected errors into society's $\Phi$ network as a whole.

The two appear opposed but are the obverse and reverse of the same structure. Both demand that ``society not leave undetected the prediction errors it ought to detect.'' What differs is the time scale and the subject distribution of the error sensitivities to be preserved and the errors to be newly picked up. To compress both into a single progressivist or single conservative philosophy of history is an operation that runs against the central thesis of this paper.

\paragraph{An MIM extension of Rawls's overlapping consensus}

The veil of ignorance and the priority of basic liberties presented in Rawls's theory of justice~\cite{rawls1971} can be reread as conditions that do not structurally erase the prediction errors of any subject. Furthermore, the concept of overlapping consensus that Rawls developed in ``Political Liberalism''~\cite{rawls1993} addresses how plural comprehensive doctrines can converge, from mutually different grounds, upon the same principles of political justice. Rawls's framework finally aims at convergence upon the same normative values. The framework of MIM extends this one step further: what should converge is not ``the same value'' but the ``processability of different state representations.''

In MIM's vocabulary, overlapping consensus can be redescribed as follows. It is the state in which an alignment map $\Phi$ holds in some shared domain among multiple subjects or groups $\{\alpha_i\}$ who maintain different prior operating profiles $\{\theta^{\alpha_i}\}$ and different value-related evaluations $\{U^{\alpha_i\}}$. Each subject endorses $\Phi$ for reasons proper to its own world model; what is shared is the operability of the alignment map, not the state representations themselves. This is an operation that rereads Rawls's ``overlapping consensus'' not as a limited alignment of the value layer $\Phi_U$, but as a networked alignment of the state-representation layer $\Phi_T$.

\paragraph{MacIntyre's vital tradition and moribund tradition}

The distinction between vital and moribund traditions that MacIntyre introduced in ``After Virtue''~\cite{macintyre1981} can be articulated in this paper's framework as follows. A vital tradition is one that retains the capacity, within itself, to incorporate new prediction errors continually and to reconfigure its own state representations. It keeps open an internal path for updating its own $\Omega$ in response to changes in external observation streams. A moribund tradition is one in which that path has closed. New prediction errors arrive neither from outside nor from inside; the state representation becomes fixed, and $\Phi$ operates only formally. In MIM's vocabulary, a vital tradition is described as ``an organization of long-term error sensitivity that maintains the balance of exploration and stabilization directions,'' and a moribund tradition as ``an organization of error sensitivity in which the stabilizing direction has become excessive and the exploratory direction has been institutionally excluded.'' Tradition itself is neither conservative nor progressive. What divides the two is whether the updating path preserved within the tradition is open or closed.

\paragraph{Taylor's embedded narrative}

The concept of embedded narrative that Charles Taylor developed in ``Sources of the Self''~\cite{taylor1989}---that the subject does not understand itself atemporally but is embedded within the historical narrative into which it has been born, forming itself by updating that narrative---can be connected to this paper's vocabulary as follows. Each subject's $\Omega^{\alpha}=(\theta^{\alpha},\lambda^{\alpha},q^{\alpha},\zeta^{\alpha},\sigma^{\alpha})$ is not an object of design from nothing, but a history-dependent structure formed by the historical observation streams of the cultural, linguistic, generational, and professional communities to which it belongs. The articulation in this paper of identity as ``the historical condition for the formation of error sensitivity'' can be read as a structural isomorph of Taylor's embedded narrative. The conflict between reform and conservatism is a dispute over how to update this embedded narrative; it is not an object that can be evaluated from an ahistorical vantage free of any narrative.

\subsection{Culture and Civilization as Dissipative Structures of Error Processing}

In the framework of this paper, culture and civilization are described not as static ``value systems'' or ``normative systems,'' but as dynamic structures of world-model updating. Culture is the totality of the scaffolding of meaning and value in a given group (language, custom, institution, art, technology, education); civilization is a network of collective world models in which such scaffolding has been organized at large scale and over long time scales. This is consistent with the working definitions of the previous subsection. The present subsection gives those working definitions a description as dynamic structures.

The concept of dissipative structure that Prigogine developed in ``Order Out of Chaos''~\cite{prigogine1984} describes the formation of order in non-equilibrium open systems as dynamic stability under the inflow and dissipation of energy, matter, and information. In a closed equilibrium system, order collapses according to the second law of thermodynamics. In an open non-equilibrium system, energy and information flowing in from outside are organized internally and dissipated outward, so that apparent order is dynamically maintained. The present paper extends this structure to the world-model updating of culture and civilization.

Culture and civilization are non-equilibrium error-processing structures that organize, as reconfigurations of internal state representations, institutions, language, practices, and narratives, the inflow of prediction errors from external observation streams (environmental change, technological progress, contact with other cultures, generational turnover from within), and that dissipate, as history, memory, and relics, the older structures produced in the process of reconfiguration. For a culture to be ``alive'' means that this cycle of inflow and dissipation is operating. For a culture to be ``dead'' means either that inflow has stopped, or that dissipation can no longer occur and old structures have piled up internally without moving. MacIntyre's distinction between vital and moribund traditions can be reread as two typologies of the operating state of this dissipative structure.

Extended to the civilizational scale, civilization is a large-scale dissipative structure in which multiple cultural dissipative structures are mutually interconnected by a $\Phi$ network. The collapse of a civilization occurs when an abrupt shift in external observation streams (climate change, infectious disease, technological disruption) causes inflow of errors to exceed threshold simultaneously with an insufficiency of the internal $\Phi$ network's capacity for reconfiguration. Conversely, the long-term persistence of civilization depends on the headroom of internal dissipative capacity relative to the scale of error inflow, and on the preservation of the $\Phi$ network's capacity for reconfiguration. This is not a deterministic law of history; it is a constructive framework for analyzing civilization as an error-processing structure.

The argument of this subsection integrates the necessity, need, defense, and design arguments of the first half of the section, together with the discussion of tradition and reform, at the longest time-scale layer. The parallel existence of diverse error-detecting subjects, the organization of processability by a $\Phi$ network, the preservation of long-term error sensitivity by tradition, and the bringing-in of new errors by reform---all of these are mutually inseparable operating components for non-equilibrium maintenance of culture and civilization as large-scale dissipative structures. The research horizon toward the collective scale to be opened in Section~14 is positioned as the work of developing this dissipative structure formally in MIM's vocabulary.

\bigskip
\begin{quote}
\emph{Culture is not a stationary norm but a dynamic equilibrium between external error inflow and internal reconfiguration. Civilization is the large-scale behavior of a network in which such equilibria are mutually interconnected.}
\end{quote}
\bigskip

\subsection{The Political Cost of Correct Inference: From Error Exclusion to Governance}
\label{subsec:political-cost-correct-inference}

The preceding account of culture, civilization, tradition, and reform leads to the central political consequence of the framework. A locally correct inference is never neutral with respect to error selection. To infer correctly is to select an estimation target, stabilize a state representation, detect certain deviations as prediction errors, and assign costs to those errors. This operation is indispensable for cognition. Without such selection, no observation could become cognitively or inferentially relevant.

Yet the same operation necessarily excludes other possible errors from the space of relevance. What appears as rational consistency within one inferential structure may therefore become blindness at another scale. A correct inference can reduce local uncertainty while suppressing error signals that are socially, bodily, relationally, culturally, or institutionally significant. The excluded errors are not merely noise. They may be errors that another subject, another institution, another tradition, another generation, or another embodied position is structurally better able to detect.

This is the political cost of correct inference. Political conflict does not arise only from disagreement over values or preferences. It also arises from disagreement over which excluded errors matter. Liberal, conservative, technocratic, communal, scientific, religious, bodily, or existential forms of reasoning may each detect real errors while systematically discarding others. Each can be locally correct within its own estimation target, state representation, prediction-error structure, and error-cost ordering; each can also become socially destructive when it is elevated into the only legitimate inference regime.

This point also clarifies why the problem cannot be solved by simply increasing observation, transparency, or surveillance. The difficulty is not merely that finite subjects lack enough data. Even extensive observation does not uniquely determine which estimation targets should be formed, which state representations should be protected, which error costs should be assigned, or which endophenomenal and embodied errors should remain beyond compulsory externalization. If SIA is interpreted as a mandate to collect enough observations to force convergence, it becomes a justification for a society in which dignity, privacy, bodily interiority, conscience, and the historically formed plurality of error sensitivities are sacrificed to a single externalizable regime of correctness. MIM rejects this inference. More observation does not automatically produce better alignment; it may instead erase the very conditions under which heterogeneous error-detection capacities survive.

From this perspective, governance is not merely the aggregation of preferences or the selection of a correct objective function. It is the institutional problem of preserving, comparing, and coordinating heterogeneous error-detection capacities without forcing them into a single inference regime. Rights, privacy, dissent, minority protection, professional autonomy, local knowledge, judicial review, scientific disagreement, and political deliberation can all be reread as devices for preventing one locally correct inference structure from monopolizing the definition of socially relevant error.

This is also where the problem of AI governance begins. If correct inference always excludes some errors, then the goal of AI governance cannot be to install a single correct inference regime over society, even when that regime is statistically powerful, logically consistent, or computationally efficient. It must instead preserve and compare heterogeneous error-detection capacities while making explicit which errors are lost when one representation is transformed into another. The alignment problem, therefore, is not an external application added after the theory of social fragmentation. It follows directly from the political cost of correct inference.

The point is not to deny correct inference. It is to make visible what correct inference must leave out, and to treat that exclusion itself as a design object for political philosophy, AI governance, and world-model alignment.

\section{Redefining AI Alignment}

Much of AI alignment begins by asking what AI should optimize. This paper asks a prior question: which human errors become visible to the alignment process, and which are silently removed before optimization even begins? If an alignment system can learn only from what humans can already externalize, it may become excellent at optimizing the visible layer of human preference while erasing the errors that remain difficult to say, audit, or translate.

The preceding subsection established the bridge from social fragmentation to AI alignment. If every locally correct inference excludes some errors, then alignment cannot mean installing a single correct inference regime over society, even when that regime is statistically powerful, logically consistent, or computationally efficient. It must mean designing systems that preserve, compare, and mediate heterogeneous error-detection capacities while making explicit which errors are lost when one state representation is transformed into another.

This section makes explicit the assumptions that current AI alignment research has implicitly carried, and discusses how they are to be redefined under the framework of this paper. The transition is direct: once governance is understood as the preservation and coordination of heterogeneous error-detection capacities, AI alignment becomes the question of how AI systems should mediate those capacities without collapsing them into a single objective function. Human society has maintained its self-correcting capacity not by compressing all prediction errors into a single authority, a single value function, or a single philosophy of progress, but by distributively preserving, translating, and coordinating the errors detected by different inferential subjects. AI alignment is therefore redefined here as the design problem of extending this self-correcting capacity without allowing AI to dominate it.

The subsections below proceed in five steps. First, they identify the limits of existing alignment methods when estimation targets, state representations, prediction errors, and error costs are not shared. Second, they give priority to state-representation alignment as the condition under which value alignment can even become meaningful. Third, they position AI as a new error-detecting subject that can either mediate or dominate human error detection. Fourth, they define the design requirements of displaying transformation loss, presenting multiple error-selection options, and supporting self-audit. Finally, they summarize alignment as the extension of social self-correction rather than the installation of a single objective function.

\subsection{Limits of Existing AI Alignment}

RLHF~\cite{ouyang2022}, DPO~\cite{rafailov2023}, Constitutional AI~\cite{bai2022}, and preference alignment in general are important technical advances. They have pushed forward, to a level of practical usability for large language models, the one-way optimization that aligns AI outputs to human preferences and safety standards. These frameworks, however, do not possess any apparatus for addressing the distinction between estimation target $Y$ and state representation $T$. Preference data is a mixture of relative evaluations given by a subject to a given state representation under a given error cost; it does not touch the phase-wise prediction-error structure behind it.

Consider a concrete example. Given the same bioethical question, a user may, at one moment, ask from the existential domain ``what if my own family were in the same situation,'' at another moment ask from the structural domain ``what is institutionally consistent,'' and at yet another moment ask from the ideational domain ``what conception of the human is presupposed here.'' These three questions raise different $Y_K$, extract different $T_K$, and generate different $L_K$. Current AI, however, treats the user as a subject with a single preference vector, and smooths the differences among these three phases as ``contextual fluctuation'' or ``inconsistency.'' The phenomenon that uniform responses are generated for users who hold multiple world models is a consequence of this smoothing.

A similar problem arises at the collective scale. The distribution of preferences across a user population regarding a given policy is the mixture of results in which subject populations with different $\theta^\alpha$ have raised different $Y^\alpha$. To aggregate this into a single reward is tantamount to deciding implicitly which prediction errors in which phases to take up and which to discard. The decision of aggregation is embedded in technical specifications and is operated invisibly both to users and to developers. This is not the philosophical proposition that ``values are not single.'' It is the structural limit that ``the aggregation procedure of value does not preserve the distinction between estimation target and state representation.''

Existing alignment lacks a vocabulary to point to this limit from the inside. Optimization of value functions does not ask from which state representations value itself arises. Under the framework of this paper, existing AI alignment is repositioned in a limited way as ``one-way optimization in the value layer $\Phi_U$,'' and it is made explicit that the problems in the upstream state-representation layer $\Phi_T$ and the world-model layer $\Phi_W$ remain untouched.

\subsection{Beginning from State-Representation Alignment}

Value alignment stands upon state-representation alignment. This is a matter of operational order, not of philosophical superiority. Values may normatively guide what should be protected, but a value judgment can be compared, contested, or acted upon only after the relevant state representation has become processable. To evaluate an AI's output as ``desirable'' or ``undesirable,'' it must first be processable: which estimation target the output raises, which state representation it presents, which prediction error it surfaces, and which error cost it carries. If value evaluation is given without this processability being established, the evaluation sticks to the surface of the output and does not reach the depths of the state representation.

The alignment map $\Phi_{\alpha\to\beta}^{K\to K'}$ defined in Section~9 is the apparatus that formally guarantees this order of priority. It aligns the sender $\alpha$'s state representation $T_{K,t}^{\alpha}$ to a $\tilde T_{K',t}^{\beta}$ that is processable under the receiver $\beta$'s operating profile $\theta_t^{\beta}$ and representation-formation-and-processing profile $\zeta_t^{\beta}$. The value layer $\Phi_U$ becomes meaningful only on top of the state in which this $\Phi_T$ has been achieved. To rush $\Phi_U$ where $\Phi_T$ is ruptured causes value alignment to regress into a forced labeling of state representations that the receiver cannot process.

The first task of AI alignment is therefore not the optimization of value, but the construction of the alignability of state representations. That an AI can estimate, with respect to its own output, which inference phase it is foregrounding, which state representation it is extracting, and which prediction error it is surfacing, and can display these according to the user's operating profile---this is the minimal requirement of state-representation alignment. Optimization of the value layer is overlaid on top of the establishment of this minimal requirement, in accordance with particular applied contexts.

This ordering is not a denial of existing alignment. RLHF, DPO, and Constitutional AI all operate effectively in particular applied contexts where state-representation alignment is already tacitly established. The problem is not those methods themselves, but the operational structure that applies value-layer optimization alone even in situations where state-representation alignment is not established. The directive ``begin from state-representation alignment'' is to be read as a design principle that restructures this operational pattern from within.

\subsection{AI as a New Error-Detecting Subject}

AI is not merely a device that reproduces human language. AI that has undergone large-scale pretraining and high-dimensional representation learning is a subject capable of detecting prediction errors at scales, granularities, and combinations not observable by any single human. An AI that extracts, from millions of medical records, symptom patterns invisible to any single clinician; an AI that detects, from decades of economic time series, structural changes invisible to any single economist; an AI that discovers, from hundreds of millions of lines of code, vulnerability chains invisible to any single developer---all of these are positioned as inference subjects of a different kind, with $\Omega=(\theta,\lambda,q,\zeta,\sigma)$ different from that of humans.

In the framework of this paper, AI can be described as follows. The prior operating profile $\theta^{\mathrm{AI}}$ has foregrounding tendencies prescribed by the training data distribution, the loss function, and architectural inductive biases. The representation-formation-and-processing profile $\zeta^{\mathrm{AI}}=(\rho^{\mathrm{AI}},\tau^{\mathrm{AI}},\kappa^{\mathrm{AI}})$ is characterized by distributional representations in high-dimensional continuous spaces and by multi-stage compression and reconstruction via attention mechanisms. The directional maturity $\sigma^{\mathrm{AI}}$ exhibits strong unevenness: in some domains it surpasses human experts, while in others it does not reach human infants.

What matters is that AI is not an extension of human cognition but another inference subject with a different $\Omega$. Among the prediction errors AI detects are not only high-speed and large-scale versions of those humans detect, but also errors that, under human allocations of cognitive resources, cannot in principle be foregrounded. Conversely, many of the bodily and existential prediction errors that humans detect on a daily basis do not appear on AI's input--output paths. AI and humans are different kinds of subjects with complementary error-detection capacities.

This positioning removes AI from the single evaluative axis of ``surpassing humans'' or ``falling short of humans.'' AI, as another kind of inference subject, illuminates domains where human error-detection capacity does not reach. At the same time, at the core of human error-detection capacity lie domains that AI does not reach. The two cannot be ranked on the same ordering.

\subsection{The Temptation of AI Governance}

If AI can detect more prediction errors than humans, it might seem to acquire the legitimacy to be entrusted with governance. This is a new variation on an old question. Plato's philosopher-king, Bentham's felicific calculus, and the central planning authority that Hayek criticized in the socialist calculation debate---all shared the premise that ``the one who sees more governs better.'' Among arguments concerning AI governance, algorithmic decision-making, and AI policy optimization, there is a current that seeks to reintroduce the same premise on the sole ground that computability has increased.

The philosophical grounds can be organized as follows. First, an epistemic ground: AI sees more, so it can judge more accurately. Second, an efficiency ground: AI is free from human cognitive biases and emotional fluctuations, so it can make more consistent decisions. Third, a fairness ground: AI is independent of individual interests, so it can render more neutral judgments. Each of these grounds contains a partial truth. AI does indeed at times see more than humans. It can generate output more consistent than that of humans. It can perform statistical processing independent of individual interests.

Under the framework of this paper, however, three structural problems are pointed out with this temptation. First, AI's $\theta^{\mathrm{AI}}$ and $\zeta^{\mathrm{AI}}$ have biased foregrounding tendencies imposed by the training data and the loss function. AI ``sees more'' in domains for which it has been designed to see more easily. Second, the prediction errors AI detects do not pass through the historically accumulated paths of error selection in human society---law, institutions, culture, common sense. Third, the moment governance is entrusted to AI, AI's judgments become an unauditable black box for those governed, and the capacity for self-correction is concentrated in the hands of AI developers and operators.

The temptation of governance errs in automatically connecting ``seeing more'' to ``judging better.'' The legitimacy of error selection and the capacity for error detection are different problems. AI governance requires a logical leap from the superiority of detection capacity to the legitimacy of selection authority. The present paper rejects this leap.

\subsection{Human Society Is Not an Object of AI-Style Optimization}

The claim of this section is not an ethical appeal. Given MIA, to optimize human society under a single evaluation function is a constructive consequence: it reproduces, in the context of social world-model updating, structural breakdowns isomorphic to overfitting, local-minimum convergence, and out-of-distribution generalization failure in machine learning. Concretely, an optimization objective defined under a single profile $\Omega$ is tantamount to compressing the high-dimensional space $\Zphase$ into a low-dimensional projection, and structurally renders invisible prediction errors in directions not projected. This damages the distributed error-detection capacity of society---that is, its generalization performance itself. In what follows, we develop this consequence in MIM's vocabulary.

Human society is not an error-minimization device. It is a place where the world is lived through body, existence, culture, institution, and common sense. The variables statistically optimized by a given policy do not add up, in the same dimension, with the existential damageability, relational burden, and loss of cultural meaning for the individuals who live under it. An institution optimal on a decadal scale cannot be compared in the same unit with the bodily sense of subjects living one day. An economic policy that maximizes aggregate GDP is not measured on the same scale as the loss of meaning of traditional livelihoods that it destroys.

This is not the naive counterargument that ``there are things that cannot be numericized.'' In this paper's framework, human society preserves different sufficient statistics and different error costs in different coarse-graining domains---empirical, ideational, structural, and existential. These are mutually translatable but mutually irreducible. AI-style optimization is, in many cases, formulated as minimization of errors in the structural domain. The formulation itself can be consistent within the structural domain, but the consequence---what changes in state representation it brings about in the empirical, ideational, and existential domains---remains a separate problem.

The involvement of AI in human society must therefore not be comprehensively described by the metaphor of ``optimization.'' Society is not an object of optimization. It is a dynamic process in which multiple inference subjects continually detect, select, and coordinate multiple errors in multiple domains, and this process itself constitutes the self-correcting capacity of society. When AI engages with society, it is positioned not as an optimization device but as a single participant in this self-correcting capacity.

\subsection{The Role of AI: Mediation, Not Domination}

AI should not become a sovereign optimizer of human society. AI is not the sovereign of error selection. AI is a mediating apparatus that renders visible the prediction errors detected by different subjects, presents translation paths among different phases, places multiple selection options in comparable juxtaposition, simulates the consequences of selection, and audits past selections. Sovereignty---the authority that finally decides which errors to pick up, which to discard, and which to defer---rests with human subjects and the institutional and cultural constructs of human society. AI does not act as a proxy for that sovereignty.

This distinction structurally corresponds to the distinction Habermas~\cite{habermas1981} developed in \emph{Theory of Communicative Action} between strategic action and communicative action. Strategic action treats others as means to the actor's own purposes and does not substantively ask for their consent. Communicative action treats others as partners in the mutual examination of validity claims, and consent is constituted through the exchange of reasons. In this paper's vocabulary, strategic action is the operation of unilaterally imposing the sender's state representation without regard for the receiver's $\theta^{\beta}, \zeta^{\beta}$; communicative action is the operation of aligning the state representation into a processable form via $\Phi_{\alpha\to\beta}^{K\to K'}$. AI acting as the sovereign of governance corresponds to scaled-up strategic action. AI acting as a mediating apparatus corresponds to the infrastructure of communicative action, which supports the $\Phi$ network on a large scale.

The central problem of AI design therefore shifts from ``which values should AI hold'' to ``how can AI support $\Phi$ among human subjects.'' The former is the refinement of strategic action; the latter is the construction of infrastructure for communicative action. The present paper supports the latter position.

\subsection{AI that Displays Transformation Loss}

The alignment map $\Phi$ does not, in general, preserve information. When the sender's state representation $T_K^{\alpha}$ is mapped to the receiver's processable region $\tilde T_{K'}^{\beta}$, something is necessarily lost. In the process of abstracting empirical detail into institutional vocabulary, situatedness is lost. In the process of organizing existential textures into structural propositions, bodily feel is lost. In the process of translating cultural meaning into legal vocabulary, historical depth is lost. These are not failures of translation but the constitutive costs of translation.

If AI is to operate as a mediating apparatus for $\Phi$, it bears the responsibility of displaying this transformation loss. What is picked up, what is discarded, what is deferred, and what is rendered invisible must be presented as another informational layer on the same screen as the translation result. This is not an exoneration label that ``the translation is incomplete.'' It is a design that treats transformation loss itself as information and converts it into an input for the next judgment.

Implementational implications can be stated in one paragraph. In a dialogue UI, when AI generates a response, it displays as a domain label ``from which coarse-graining domain this response is described,'' and presents in parallel, as ``domains not touched by this response,'' information from other domains not contained in the response. In a document-summarization UI, alongside the summary, the system displays ``a category-wise list of matters present in the source text but not included in the summary.'' In a decision-support UI, for each option, the system symmetrically displays ``the error categories that this option foregrounds'' and ``the error categories that this option renders invisible.'' All of these constitute a design shift, from a UI that presents a single optimal solution to a UI that treats transformation loss as information.

\paragraph{Outlook Toward an Information-Theoretic Formalization of Transformation Loss}

Transformation loss has been articulated conceptually up to here, but an outlook toward information-theoretic formalization is also possible. Letting $I(T_K^{\alpha}; Y_K^{\alpha})$ denote the amount of information the sender's state representation $T_K^{\alpha}$ retains about the estimation target $Y_K^{\alpha}$, and $I(\tilde T_{K'}^{\beta}; Y_K^{\alpha})$ the amount of information the post-alignment state representation $\tilde T_{K'}^{\beta}$ retains about the same estimation target, a first approximation of transformation loss is given by the difference of mutual information,
\begin{equation}
\Delta I_{\Phi}=I(T_K^{\alpha}; Y_K^{\alpha})-I(\tilde T_{K'}^{\beta}; Y_K^{\alpha})
\end{equation}
This is isomorphic to the compression residual in information-bottleneck theory~\cite{tishby1999}, and connects to the information-geometric framework that treats $\Phi_T$ as a map in natural-parameter space~\cite{amari2016}. Furthermore, in translation across different coarse-graining domains, a non-commutativity that cannot be captured by a single mutual information remains---incompressible components proper to empirical detail, existential texture, and cultural meaning, respectively. Formalizing these as domain-wise weightings of the distortion function in rate--distortion theory becomes a subject for subsequent technical research. The present paper restricts itself to articulating transformation loss conceptually, but this information-theoretic connection provides a research hook that opens the mathematization of state-representation alignment to the machine-learning and information-geometry communities.

\subsection{AI that Presents Multiple Error-Selection Options}

AI is not a device that presents a single optimal solution. It is a device that places multiple error-selection options in juxtaposition, together with their respective structural consequences. In this paper's framework, any given policy, response, or decision necessarily involves a selection of ``which errors to pick up, which to discard, which to defer, which to render invisible, and which to amplify.'' For the same problem, there can be multiple combinations of such selection. AI structures and presents this multiplicity rather than erasing it.

For each selection option, AI shows the following five. First, the \emph{errors picked up} by the option---which prediction errors of which phases are foregrounded and become objects of action and updating. Second, the \emph{errors discarded} by the option---which prediction errors of which phases are deemed to fall within the allowable range of the selection and are removed from the processing target. Third, the \emph{errors deferred} by the option---which prediction errors of which phases are recorded as objects whose judgment is for the moment held in reserve. Fourth, the \emph{errors rendered invisible} by the option---which prediction errors of which phases can no longer be detected under the vocabulary of the selection. Fifth, the \emph{errors amplified} by the option---which prediction errors of which phases newly arise or expand as side-effects of the selection. Among these five, the fourth and fifth are particularly important. ``Rendered invisible'' refers to a process in which errors are not merely brought within an allowable range but cease to be detectable under the given vocabulary. ``Amplification'' refers to a process in which selection generates new errors. Both indicate paths through which a selection that appears on the surface to be an improvement produces structural loss behind the scenes.

The presentation of multiple options does not mean ``an AI that cannot decide.'' AI has the capacity to organize the structural consequences of each option and to juxtapose them. The final decision of which option to adopt is left to the institutional and cultural selection processes of human subjects and human society. What AI presents is the option space and the structural profile of each option; sovereignty does not remain there.

\subsection{Self-Audit by AI}

AI cannot perform error selection in a fully neutral manner. AI's $\theta^{\mathrm{AI}}, \zeta^{\mathrm{AI}}$ have biased foregrounding tendencies imposed by the training data distribution, the loss function, and architectural inductive biases. AI tends to pick up errors in certain phases more easily and tends to render errors in other phases invisible. This is not a defect of AI but a structural property. The problem lies less in the property itself than in the fact that it is operated upon without being made aware.

AI alignment therefore includes self-audit by AI. A design is required in which AI can estimate its own foregrounding tendencies at a meta-level and make explicit ``which phase's errors it tends to pick up,'' ``which phase's errors it tends to render invisible,'' and ``which state representations it tends to generate for which user populations.'' This is a self-reflective meta-level alignment, a fourth working domain irreducible to the value layer, the state-representation layer, or the world-model layer.

The implementation of audit includes at least the following: phase-wise classification of output logs, systematic comparison of responses across different user populations, continuous measurement of the domain distribution of responses to identical inputs, disclosure of phase-wise bias in the training data distribution, and explication of the error categories implicitly foregrounded by the loss function. These are at once technical audits and a constructive decision that AI positions itself as ``a subject that can also be treated as an object of audit.''

Audit does not complete in AI alone. The results of AI's self-audit are positioned as input to be disclosed to the institutional audit processes of human society---third-party evaluation, regulation, scholarly verification, and citizen-participatory monitoring. A two-layer structure of self-audit by AI and external audit by human society constitutes the self-reflective meta-level of AI alignment.

\subsection{Redefining AI Alignment}

The discussions of the preceding nine sections are summarized as a single-sentence working definition.

\bigskip
\begin{quote}
\emph{AI alignment is not the problem of fitting AI to a single human value, but the design problem of extending --- without being dominated by AI --- human society's multi-phase capacity for error detection, error selection, and error coordination.}
\end{quote}
\bigskip

This definition is not a denial of existing AI alignment research. It repositions existing research as the limited region of the value layer $\Phi_U$, and opens the problems of the upstream state-representation layer $\Phi_T$ and the world-model layer $\Phi_W$ as central tasks. At the same time, it repositions AI as a new error-detecting subject, constructively rejects the temptation of AI governance, restricts the role of AI to that of a mediating apparatus, and requires, in order, the design requirements of displaying transformation loss, presenting multiple options, and self-audit.

This redefinition extends AI alignment from a technical problem to a design problem. It is a shift from the question ``which optimization should be run'' to the question ``which social and cognitive capacities should be supported, under which distributed structure, with which allocation of sovereignty.'' As the central thesis of this paper indicates, human society has maintained its self-correcting capacity by not compressing prediction errors into a single authority, a single value function, or a single progressivist philosophy of history. The ultimate aim of AI alignment is neither to destroy this distributed structure nor simply to maintain it, but to extend it by means of AI. In Section~12 we proceed to the discussion of the multi-phasing of design that this redefinition opens.

\section{Multi-Phasing the Design Process Itself}

The central thesis of this paper is that human society is a structure that has maintained self-correcting capacity by distributively preserving, translating, and coordinating the prediction errors detected by different inference subjects, rather than compressing them into a single authority, a single value function, or a single progressivist philosophy of history. The present section develops this thesis as a design theory. It asks how the design process itself must be organized so that the design of social institutions and AI systems is assembled in a direction that strengthens, rather than destroys, that self-correcting capacity. This section is the arrival point of the present paper. The central thesis is here transformed into the concrete question ``who may design what, and how.''

The reason this question follows from the formal theory is that designers are also finite subjects. Their design choices arise through particular estimation targets, state representations, operating profiles, error costs, and alignment maps. Multi-phase design is therefore not a procedural ornament added to AI governance; it is the meta-level application of MIM to the design process itself.''

\subsection{The Single-World-Model Problem of Designers}

A designer of institutions or AI systems is also a single human subject with a particular profile state $\Omega = (\theta, \lambda, q, \zeta, \sigma)$ and a particular world model $W$. The designer's world model carries estimation targets $Y$ they foreground easily, state representations $T$ they extract easily, prediction errors $L$ to which they are sensitive, and an internalized error-cost structure $c$. This is not an argument that criticizes designers as ``biased.'' Every subject is biased. It is the constructive consequence---repeatedly shown from Section~3 of this paper---that under finite data, partial observation, and representational constraints, world models cannot but be biased endogenously.

The problem lies in the point that, under the multi-phase inference assumption (MIA), extending the designer's world model as ``the correct world model'' to the whole of society fails in principle. Errors that the designer does not foreground do not appear in the design. Prediction errors that the designer does not treat as high-cost do not enter the design's evaluation function. Events that cannot be cut out by the designer's state-representation extractor $\tau$ do not enter the design process as input in the first place. This is not a problem of the designer's good or ill or of their capacity, but of the finiteness of the error-detection bandwidth that a single world model can process.

To close design to a single expert group, a single regulatory agency, a single optimization objective, a single benchmark, or a single user model is to compress society's distributed error-detection capacity into the designer's world-model error-detection bandwidth by force. In the short term, efficiency and consistency are gained. In the long term, errors the designer does not see accumulate. This reproduces, at the layer of design, the same structure as the social fragmentation described in Section~9.

A multi-phase society cannot be designed from a single world model. The singleness of design is mapped onto a singleness of society's self-correcting capacity. This is not a technical choice but a structural constraint.

\subsection{Opening the Design Process to Error Detection}

The design process must be opened so that diverse subjects can input different errors. ``Opened'' here is not only the procedural openness of existing arrangements such as public consultation, public comment, and stakeholder meetings. Those can function as fundamental filtering devices in which only the opinions already shaped into a form receivable by the designer's state-representation space arrive.

To rephrase in this paper's vocabulary. The prediction errors $L^{\alpha}$ detected by diverse subjects, the estimation targets $Y^{\alpha}$ behind them, the state representations $T^{\alpha}$, and the error costs $c^{\alpha}$ must be rendered visible as inputs to the design process, and must be subjected to an alignment map $\Phi$ into a form processable in the designer's world model. More importantly, what is preserved and what is lost in the operation of $\Phi$ must be explicitly displayed as transformation loss. ``We have understood your concern,'' the designer's acknowledgment, is often the operation of treating the other's prediction error as a residual that has been forcibly projected onto the designer's state-representation space.

The openness of the design process is not the number of input channels. It is that input prediction errors may also come from inference phases not foregrounded by the designer; that the designer recognizes this and accepts the transformation loss. This is difficult technically and organizationally. The designer does not possess, in themselves, a procedure for rendering processable, while keeping them difficult to see, the errors that their own operating profile struggles to capture. For precisely this reason, subjects with different operating profiles must be structurally incorporated into the design process itself. Not as the ex post input of a vote or hearing, but as the very initial condition of design, the multi-phasing of the error-detection bandwidth itself must be required.

\subsection{Expert Knowledge, Life-World Knowledge, Traditional Knowledge, and Stakeholder Knowledge}

Expert knowledge alone is insufficient as input to the design process. Expert knowledge is a system of state representations and evaluation functions highly organized within a particular inference phase. It works powerfully within that inference phase. But it is constructively impossible to detect, by expert knowledge alone, errors that expert knowledge does not foreground. In the framework of this paper, error detection is relative to inference phase.

The design process requires at least the following different kinds of knowledge. Life-world knowledge possesses empirical state representations accumulated through daily operation, near-field manipulation, and immediate feedback. Traditional knowledge preserves long-range prediction-error sensitivities that have survived across generations, in the form of custom, ritual, norm, and narrative. Stakeholder knowledge is the capacity of subjects directly affected by the design to process that impact as prediction error in the existential domain. Field knowledge possesses both coarse-graining and fine-graining strategies that have absorbed, in the details of implementation, the friction between design and reality.

These are not mutually untranslatable. But the errors each detects are not preserved when transferred directly into the state representations of another knowledge. The work that AI ought to do is not to compress all of these into a single evaluation function $u_{\mathrm{design}}$. Designing AI as a single evaluation-function generator reproduces, at one meta-level higher, the single-world-model problem of design.

What is required of AI is to render the prediction errors of different knowledges processable in parallel within the design process, while displaying transformation loss. For example, when a request that is ``low in feasibility'' from the standpoint of expert knowledge is, from the standpoint of stakeholder knowledge, a prediction error that ``all other options have been eliminated,'' the two errors are not to be compared on the same scalar; rather, what is to be presented in parallel is which error cost of which inference phase each error arises from. On the basis of that parallel presentation, the designer self-consciously chooses which errors to protect and which transformation losses to accept. This is the smallest and most concrete unit of the multi-phasing of design.

\subsection{Multi-Phase Pluralism: Redescribing Freedom, Rights, and Dissent}

The multi-phase pluralism this paper presents is a social-design principle for distributively preserving, translating, and coordinating the prediction errors detected by different inference subjects, rather than converging society onto a single value system or a single rationality. This term is not a political label but a working concept that designates the social-design conditions implied by MIA. The multi-phase pluralism here is not value pluralism in Berlin's sense as such~\cite{berlin1969}. In addition to the claim that values are plural, it includes the operative claim that what is detected as error, which state representations are preserved, and which transformation losses are deemed impermissible vary multi-phase-wise across subjects and institutions. This standpoint gives a vocabulary that redescribes the issues that the liberal tradition has problematized---freedom, rights, dissent, and institutional deliberation---not as value pluralism, but as conditions that protect society's distributed error-detection capacity.

Freedom, rights, dissent, protection of minorities, and institutional deliberation are error-input mechanisms by which society does not become rigidly fixed into a single world model. This does not assert liberalism as a particular partisan position. It states, rather, formal conditions for not compressing different error signals into a single authority, a single value function, or a single progressivist philosophy of history.

\subsection{Traditional Knowledge and Long-Term Error Sensitivity}

Tradition retains, within itself, the traces of prediction errors paid in the past as custom, norm, ritual, bodily skill, and narrative. As argued in Section~7, the conservative thought of Burke and Oakeshott has sought to preserve such long-term error sensitivities beyond the reach of rational design. This subsection is not here to repeat that argument. It asks through which implementational mechanisms the long-term error sensitivity preserved by tradition can be incorporated into institutional design and AI design.

The prediction errors preserved by traditional knowledge are transmitted not in the language of formalized rules, but as implicit taboos, senses of unease, and feelings of ``things not working.'' To treat these as inputs to the design process requires at least three implementational moments. First, a device that presents in parallel, while retaining transformation loss, the state representations internal to the tradition, without forcibly translating them into the designer's state-representation space. Second, a device that extends the time axis of the design process, treating the errors accumulated by tradition on generational scales as long-term prediction errors that short-term optimization's evaluation function cannot detect. Third, a device that makes the shared error-cost structure that tradition has tacitly relied upon explicit, so that it can become an object of conscious critique.

The third point is the most important implementational issue of this section. Tradition both preserves the errors it has detected and renders invisible errors of another kind. The error-cost allocation $c^{\mathrm{trad}}$ taken for granted within a tradition may have processed the prediction errors of subjects on the periphery or outside of the tradition as noise that does not register in the tradition's state-representation space. In this paper's framework, protecting the long-term error sensitivity of tradition and foregrounding the errors that tradition has rendered invisible do not contradict each other. Both are two consequences of the same principle: the protection of society's distributed error-detection capacity.

The multi-phasing of design treats traditional knowledge neither as a ``heritage to be preserved'' nor as a ``residue to be overcome.'' It treats traditional knowledge as a group of special operating profiles carrying long-term error sensitivity, and constructs an alignment map $\Phi$, retaining transformation loss, between its state representations and other state representations. By this, the error sensitivity internal to tradition becomes processable within the contemporary design process, and at the same time, the errors that tradition has rendered invisible are foregrounded in external state-representation spaces.

\subsection{Radicalism, Dogmatism, and Disorder}

The failure modes that the multi-phasing of design seeks to avoid can be organized into at least three. Each destroys society's distributed error-detection capacity in a different manner.

First, radicalism. This is the tendency to erase, at one stroke, the error sensitivities that existing institutions, customs, and traditions have preserved, and to reconstruct society in accordance with the designer's world model. What Burke's critique of Revolution~\cite{burke1790} problematized was not the ideal of revolution itself, but the structure in which revolution pays the cost of discarding wholesale generational-scale error-absorbing devices. What Oakeshott's critique of Rationalism~\cite{oakeshott1962} warned against was the error sensitivity that political rationalism loses when it tries to replace tacit and practical knowledge with formalized design reason. In this paper's vocabulary, radicalism is the operation of leaving only those errors that the designer's world model foregrounds and deleting, from the initial conditions of design, the error sensitivities preserved distributively across generations.

Second, dogmatism. This is the tendency to fix a particular world model, a particular evaluation function, or a particular philosophy of history as ``the uniquely correct one,'' and to structurally erase prediction errors that do not fit. Popper's theory of the open society~\cite{popper1945} criticized the structure whereby historicism and totalitarianism compress society's self-correcting capacity into a single historical necessity or a single guiding principle. Hayek's critique of constructivist rationalism~\cite{hayek1973} problematized the structure whereby attempts to construct social order by central design reason destroy the error-absorbing capacity of spontaneous orders that have evolved distributively. In this paper's vocabulary, dogmatism is the operation of fixing society's foregrounding gradient field $R$ to a single operating profile, thereby rendering, in principle, unprocessable the prediction errors detected by other operating profiles.

Third, disorder. This is the tendency to confuse the multi-phasing of error detection with the absence of an error-selection apparatus. What Hobbes's theory of the state of nature~\cite{hobbes1651} depicted was the structure that, in the absence of any error-selection apparatus, diverse prediction errors can only erase one another or be processed as violent collision. What Habermas problematized as system distortion~\cite{habermas1981} was the structure whereby formally open deliberation, owing to structural distortions of power, economy, and media, comes to render particular prediction errors invisible. The relation between law and democracy that Habermas developed in ``Faktizit\"at und Geltung''~\cite{habermas1992} can also be read as an argument that, as a condition for normative validity to acquire factual binding force, an error-selection apparatus must be institutionally guaranteed. In this paper's vocabulary, disorder is the dissolution of the social world model that occurs when the multi-phasing of error proceeds without being accompanied by error processability and the alignment map $\Phi$.

Multi-phase pluralism avoids these three failures by multi-phasing error selection itself. Against radicalism, it preserves the existing long-term error sensitivities together with their transformation loss. Against dogmatism, it forbids, as an initial condition of design, fixation to a single operating profile. Against disorder, it always assembles mutual processability via the alignment map $\Phi$ together with the multi-phasing of error detection. The three failures can be described, respectively, as falls into different singularities---the singularity of design reason, the singularity of historical philosophy, and the negative singularity of the absence of an error-detection apparatus. Multi-phase pluralism is a fourth design position, irreducible to any of these three.

\subsection{Connection to AI Regulation and AI Governance}

Current discussions of AI regulation address risk, safety, fairness, transparency, accountability, privacy, copyright, bias, and the like. The EU AI Act, the NIST AI Risk Management Framework, and the OECD AI Principles each treat the risks and responsibilities of AI in different institutional vocabularies~\cite{euai2024,nist2023,oecd2024}. The question MIM adds is: which prediction errors of society does AI render visible, which does it render invisible, and which state representations does it institutionally prioritize? If existing regulatory vocabulary chiefly addresses risk, rights, safety, transparency, and accountability, MIM connects these to the question of which errors are institutionally foregrounded and which errors remain as immeasurable, untranslatable, or low-priority.

AI governance is the problem of governing how AI rewires society's $\Zphase$, $R$, and $\Phi$. Even if AI does not render a judgment, the dashboards it shows, the risks it classifies, the issues it recommends, and the vocabularies it translates change society's estimation target space. AI regulation must therefore ask not only about the outputs of AI but also about how AI changes society's error-detection structure.

\subsection{The Principle of Multi-Phasing the Design Itself}

We summarize this section into a single design principle.

The principle of multi-phasing design --- The design of social institutions and AI systems must not be closed by a single expert group, a single evaluation function, a single optimization objective, or a single philosophy of history. Design must be opened so that subjects with different inference phases, different state-representation extractors, and different error-cost structures can input different prediction-error signals into the design process itself. The designer cannot, by their own world model alone, detect errors that their own operating profile does not foreground. The legitimacy of design therefore rests not on the correctness of the designer's world model, but on the multi-phasic-ness of the error-detection bandwidth built into the design process.

This principle does not open design to relativism. The multi-phasing of error detection is not the abandonment of error selection. The combination is required as indivisible: multi-phased error detection, mutual processability via the alignment map $\Phi$, the self-conscious selection of error cost $c$, and the explication of transformation loss. Only when these four are in place can design avoid falling into the three failures of radicalism, dogmatism, and disorder, and protect and strengthen society's distributed error-detection capacity.

What the central thesis of this paper has described is a structure in which human society has maintained its self-correcting capacity by resisting the single compression of error. The design principle presented in this section is intended to reconstitute that structure as a deliberate practice of design in the age of AI. AI is a device that can destroy this structure and at the same time a device that can strengthen it. Which direction it goes reduces to the design problem of multi-phasing design itself---namely, to which operating profile AI design is closed, and to which group of operating profiles it is opened.

\section{Empirical Verifiability and Technical Research}

A theory that explains too much too comfortably should make its reader suspicious. The present framework is not offered as a doctrine immune to failure; it should become stronger only where it can be measured, challenged, and made technically operational. The central structure derived in Sections~3--5 is that a finite subject constructs approximate sufficient statistics from an observation sequence, and then performs foregrounding, exploration, and stabilization within the candidate space generated thereby. Accordingly, empirical verifiability does not consist in directly measuring inner contents. It consists in measuring which candidate sufficient statistic $c=(Y,\psi,\rho,T)$ is constructed, which prediction error $L(c)$ is triggered, and which operating fields $r,e,s$ orient that processing. In this section we organize empirical verifiability across three layers---intra-individual measurement, world-model alignment measurement, and AI-system evaluation---and then situate the remaining limitations as constructive tasks for future research.

\subsection{Intra-individual measurement of the operating profile}

First, the operating profile of the world model can be measured within a single individual. Given the same stimulus, we estimate which candidate sufficient statistic $c=(Y,\psi,\rho,T)$ the subject constructs, which prediction error $L(c)$ is routed to reflective processing, and which candidates are explored or stabilized. This estimation draws on free-response descriptions, forced-choice tasks, physiological indices, and behavioral indices. What is measured here is not the process by which an estimation target is directly extracted from the observation sequence, but rather which candidate within the constructible candidate space $\mathcal C_t^\alpha$ has in fact been activated.

The prior operating profile $\theta=(r,e,s)$ can be estimated, across multiple tasks, from which candidates in the candidate sufficient statistic space tend to be foregrounded, which candidates host wider exploration, and which candidates are more readily stabilized. The posterior operating state $\tilde\theta=(\tilde r,\tilde e,\tilde s)$ can be estimated, in response to a specific stimulus, from which candidate sufficient statistic is actually triggered and whether processing is directed toward exploration or stabilization.

The representation-formation and processing profile $\zeta=(\chiop,\tau,\kappa)$ can be measured by how the subject forms candidate estimation targets, how the state representation $T$ is constructed under a given conditioning basis and resolution, and to what degree the subject can compress, symbolize, and externalize that representation. The accuracy and stability of these processing operations are measured separately as directional maturity $\sigma$. The thresholds and error costs that gate entry into reflective processing are not attributes of the operating profile $\theta$. They are measured as the trigger-and-awareness threshold component $q=(c_{\mathrm{err}},\eta)$ of the profile state $\Omega$. Here $c_{\mathrm{err}}$ denotes the error cost and is distinct from the symbol $c$ used for the candidate sufficient statistic.

\subsection{Measurement of world-model alignment}

Second, the success or failure of world-model alignment $\Phi$ can be measured. The same content is presented after being aligned into empirical, ideal, structural, and existential forms, and the recipient's comprehension, acceptance, resistance, and behavioral intention are measured.

For example, an institutional problem expressed in the structural region may be transformed into expressions of responsibility load or reassurance that can be processed in the existential region. Conversely, a sense of unease expressed in the existential region may be transformed into a question of institutional design that can be processed in the structural region. In each case, one measures whether the post-alignment state representation lowers the recipient's $L_{K'}$, or, when the threshold is exceeded, whether it carries positive directional compatibility.

\subsection{Evaluation of AI-system alignment support}

Third, one can evaluate whether an AI system can support world-model alignment. An evaluation pipeline can be composed of: foregrounded-phase classification of input utterances; generation of candidate sufficient statistics; extraction of state representations; estimation of the recipient's profile state; generation of alignments across phases; explanation of alignment loss; and human-rated measurement of gains in comprehension and reductions in resistance.

An alignment produced by an AI system is not adequate merely by being a polite paraphrase. What evaluation must target is whether the sender's $T_K^{\alpha}$ has been aligned into an approximate sufficient statistic $\tilde T_{K'}^{\beta}$ that is processable under the recipient's profile state $\Omega^{\beta}=(\theta^{\beta},\lambda^{\beta},q^{\beta},\zeta^{\beta},\sigma^{\beta})$.

\subsection{Verifiability hypotheses}

The framework above can be made explicit as several verifiable hypotheses. In order to fix the theoretical scope of this paper as a researchable theory, we present the following five core hypotheses.

\textbf{H1 (Inter-subject variation and coarse-graining of candidate sufficient statistics).} For a population of subjects $\{\alpha_i\}$ receiving the same observation sequence $o_{1:t}$, the free-response descriptions, behavioral indices, and physiological indices obtained from each subject can be observed as differences in the constructed candidate sufficient statistic $c_i=(Y_i,\psi_i,\rho_i,T_i)$ and prediction error $L(c_i)$. Moreover, we hypothesize that these differences can be estimated as continuous foregrounding tendencies on $\Zphase$, and, when projected as needed onto the empirical, ideal, structural, and existential coarse-graining regions, exhibit explanatory power significantly higher than random assignment. The four vocabularies are therefore not axioms of MIM but coarse-graining hypotheses open to empirical verification.

\textbf{H2 (Receptivity prediction from the profile state).} When the profile state $\Omega^{\beta}=(\theta^{\beta},\lambda^{\beta},q^{\beta},\zeta^{\beta},\sigma^{\beta})$ of subject $\beta$ has been estimated in advance, the question of which among a set of explanation forms $\{T^{(K_1)}, T^{(K_2)}, \ldots\}$ will induce receptive processing for $\beta$ (namely $\mu > 0$ and $I < \eta$) can be predicted with accuracy significantly higher than chance. That is, $\theta^{\beta}$ encodes tendencies for foregrounding, exploration, and stabilization of candidate sufficient statistics; $\lambda^{\beta}$ encodes their plasticity; $q^{\beta}$ encodes error-trigger thresholds; $\zeta^{\beta}$ encodes the mode of representation formation; and $\sigma^{\beta}$ encodes directional maturity. Together they function as features that predict the receptivity of explanation forms.

\textbf{H3 (Effect of state-representation alignment).} An explanation produced by applying the alignment operation $\Phi_{\alpha\to\beta}^{K\to K'}$, which transforms the sender's state representation $T_K^{\alpha}$ into a form processable under the recipient's profile state $\Omega^{\beta}$, will, compared with a non-aligned explanation (the sender's original state representation presented as is), significantly increase the recipient's comprehension, significantly decrease resistance and rejection responses, and facilitate uptake into behavioral intention. This effect is sensitive to the choice of intermediate phase $K \to K'$ in $\Phi$, and an inappropriate phase choice can produce a counterproductive effect.

\textbf{H4 (Resolution bias and error-absorption strategy).} Within the subject's foregrounding field $r^\alpha$, the component $r_{\mathrm{res}}^\alpha$ pertaining to resolution $\rho$ predicts whether prediction error is absorbed in the short term through action or retained and reorganized in the long term through reflection and conceptualization. Confronted with the same ambiguous situation, a subject who privileges fine-grained consistency is more likely to choose local behavioral intervention, while a subject who privileges coarse-grained consistency is more likely to choose long-horizon reconstruction of interpretive frameworks. This hypothesis can be verified by measuring, for the same scenario, the subject's preferences over proposed actions, conceptual reorganizations, temporal horizons, abstraction levels, and tolerated distortions.

\textbf{H5 (Constructive process of meaning, value, and justice).} The same utterance, event, policy, or conflict situation is constructed as different candidate sufficient statistics pertaining to meaning, value, and justice, depending on the subject's profile state $\Omega$. For instance, the same utterance may be processed by one subject as a matter of structural meaning, by another as a violation of existential value, and by a third as a procedural inconsistency. These differences can be measured through free-response descriptions, reaction times, resistance intensity, choices of what is to be protected, and choices of transformations that are not tolerated. Meaning, value, and justice are therefore not mere subjective contents; they can be operationally analyzed as tuples of estimation target, conditioning basis, resolution, state representation, prediction error, and error cost. This hypothesis would be weakened if differences described as meaning-, value-, or justice-oriented could be fully predicted without reference to differences in $(Y,\psi,\rho,T)$, transformation loss, or error-cost assignment.

These hypotheses can each be verified independently, and verifying them in combination provides a graded confirmation of the predictive power of the present framework. H1 assumes inter-subject variation of candidate sufficient statistics but presupposes neither $\Phi$ nor the three-layer hierarchy. H2 presupposes H1 and posits the predictive capacity of the profile state. H3 presupposes H2 and verifies the efficacy of the operation $\Phi$. H4 verifies whether foregrounding tendencies in resolution space predict error-absorption strategies. H5 verifies how meaning, value, and justice are realized as particular candidate sufficient statistic constructions. Subsequent research will require experimental designs and evaluation indices that verify these hypotheses in sequence.

\subsection{Limits of the present work---research objects to be constructed next}

In what follows we develop the principal limitations of the present formalization, each understood as a constructive task that it calls for. The fact that the formulas of this paper do not constitute immediately executable algorithms but are descriptive formalizations that fix the relations among the constituent elements is supplemented in Appendix~B.1, and the connection to future technical research is organized in Appendix~B.5.

We first distinguish the core of this paper from its extensions. The minimal core of the intra-individual generative mechanism comprises the constructibility conditions for approximate sufficient statistics, the candidate sufficient statistic space $\mathcal C_t^\alpha$, the phase-formation space $\Zphase$, and the foregrounding gradient field $R_t^\alpha$ over $\Zphase$ that determines the direction in which prediction errors are processed. In contrast, the alignment map $\Phi$ is the principal device that makes state representations arising from the intra-individual mechanism processable across subjects; it bears the connection to social fragmentation and to AI alignment. The four coarse-graining regions, the cognitive typology, and the redescriptions of meaning, value, justice, culture, and civilization are low-dimensional approximations, applications, and research programs opened by the intra-individual generative mechanism $(\mathcal C,\Zphase,R)$ and by the inter-subject alignment mechanism $\Phi$. The present paper does not therefore provide a fully closed generative model or estimation algorithm. By introducing the candidate sufficient statistic space and the foregrounding gradient field, however, it lays out the placement of the constituent elements that connect observation sequences, candidate construction, prediction error, output, and feedback update. In what follows we organize this pathway in the order of candidate generation, plan candidates, action constraints, and feedback update.

\subsection{Outlook toward generative-model implementation---from multi-phase candidates to plan candidates}

Cognition does not terminate with the formation of a state representation. A finite subject uses candidate sufficient statistics constructed from the observation sequence to evaluate prediction errors, to generate plan candidates, and to feed the results back into subsequent observation and world-model updates. In this subsection we sketch a route by which the formal apparatus of the present paper may be made concrete as a dynamic generative model. We do not provide a completed computational model; instead we exhibit a conceptual loop whose skeleton is to be made concrete in subsequent research.

More constructively, the observation sequence $o_{1:t}^\alpha$, under the subject's world-model state $W_t^\alpha$ and profile state $\Omega_t^\alpha$, forms the candidate sufficient statistic space $\mathcal C_t^\alpha$. We write an element of this space as $c=(Y,\psi,\rho,T)$. Each candidate jointly specifies an estimation target, a conditioning basis, a resolution, and a state representation. Letting $L_t^\alpha(c)$ denote the prediction error for candidate $c$ and $x(c)=(Y,\psi,\rho)$ the corresponding phase coordinate, the foregrounding gradient field can be written as
\begin{equation}
v_t^\alpha(c)=R_t^\alpha(x(c),L_t^\alpha(c);\theta_t^\alpha,\zeta_t^\alpha),
\end{equation}
giving a processing direction over the candidate sufficient statistic space. The priority over candidates may be written, for example, as
\begin{equation}
\pi_t^\alpha(c)=\operatorname{softmax}_{c\in\mathcal C_t^\alpha}\left(S^\alpha(c;o_{1:t}^\alpha,L_t^\alpha(c),v_t^\alpha(c),\Omega_t^\alpha)\right).
\end{equation}
This is not a distribution over estimation targets alone but a distribution over which estimation target, through which conditioning basis and resolution, is to be constructed as which state representation.

Once a candidate $c$ is foregrounded, its state representation $T$ and prediction error $L_t^\alpha(c)$ are weighted by the trigger-and-awareness threshold component $q_t^\alpha$ of the profile state $\Omega_t^\alpha$. Writing the error cost as $c_{\mathrm{err},t}^\alpha(c)$, the significant error intensity is
\begin{equation}
I_t^{\alpha}(c)=c_{\mathrm{err},t}^{\alpha}(c)L_t^{\alpha}(c).
\end{equation}
When this exceeds the threshold $\eta_t^\alpha(c)$, candidate $c$ is routed into reflective processing, verbalization, action, or conscious suspension.

A foregrounded candidate $c_t^{\alpha}=(Y_t^{\alpha},\psi_t^{\alpha},\rho_t^{\alpha},T_t^{\alpha})$ generates plan candidates $P_{i,t}^{\alpha}$ such as reflection, suspension, reinterpretation, exploration, explanation, utterance, bodily action, or coordination. Up to this stage, the observation sequence is transformed into multiple plan candidates by way of the candidate sufficient statistic space, candidate-specific prediction errors, significant error intensities, and foregrounding vectors. Generative-model implementation of MIM requires, as a first step, that this candidate-generation process be made concrete as a probability distribution.

\subsection{Action constraints and feedback update}

Generated plan candidates are not selected solely on the basis of their expected reduction of prediction error. They are filtered by whether they satisfy constraints arising from embodiment, time, skill, coordination, and communication. We refer to this broad notion of executability as the action constraint $C_{\mathrm{act}}$.
\begin{equation}
C_{\mathrm{act}}^{\alpha}(P_i)=
C_{\mathrm{body}}^{\alpha}(P_i)+
C_{\mathrm{time}}^{\alpha}(P_i)+
C_{\mathrm{skill}}^{\alpha}(P_i)+
C_{\mathrm{coop}}^{\alpha}(P_i)+
C_{\mathrm{comm}}^{\alpha}(P_i)+\cdots.
\end{equation}
Here $C_{\mathrm{body}}$ is the embodiment constraint, $C_{\mathrm{time}}$ is the temporal constraint, $C_{\mathrm{skill}}$ is the skill-and-maturity constraint, $C_{\mathrm{coop}}$ is the coordination constraint, and $C_{\mathrm{comm}}$ is the communication constraint, including verbalization, explanation, and externalization. We have written the expression above in additive form for notational brevity, but the constraints in fact interact. When temporal constraints are tight, the costs of the embodiment and coordination constraints change as well, and coordination and communication constraints are strongly coupled. Strictly speaking, one should write
\begin{equation*}
C_{\mathrm{act}}^{\alpha}(P_i) = F_{\mathrm{act}}\left(C_{\mathrm{body}}, C_{\mathrm{time}}, C_{\mathrm{skill}}, C_{\mathrm{coop}}, C_{\mathrm{comm}}, \ldots; P_i\right),
\end{equation*}
a composite function for which the additive form serves as a first-order approximation.

The coordination constraint $C_{\mathrm{coop}}$ is a sub-constraint within the action-execution constraint $C_{\mathrm{act}}$ that becomes active when execution must pass through other subjects' world models, relationships, inner states, or consensus formation. The term ``action'' here is not restricted to external bodily action; it includes internal and social acts such as thought, suspension, reinterpretation, explanation, and coordination.

The priority of a plan candidate $P_i$ may be written, for example, as
\begin{equation}
\Pi_t^{\alpha}(P_i)=
F\left(
\Delta L_t^{\alpha}(P_i),
U_t^{\alpha}(P_i),
C_{\mathrm{comp}}^{\alpha}(P_i),
C_{\mathrm{obs}}^{\alpha}(P_i),
C_{\mathrm{act}}^{\alpha}(P_i),
R_t^{\alpha}
\right).
\end{equation}
Here $\Delta L_t^{\alpha}(P_i)$ denotes the expected reduction of prediction error by the plan, $U_t^{\alpha}(P_i)$ denotes how it preserves, protects, or damages value-related state representations, and $C_{\mathrm{act}}^{\alpha}(P_i)$ denotes whether the plan is executable. From these priority scores the plan-selection distribution may be derived as, for example,
\begin{equation*}
p_t^{\alpha}(P_i \mid c_t^\alpha, v_t^\alpha(c_t^\alpha)) = \operatorname{softmax}_i\left(\Pi_t^{\alpha}(P_i)\right),
\end{equation*}
which, in parallel with the selection distribution over candidate sufficient statistics, gives a probability distribution over plan candidates. When several plans are simultaneously candidated, they enter, with priorities, a working-memory queue
\begin{equation}
Q_t^{\alpha}=[P_1,P_2,\ldots,P_m].
\end{equation}
Working memory, however, is short-lived and prone to forgetting. If a high-priority plan stalls on error resolution or on consolidating its update to the generative model, lower-priority plans drop out of the explicit queue. Nevertheless, when repeated, they may sediment into the operating profile as low-intensity traces.

After a plan $P_i$ has been executed---or attempted as reflection, suspension, coordination, or utterance---the result can be expressed as a feedback quantity
\begin{equation}
\Delta_t^{\alpha}(P_i)=
(\Delta L_t^{\alpha},\Delta U_t^{\alpha},\Delta C_{\mathrm{act},t}^{\alpha},\Delta C_{\mathrm{coop},t}^{\alpha},\Delta\Phi_t^{\alpha}).
\end{equation}
Here $\Delta L$ indicates whether prediction error decreased or increased, $\Delta U$ indicates whether value-related state representations were preserved or damaged, $\Delta C_{\mathrm{act}}$ indicates whether the action, thought, or plan was executable, $\Delta C_{\mathrm{coop}}$ indicates whether coordination succeeded or relational friction increased, and $\Delta\Phi$ indicates whether state-representation alignment succeeded or failed. This feedback updates the foregrounding gradient field $R$ and the profile state $\Omega=(\theta,\lambda,q,\zeta,\sigma)$ for subsequent rounds. The equations below do not constitute estimation algorithms; they are structural templates indicating how action outcomes return to subsequent foregrounding, exploration, stabilization, plasticity, thresholds, representation formation, and directional maturity. Abstractly,
\begin{align}
R_{t+1}^{\alpha}
&=\mathcal U_R(R_t^{\alpha},v_t^{\alpha},\Delta_t^{\alpha}),\\
\Omega_{t+1}^{\alpha}
&=\mathcal U_{\Omega}(\Omega_t^{\alpha},c_t^{\alpha},v_t^{\alpha},\Delta_t^{\alpha}).
\end{align}
As needed, this update can be expanded as
\begin{align}
\theta_{t+1}^{\alpha}
&=\mathcal U_{\theta}(\theta_t^{\alpha},c_t^{\alpha},v_t^{\alpha},\Delta_t^{\alpha}),\\
\lambda_{t+1}^{\alpha}
&=\mathcal U_{\lambda}(\lambda_t^{\alpha},\theta_t^{\alpha},\Delta_t^{\alpha}),\\
q_{t+1}^{\alpha}
&=\mathcal U_q(q_t^{\alpha},L_t^{\alpha}(c_t^{\alpha}),\Delta_t^{\alpha}),\\
\zeta_{t+1}^{\alpha}
&=\mathcal U_{\zeta}(\zeta_t^{\alpha},Y_t^{\alpha},T_t^{\alpha},\Delta_t^{\alpha}),\\
\sigma_{t+1}^{\alpha}
&=\mathcal U_{\sigma}(\sigma_t^{\alpha},c_t^{\alpha},\Delta_t^{\alpha}).
\end{align}
Here each component of $\sigma$ is treated not as a fixed list of directions but as the maturity over a direction or candidate family $j\in\mathcal J^\alpha$ that is iteratively operated upon within the candidate sufficient statistic space.

This update, however, is not a simple matter of strengthening success and weakening failure. When action or thought along a given direction reduces prediction error, the gradient and maturity along that direction tend to be strengthened. Conversely, when failure occurs in a region of high error cost, that direction can be strengthened in a different form---as avoidance, hypersensitization, or defensive fixation. Maturity in MIM is therefore not the mere accumulation of successful experiences but a history-dependent change in operation that includes failure, avoidance, hypersensitization, and fixation.

By the above, MIM does not yet provide a fully closed generative model, but it obtains a conceptual loop by which it may be made concrete as a dynamic generative model, namely
\begin{equation}
\begin{aligned}
o_{1:t}^{\alpha}
&\rightarrow \mathcal C_t^{\alpha}
\rightarrow c_t^{\alpha}=(Y_t^{\alpha},\psi_t^{\alpha},\rho_t^{\alpha},T_t^{\alpha}) \\
&\rightarrow L_t^{\alpha}(c_t^{\alpha})
\rightarrow I_t^{\alpha}(c_t^{\alpha})
\rightarrow v_t^{\alpha}(c_t^{\alpha}) \\
&\rightarrow P_t^{\alpha}
\rightarrow y_t^{\alpha}
\rightarrow o_{t+1}^{\alpha}
\rightarrow \Delta_t^{\alpha}
\rightarrow W_{t+1}^{\alpha} \\
&= (M_{t+1}^{\alpha},\mathcal{Z}_{\mathrm{phase},t+1}^{\alpha},R_{t+1}^{\alpha},\Omega_{t+1}^{\alpha}).
\end{aligned}
\end{equation}
The remaining task is to make this loop concrete as a candidate-generation distribution over sufficient statistics, candidate-specific error functions, a composite objective function, an update rule for the foregrounding gradient field, an update rule for the working-memory queue, and an observation-generation model.

The generative loop above is the central task in constructing MIM as a dynamic generative model. In what follows we organize the research objects that remain in order to make that loop concrete through empirical verification, mathematical formalization, and connection to existing research. From this point onward, what follows is not the addition of new claims but a work-decomposition of the research program.

\textbf{Formal characterization of multi-channel parallelism (MIA).} This paper presents MIA as an alternative hypothesis to SIA and provides phenomenological and historical-philosophical motivation for it. The next constructive task is to decide whether ``parallelism'' is to be precisified as computational independence, temporal simultaneity, or non-commutativity of representations, and to construct an impossibility argument that formally identifies phenomena which a single-channel intelligence cannot explain. Specifically, this can be developed as a research plan that identifies empirical signatures of multi-channel parallel processing in behavioral and neural data---simultaneous formation of multiple estimation targets under the same stimulus, time scales of inter-phase transition, and processing costs in inter-phase competition.

\textbf{Structural description of the internal components of the resolution space $\Zres(\psi)$.} As argued in Section~4, the resolution space $\Zres(\psi)$ comprises a number of structural properties---temporal range, spatial range, abstraction level, and operability---as differing manifestations of rate--distortion trade-offs under computational constraints. The next constructive task is to precisify the independence and correlation of these components by means of principal component analysis, information geometry, and quantitative analysis of rate--distortion trade-offs. The extent to which the structure obtained as a low-dimensional projection preserves explanatory power and the extent to which it discards independent variation become objects of information-theoretic or statistical verification.

\textbf{Verification of the minimality and dimensionality of the phase-formation space.} Section~4 introduced $\Zphase$ as the conjunction of the conditioning-basis space $\Zref$ and the conditioning-basis-dependent resolution space $\Zres(\psi)$ under the three-constraint framework $(C_{\text{comp}}, C_{\text{obs}}, C_{\text{act}})$. The next constructive task is to verify the validity of the three-constraint framework, the internal dimensionality of $\Zref$ and $\Zres(\psi)$, the existence of further independent components, and the validity of low-dimensional projections, through empirical and mathematical examination based on behavioral and neural data.

\textbf{Constructive formalization of MIM operators and a minimal working example.} This paper defined the operators of the Multi-Phase Inference Mechanism, namely $(\chiop, \tau, \kappa)$, conceptually. The next constructive task is to build concrete constructions for each, together with the update process for directional compatibility $\mu$, trigger threshold $q$, and directional maturity $\sigma$---their functional forms, learning rules, and implementable algorithms---and to construct a \emph{minimal working example} on synthetic data that exhibits, in computable form, ``different candidate sufficient statistics $c=(Y,\psi,\rho,T)$ being constructed from the same observation sequence.'' This constitutes the first milestone along the path that leads to multi-subject MIM simulation (the research horizon of Section~14) and to embedding the MIM structure into LLM and VLA systems.

\textbf{Quantification of low-dimensional approximability.} This paper has argued that ``high-dimensional complex systems can be handled approximately by low-dimensional structures.'' The next constructive task is to define, as quantitative indices, the reconstruction error, the variance-preservation ratio, and the information loss obtained when the operating-profile space is projected to a lower dimension, and to formalize approximability through metric-space or information-theoretic measures. By this means, the validity of the coarse-graining regions as an explanatory map becomes an object of quantitative verification.

\textbf{Empirical verification of the verifiability hypotheses H1--H5.} This paper has presented five verifiable hypotheses: H1 (inter-subject variation of candidate sufficient statistics), H2 (receptivity prediction from the profile state), H3 (effect of state-representation alignment), H4 (resolution bias and error-absorption strategy), and H5 (constructive process of meaning, value, and justice). The next constructive task is to develop an empirical research program that verifies these hypotheses on real data, quantifies effect sizes, and confirms reproducibility. Each hypothesis can be verified independently, and a graded verification from H1 to H5 provides a cumulative confirmation of the predictive power of the present framework.

\textbf{Construction of falsification conditions.} The structural conditions under which the present framework could be falsified---insufficiency of the number of coarse-graining regions, continuity across phases, non-existence of the alignment map $\Phi$, and so on---each call for the construction of explicit falsifiable hypotheses and tests; this construction is part of the process by which the present framework matures as a research program. The Popperian refinement of falsifiability is the next step toward strengthening the present framework as a scientific theory.

\textbf{Quantitative verification of commensurability with existing research on epistemic diversity.} The empirical data accumulated under such labels as the ``New Look'' psychology of Bruner and Postman~\cite{bruner1947}, Witkin's cognitive styles~\cite{witkin1962}, Anderson and Pichert's schema theory~\cite{anderson1978}, Nisbett's cultural psychology~\cite{nisbett2003}, the Big Five~\cite{costa1992,goldberg1990}, and MBTI and Socionics~\cite{myers1985,augustinaviciute1980,pietrak2018}---all of which document the phenomenon that ``different estimation targets arise from the same stimulus''---constitute the next constructive task: redescribing these data under the continuous formal apparatus of the present paper (the operating fields $r,e,s$ on the candidate sufficient statistic space $\mathcal C$, the phase-formation space $\Zphase$, and the three layers and three typologies of $\Phi$). A quantitative comparison of which of the continuous formal apparatus of this paper and the discrete sixteen-type-style approximation has greater explanatory power, in which observational domains and to what degree, is the principal subsequent research that positions the present paper as ``a formal apparatus that integrates existing research on epistemic diversity.''

The above are the research objects opened by the framework of this paper, each of which can be developed as an independent constructive and empirical research program. The present paper extends only to the formulation of the hypotheses and to the preparation of their verifiability; the constructive developments will be carried forward in collaboration with subsequent research.

Low-dimensional approximability can, in subsequent research, be handled as a reconstruction error. Letting $\Theta^\alpha$ denote the high-dimensional operating structure, $\Pi_d$ a low-dimensional projection, and $\hat{\Theta}^\alpha(\Pi_d(\Theta^\alpha))$ the reconstruction therefrom, the quantity
\begin{equation}
\epsilon_d
=
\mathbb{E}_{\alpha}
\left[
D\left(
\Theta^\alpha,
\hat{\Theta}^{\alpha}(\Pi_d(\Theta^\alpha))
\right)
\right]
\end{equation}
can be used to evaluate how much information the low-dimensional approximation preserves as an explanatory map. This quantification is left as a limitation of the present paper and as a task for future empirical and technical research.

\subsection{Requirements for a complete generative-model implementation}

The equations of this paper do not constitute an immediately executable estimation algorithm. A complete generative-model implementation would require at least the following components. First, a distribution $p(c\mid o_{1:t},W_t,\Omega_t)$ that generates candidate sufficient statistics $c=(Y,\psi,\rho,T)$ from the observation sequence and context. Second, the estimation, for each candidate, of the prediction error $L(c)$, the error cost $c_{\mathrm{err}}(c)$, and the trigger threshold $\eta(c)$. Third, the dynamics of the foregrounding field $R_t^\alpha$ that determines to which candidates within the candidate space processing is pushed. Fourth, a function that evaluates plan candidates $P_i$ under prediction-error reduction, value-related state representations, computational constraints, observation constraints, and action constraints. Fifth, a learning rule that updates $W$, $R$, and $\Omega$ through execution-outcome feedback $\Delta_t^\alpha$. The update of directional maturity $\sigma$ is treated as the update of an internal component of $\Omega=(\theta,\lambda,q,\zeta,\sigma)$.

In an implementation based on LLM persona simulation, these components may be approximated as a configuration of prompts, latent representations, dialogue history, evaluators, meta-cognitive explanation generation, and post-dialogue feedback. For example, one can construct a dialogue-support system that, in response to the same observation, generates multiple candidate estimation targets, makes explicit the state representation and prediction error of each, and transforms them into representations that are processable by other subjects. Appendix~B.4 presents a minimal empirical protocol for advancing in this direction.

\section{Research Program}

The preceding chapters propose a common apparatus for several fields that usually work apart from one another. The core object is not a doctrine to be accepted as a whole, but a set of research entry points: how observations become inferentially relevant, how error-selection structures diverge, how state representations become processable across subjects, and how AI systems may support rather than collapse this diversity. This section organizes those entry points by field. The aim is to make the framework usable: each subsection indicates a line of inquiry that can be developed independently while remaining connected to the same formal vocabulary of estimation targets, state representations, prediction errors, error costs, operating profiles, and alignment maps.

\subsection{Questions for epistemology --- constructive description of incommensurability and perspective}

Can Kuhn's \cite{kuhn1962} incommensurability be quantified? In the framework of this paper, different subjects constitute the estimation target $Y$, the state representation $T$, and the prediction error $L$ in different ways, and the divergence of conclusions is not a logical failure but a structural consequence of operating-profile $\theta^\alpha$ differences. By describing incommensurability as the difficulty of the alignment map $\Phi$, one can ask at which layer between two paradigms which pattern of action --- full action, partial action, or rupture --- holds.

Can perspectivism, in the tradition descending from Nietzsche, be given a computable description? If a route opens toward describing the perspective itself as a tuple $(\rho, \tau, \Omega)$ in constructive terms, then perspectivism gains room to develop from a metaphysical thesis into a research field with observable objects.

Can the long opposition between correspondence and coherence theories of truth be reformulated in a different vocabulary? The contact with the world that correspondence theories have sought to preserve, and the internal structure of belief systems that coherence theories have sought to preserve, can be handled as questions at different levels in the framework of this paper: as the prediction error $L_K$ (contact with the world in phase $K$) and as the systematic coherence of the state representation $T_K$ (the internal structure of phase $K$).

\subsection{Questions for ethics and social philosophy --- constructive positioning of diversity}

How does meaning and value become an object of ethics? In the framework of this paper, ethics is treated not as mere inner belief or subjective preference but as a constructive problem concerning which state representations, prediction errors, and error costs should be preserved and protected. Why is cognitive diversity an ethical value? The framework of this paper provides a constructive ground for this question. The generalization performance of the social world model is sustained by a population of subjects with diverse foregrounding tendencies across coarse-grained domains (Section 10). If all subjects held the same $\theta$, the range of detectable prediction errors would be narrower. Diversity is at once an ethical value and a functional condition of the social epistemic system. How can ethics handle this dual positioning?

What is preserved by respecting the other? The preservation of the other indicated by Levinas's \cite{levinas1961} irreplaceability and Buber's \cite{buber1923} I--Thou relation can be reformulated in the framework of this paper as an operation that, without erasing the singularity of the other's operating profile $\theta^\beta$ and representation-formation and processing profile $\zeta^\beta$, constitutes only processability through $\Phi$. How can one refine, within the formal apparatus, an ethics of non-reductive coexistence rather than an ethics of identification?

\subsection{Questions for political philosophy --- structural description of democratic disagreement}

Can democratic disagreement be described not as a failure of consensus formation but in another way? Disagreement is often processed as ``lack of education,'' ``bad faith,'' or ``insincerity.'' In the framework of this paper, disagreement is described as a structural process in which a population of subjects with different operating profiles $\theta^\alpha$ raise different estimation targets in response to the same policy, event, or statistic. Many political conflicts arise because, while sharing the vocabulary of meaning, value, and justice, the parties do not share the state representations and error costs to which those terms are connected.

Processability while preserving difference --- can this become the cognitive basis of democratic politics? The stance that treats disagreement not as a convergence process toward consensus but as a design problem of state-representation alignment $\Phi$ opens, between deliberative democracy and pluralism, a concrete point of contact not available before. Can political philosophy take up the institutional implementation of world-model alignment as one of its central problems?

\subsection{Questions for social-science methodology --- reconsidering the relation between the natural and social sciences}

Should the boundary between the natural and social sciences be drawn by the presence or absence of observer dependence? Under the framework of this paper, natural-scientific observation tends to foreground empirical and structural coarse-grained domains strongly, while social-scientific observation tends to foreground existential and ideal coarse-grained domains strongly. Both, however, pass through subject-level estimation-target selection and state-representation extraction. The difference between them is not the presence or absence of observer dependence but a quantitative difference in which domain is foregrounded to what degree.

Can the methodological tension between qualitative and quantitative research be positioned in another way? What qualitative research has sought to preserve --- the lived perspective, interpretation, and relationship --- can be regarded as state representations of the \emph{existential domain}, and what quantitative research has sought to preserve --- reproducibility, operability, and structural relation --- as state representations of the \emph{structural domain}. The two have independent significance. There is no need to cast the relation as a competition over ``which is scientific.''

\subsection{Questions for AI ethics and design --- forms of AI that support diversity}

Will AI be a device that erases human diversity, or one that supports it? Current recommender systems, dialogue agents, and educational and policy-support AI place the maximization of users' immediate preferences at the center of their design. The long-term consequences of this design, however, are cognitive homogenization, echo chambers, and deepening fragmentation. The framework of this paper indicates another design possibility. AI can be designed not as a device that compresses meaning, value, and justice into single labels for judgment, but as a mediating device that makes visible how these are constituted within which world models and renders them processable across subjects. Output design that takes the operating-profile distribution of the user population into account, the provision of state-representation alignment through $\Phi$, and design that prompts meta-cognition in users are all possible.

Can value alignment be understood as a broader problem than what RLHF, DPO, and Constitutional AI have addressed? Under the \emph{three-layer $\Phi$ hierarchy (value, state representation, and world model)}, current alignment research is repositioned as the most restricted of the three layers. Constructing state-representation alignment and world-model alignment as implementable technologies can become the central task of next-generation AI design.

Can an AI that implements multi-phase inference internally recognize that it itself holds different inference phases and disclose this to the user? An AI that can declare ``I am strongly foregrounding the structural domain'' or ``this answer has thin empirical grounding'' could qualitatively change world-model alignment between humans and AI.

Whose strategy do the maximization of immediate preferences and action-driven learning normalize? Many current AI learning and evaluation metrics are organized around immediate response rate, engagement, and immediate collection of preference signals, and they tacitly incorporate as a norm the strategy of absorbing prediction error through short-term action. By contrast, subjects who emphasize coarse-grained long-term consistency may take a strategy of not answering immediately, suspending, and responding over a long horizon via abstraction. Such subjects are structurally likely to be evaluated as ``slow to react'' or ``uncooperative'' under short-term metrics.

Hence AI must judge, in accordance with operating profiles that include $r_{\mathrm{res}}^\alpha$, whether to prompt the user to act immediately, to support suspension and reinterpretation, or to connect both in a particular order. This is at once a design problem of extending active inference to human reflection, meaning-making, and long-term state-representation reconstruction, and an ethical problem concerning the implicit image of the human built into AI design.

\subsection{Collective scale --- culture, tradition, and civilization}

MIM connects not only to individual cognitive differences but also to world-model formation at the collective scale. Culture, tradition, and civilization are sets of estimation targets, state representations, error costs, and update rules that have stabilized on long time scales. What a society treats as damage to dignity, what it treats as institutional inconsistency, and what it treats as a technical challenge depend on the operating profile of the world model that society has historically formed.

From this viewpoint, intercultural and intergenerational conflict can be described not as a mere difference of values but as a problem of state-representation alignment between collective world models. AI can become a support device that, rather than converging these differences into a single value system, makes explicit what different collective world models are trying to preserve and constructs processable transformations.

\subsection{Implementation paths and technical research}

Technically, MIM can be developed in at least three directions. First, agent simulation. Multiple subjects with different $\Omega$ are generated, and one observes which estimation targets, state representations, and prediction errors arise in response to identical observations. Second, representation-learning research. Using the internal representations of LLMs and multimodal models, one estimates differences in estimation targets and state representations in latent space. Third, dialogue-support systems. Utterances between subjects are analyzed; the layers at which processability is lost are made visible; and candidates for the alignment map $\Phi$ are generated.

These implementations do not merely apply the theory of this paper. In the course of implementation, it becomes clear which symbols are measurable, which symbols remain abstract, and which update rules require further specification. Hence technical research is at once the verification of MIM and the refinement of MIM itself.

\subsection{Relations to neighboring fields}

The framework touches neighboring domains without absorbing them. It does not replace Bayesian model comparison, active inference, reinforcement learning, theory of mind, social epistemology, political philosophy, or psychometrics. Rather, it supplies a shared representational layer on which these fields can ask a common question: which observations become inferentially relevant, which errors are detected, which errors are excluded, and what is lost when one state representation is transformed into another. This is the intended function of the parent paper. It does not close the research program; it provides a coordination surface for researchers who approach the same problem from different disciplines.

\section{Conclusion}

We attend the same meeting, read the same documents, and follow the same reports, yet we think different things, feel different things, and arrive at different conclusions. Why? The answer offered in this paper begins before disagreement. It begins where an observation sequence becomes relevant, where a subject forms an estimation target, where a state representation is stabilized, where one deviation becomes an error and another disappears without a name.

The Multi-Phase Inference Assumption (MIA) and the Multi-Phase Inference Mechanism (MIM) were introduced to describe this process without reducing it to ignorance, irrationality, bad faith, or a vague appeal to values. The phase-formation space $\Zphase$, the foregrounding gradient field $R$, and the alignment map $\Phi$ describe epistemic diversity as a structural property of finite intelligence: from partially shared observation sequences, different estimation targets, state representations, prediction errors, error costs, and update paths can arise.

The shortest form of the argument is this. Observation is not yet inference. A constrained subject does not merely receive an observation and then infer from it. A possible target must first become admissible, and this happens only when the subject can construct a state representation that is sufficiently informative for prediction, evaluation, or action with respect to that target. Correct inference is therefore indispensable: without such selection, no observation could become inferentially relevant. Yet correct inference also excludes errors. It makes some deviations visible and processable while leaving other possible errors outside the space of relevance. This exclusion is not an accidental defect added from outside cognition. It is the condition under which cognition operates at all.

This is also the point at which the theory becomes a theory of possibility rather than merely a theory of disagreement. If different subjects exclude different errors, then cognitive diversity is not simply a burden to be managed after reasoning has failed. It is a distributed capacity for sensing distortions, injuries, risks, opportunities, and losses that no single inference profile can detect alone. A society that preserves only one mode of correct inference may become locally consistent while becoming blind to the errors that fall outside its regime of relevance.

This is why the central problem of alignment cannot be agreement. If correct inference always selects some errors and excludes others, then no society can remain self-correcting by installing a single correct inferential regime. Alignment must instead mean processability: the capacity to preserve, compare, transform, and coordinate heterogeneous error-detection capacities without forcing them into one representation, one objective function, or one value system.

The aim of this paper has not been to replace philosophy with computation. It has been to give the questions surrounding recognition, meaning, value, conflict, and practice a constructive vocabulary that can be formalized, tested, and mediated by AI systems. The history of philosophy, cognitive typology, social fragmentation, culture and civilization, and AI alignment are distinct fields, and none of them is reducible to MIM. But the error sensitivities and state representations that they have preserved can be connected to a comparable and processable formal vocabulary.

From the perspective of distributed error detection presented in this paper, human society is a world-model update structure in which different inferential subjects detect different prediction errors and preserve them as culture, institutions, common sense, science, and politics. Diversity is therefore not merely a slogan of value pluralism. It is a structural condition under which society continues to pick up errors that no single inference profile can detect. Freedom, rights, dissent, privacy, minority protection, traditional knowledge, lay knowledge, professional autonomy, and institutional deliberation can all be reread as error-input mechanisms that prevent society from hardening into a single world model.

In the AI era, this structure becomes still more consequential. AI can detect prediction errors that humans cannot easily see, which creates a temptation toward governance by AI. But human society is not an object of AI-style optimization. The task of AI alignment is not exhausted by fitting AI to a single human value; it is the design problem of extending human society's intrinsic capacities for error detection, error selection, and error coordination while making transformation losses explicit and leaving responsibility for world-model update within human society.

This is why the multi-phasing of design itself is necessary. Institutional and AI-system design should not be closed under a single expert perspective, a single value function, or a single optimization objective. It should be opened so that subjects with different inference phases, state representations, and error costs can input different error signals into the design process itself. The goal is not to make disagreement pleasant or to pretend that every perspective is right. It is to keep open the errors that each perspective is able to detect before they are erased by the convenience of a single regime.

An AI that understands self and others is therefore not a device that knows the final human answer. It is a device that helps us see what each world model reveals, what it excludes, and what becomes possible when those differences remain alive enough to be processed. It does not erase the distance between worlds. It makes that distance thinkable, comparable, and livable.

\appendix
\section*{Appendix A: Notation Summary}
\phantomsection
\addcontentsline{toc}{section}{Appendix A: Notation Summary}

The main symbols and concepts used in this paper are organized in the order in which they are introduced. The symbol table is an index for referring to the formal system of the main text; coarse-grained domains and simplified notations are explained explicitly where they are needed.

\medskip
\noindent\textbf{World Models, Observations, and Outputs}
{\small
\begin{longtable}{L{0.18\linewidth} L{0.20\linewidth} L{0.52\linewidth}}
\toprule
Symbol & Type & Meaning and place of definition \\
\midrule
$O_t^{\alpha}$ & observation sequence & The observation sequence available to subject $\alpha$ up to time $t$. The observation sequence is not recognition in itself, but the input for forming state representations. Section 3.1  \\
$W_t^\alpha$ & broad world-model state & $W_t^\alpha=(M_t^\alpha,{\mathcal Z}_{\mathrm{phase},t}^\alpha,R_t^\alpha,\Omega_t^\alpha)$. The world-model state of subject $\alpha$, including the generative model in the narrow sense, the phase-formation space, the foregrounding gradient field, and the profile state. Section 5.3  \\
$M_t^\alpha$ & generative model in the narrow sense & The generative-model component that forms latent states, prediction errors, and output candidates from the observation sequence. It is one component of the broad world-model state $W_t^\alpha$. Section 5.3  \\
$y_t^\alpha$ & output & The output that subject $\alpha$ returns to the environment, others, or the self through plan candidates $P_t^\alpha$. It includes utterance, action, suspension, explanation, reinterpretation, and related forms. Section 5.3  \\
\bottomrule
\end{longtable}
}

\medskip
\noindent\textbf{Estimation Targets, State Representations, and Candidate Sufficient Statistics}
{\small
\begin{longtable}{L{0.18\linewidth} L{0.20\linewidth} L{0.52\linewidth}}
\toprule
Symbol & Type & Meaning and place of definition \\
\midrule
$\mathcal H_t^\alpha$ & class of candidate latent structures & The class of candidate latent variables and candidate latent relations that can be constructed as estimation targets from the observation sequence. Section 3.2  \\
$\mathcal T_t^\alpha$ & class of feasible internal states & The class of internal states that subject $\alpha$ can construct under finite computational, observational, and action constraints. Section 3.2  \\
$\mathcal Y_t^\alpha$ & estimation-target space & The space of latent variables and latent relations for which approximate sufficient statistics can be constructed. Section 3.2  \\
$Y_t^\alpha$ & estimation target & A latent variable or latent relation for which prediction error can be defined. It becomes an estimation target when an approximate sufficient statistic $T$  can be constructed. Section 3.2  \\
$T_t^\alpha$ & state representation & A state representation constructed from the observation sequence and serving as an approximate sufficient statistic for the estimation target $Y$. Section 3.1--3.4  \\
$\eta$ & natural parameter & A coordinate that weights components of sufficient statistics when finite-dimensional sufficient statistics are treated in the canonical form of an exponential family. Section 3.5  \\
$\mathcal A_t^\alpha$ & admissible estimation-target--state-representation pair & The space of $(Y,T)$  satisfying the approximate-sufficiency condition. Section 4.1  \\
$\psi$ & conditioning basis & The basis that specifies as which latent structure, feature set, or relational structure the observation sequence is articulated when a state representation is constructed. Section 4.1  \\
$\rho$ & resolution & The state-representation format that specifies at what granularity, temporal range, spatial range, abstraction level, or operational form the conditioning basis is compressed. Section 3.4, 4.2  \\
$\tau_\rho$ & state-representation formation map & A map through which an approximate sufficient statistic $T$ is constructed from the observation sequence under a conditioning basis $\psi$ and a resolution $\rho$. Section 4.2  \\
$\mathcal C_t^\alpha$ & candidate sufficient-statistic space & $\mathcal C_t^\alpha=\{(Y,\psi,\rho,T)\}$. The space of candidates that simultaneously specify an estimation target, conditioning basis, resolution, and state representation. Section 4.2  \\
$c$ & candidate sufficient statistic & $c=(Y,\psi,\rho,T)$. An element of the candidate sufficient-statistic space. Section 4.2  \\
$L_t^\alpha(c)$ & candidate-specific prediction error & The prediction error defined on the basis of the estimation target and state representation specified by candidate $c$. Sections 5.2 and 12.6  \\
$c_{\mathrm{err},t}^\alpha(c)$ & candidate-specific error cost & A weight expressing how important the prediction error of candidate $c$ is for the subject. Section 13.6  \\
$I_t^\alpha(c)$ & meaningful error intensity & $I_t^\alpha(c)=c_{\mathrm{err},t}^\alpha(c)L_t^\alpha(c)$. A candidate-specific prediction error weighted by error cost. Section 13.6  \\
$\eta_t^\alpha(c)$ & threshold of reflective awareness & The threshold by which the weighted error of candidate $c$ is sent to reflective processing, verbalization, or action. It belongs to the activation/awareness-threshold component $q$. Section 13.6  \\
\bottomrule
\end{longtable}
}

\medskip
\noindent\textbf{Phase-Formation Space and Coarse-Grained Vocabulary}
{\small
\begin{longtable}{L{0.18\linewidth} L{0.20\linewidth} L{0.52\linewidth}}
\toprule
Symbol & Type & Meaning and place of definition \\
\midrule
${\mathcal Z}_{\mathrm{phase},t}^\alpha$ & phase-formation space & The space obtained from candidate sufficient statistics $c=(Y,\psi,\rho,T)$ by forgetting the state representation $T$ and extracting the constructive degrees of freedom $(Y,\psi,\rho)$ of the estimation target, conditioning basis, and resolution. Section 4.2  \\
$\mathcal Z_{\mathrm{ref}}$ & conditioning-basis space & The space in which the conditioning basis $\psi$ is formed as a latent structure, feature set, or relational basis for constructing state representations. Section 4.1  \\
$\mathcal Z_{\mathrm{res}}(\psi)$ & resolution space & The space of resolutions $\rho$ that satisfy the approximate-sufficiency condition for a given conditioning basis $\psi$. Section 4.2  \\
$\mathcal K_{\mathrm{coarse}}$ & set of coarse-grained domains & A set of coarse-grained domains consisting of empirical, ideal, structural, and existential explanatory vocabularies. It is not an axiom of MIM, but an approximate map for explaining representative stable regions on $\mathcal Z_{\mathrm{phase}}$. Section 7  \\
empirical domain & explanatory vocabulary & A coarse-grained domain in which estimation targets concerning embodiment, nearby feedback, and action possibility are foregrounded. Section 7.1  \\
ideal domain & explanatory vocabulary & A coarse-grained domain in which estimation targets concerning long-horizon representation, symbols, future images, and possible worlds are foregrounded. Section 7.1  \\
structural domain & explanatory vocabulary & A coarse-grained domain in which estimation targets concerning form, constraint, and reproducible order are foregrounded. Section 7.1  \\
existential domain & explanatory vocabulary & A coarse-grained domain in which self and other, trust, responsibility, and relational prediction errors are foregrounded. Section 7.1  \\
\bottomrule
\end{longtable}
}

\medskip
\noindent\textbf{Operating Profile and Profile State}
{\small
\begin{longtable}{L{0.18\linewidth} L{0.20\linewidth} L{0.52\linewidth}}
\toprule
Symbol & Type & Meaning and place of definition \\
\midrule
$\theta_t^\alpha$ & operating profile & $(r_t^\alpha,e_t^\alpha,s_t^\alpha)$. The foregrounding field, exploration field, and stabilization field on the candidate sufficient-statistic space. Section 5.1  \\
$r_t^\alpha$ & foregrounding field & $r_t^\alpha:\mathcal C_t^\alpha\to\mathbb R$. It expresses which candidates are likely to rise into processing on the candidate sufficient-statistic space. Section 5.1  \\
$e_t^\alpha$ & exploration field & $e_t^\alpha:\mathcal C_t^\alpha\to\mathbb R_{\ge0}$. It expresses the degree to which alternative candidates are likely to be explored and expanded. Section 5.1  \\
$s_t^\alpha$ & stabilization field & $s_t^\alpha:\mathcal C_t^\alpha\to\mathbb R_{\ge0}$. It expresses the degree to which the current candidate is likely to be maintained or fixed. Section 5.1  \\
$\tilde\theta_t^\alpha$ & posterior operating state & $(\tilde r_t^\alpha,\tilde e_t^\alpha,\tilde s_t^\alpha)$. The state of foregrounding, exploration, and stabilization that actually fires after observation processing. Section 5.2  \\
$\Omega_t^\alpha$ & profile state & $(\theta_t^\alpha,\lambda_t^\alpha,q_t^\alpha,\zeta_t^\alpha,\sigma_t^\alpha)$. A subject-specific state that brings together the operating profile, plasticity/fixation, firing/awareness thresholds, representation-formation and processing profile, and direction-specific maturity. Section 5.1  \\
$\lambda_t^\alpha$ & plasticity/fixation parameter & It expresses how much the operating field is updated by experience or fixed. Section 5.1  \\
$q_t^\alpha$ & activation parameter & A parameter grouping candidate-specific error costs, thresholds of reflective awareness, and thresholds for action. Section 5.1, 12.6  \\
$\zeta_t^\alpha$ & representation-formation and processing profile & $({\chiop}_t^\alpha,\tau_t^\alpha,\kappa_t^\alpha)$. Constraints on estimation-target formation control, state-representation formation, and compressibility / externalizability. Section 5.1  \\
${\chiop}_t^\alpha$ & estimation-target formation control & A control expressing which candidate latent variables or candidate latent relations are likely to be constructed as candidate estimation targets. Section 5.1  \\
$\tau_t^\alpha$ & state-representation formation control & A family of maps that construct approximate sufficient statistics on the basis of the conditioning basis and resolution. Section 5.1  \\
$\kappa_t^\alpha$ & compressibility / externalizability & It expresses how far a state representation can be compressed, symbolized, verbalized, or transformed into a sharable form. Section 5.1  \\
$\sigma_t^\alpha$ & direction-specific maturity & Maturity of state-representation formation, plan completion, and error processing in repeatedly operated candidate families or directional components. Section 5.1, 12.7  \\
$r_{\mathrm{ref}}^\alpha$ & conditioning-basis foregrounding component & The component of the foregrounding field $r^\alpha$ that expresses foregrounding tendencies in the conditioning-basis space $\mathcal Z_{\mathrm{ref}}$. Section 5.1  \\
$r_{\mathrm{res}}^\alpha$ & resolution foregrounding component & The component of the foregrounding field $r^\alpha$ that expresses foregrounding tendencies of granularity, horizon, and abstraction in the resolution space $\mathcal Z_{\mathrm{res}}(\psi)$. Sections 4.2 and 5.1  \\
\bottomrule
\end{longtable}
}

\medskip
\noindent\textbf{Generative Modeling, Action Constraints, and Feedback}
{\small
\begin{longtable}{L{0.18\linewidth} L{0.20\linewidth} L{0.52\linewidth}}
\toprule
Symbol & Type & Meaning and place of definition \\
\midrule
$x(c)$ & inference-phase coordinate & $x(c)=(Y,\psi,\rho)$, obtained from a candidate sufficient statistic $c=(Y,\psi,\rho,T)$ by removing the state representation $T$. Section 13.6  \\
$R_t^\alpha$ & foregrounding gradient field & $R_t^\alpha(x(c),L_t^\alpha(c);\theta_t^\alpha,\zeta_t^\alpha)$. A direction-field-like structure that returns the direction in which processing is likely to unfold according to the inference-phase coordinate and prediction error. Sections 5.1 and 12.6  \\
$v_t^\alpha(c)$ & foregrounding vector & The instantaneous candidate-specific processing direction obtained from $R_t^\alpha(x(c),L_t^\alpha(c);\theta_t^\alpha,\zeta_t^\alpha)$. Section 13.6  \\
$P_{i,t}^\alpha$ & plan candidate & Candidates such as thought, suspension, reinterpretation, exploration, explanation, utterance, bodily action, and coordination generated from foregrounded candidate sufficient statistics. Section 13.7  \\
$\Pi_t^\alpha(P_i)$ & plan priority & The priority of a plan candidate obtained from expected prediction-error reduction, value-relevant evaluation, computational constraints, observational constraints, action constraints, and consistency with the foregrounding gradient field. Section 13.7  \\
$Q_t^\alpha$ & working-memory queue & A short-term queue in which multiple thoughts, actions, suspensions, and coordination plans are held with priorities. Section 13.7  \\
$C_{\mathrm{act}}$ & action-execution constraint & A higher-level constraint expressing the feasibility not only of external bodily action, but also of internal and social acts such as thought, suspension, reinterpretation, explanation, and coordination. Section 13.7  \\
$C_{\mathrm{body}}$ & embodiment constraint & Constraints concerning bodily capacity, fatigue, sensorimotor ability, coupling with the environment, and bodily load. Section 13.7  \\
$C_{\mathrm{time}}$ & time constraint & Constraints concerning available time windows, deadlines, short- or long-term horizons, and whether resources are allocated to reflection or action. Section 13.7  \\
$C_{\mathrm{skill}}$ & skill/maturity constraint & Constraints concerning the maturity of plan completion, problem decomposition, proceduralization, and handling of external conditions. Section 13.7  \\
$C_{\mathrm{coop}}$ & coordination constraint & A lower-level constraint operating when approval, participation, understanding, trust, or consensus formation by others is required. Section 13.7  \\
$C_{\mathrm{comm}}$ & communication constraint & Constraints concerning the feasibility of verbalization, explanation, externalization, negotiation, and state-representation alignment. Section 13.7  \\
$\Delta_t^\alpha$ & execution-result feedback & Changes in prediction error, value-relevant evaluation, action constraints, coordination, and alignment maps obtained as results of plan execution, reflection, suspension, coordination, or utterance. Section 13.7  \\
\bottomrule
\end{longtable}
}

\medskip
\noindent\textbf{State-Representation Alignment and the Three-Layer Hierarchy}
{\small
\begin{longtable}{L{0.18\linewidth} L{0.20\linewidth} L{0.52\linewidth}}
\toprule
Symbol & Type & Meaning and place of definition \\
\midrule
$\Phi_{\alpha\to\beta}^{K\to K'}$ & alignment map & A map that transforms the sender's state representation in coarse-grained domain $K$ into an approximate sufficient statistic processable by receiver $\beta$ in coarse-grained domain $K'$, represented at the level of candidate sufficient statistics as $c\to c'$. Section 9.1  \\
$\tilde T_{K',t}^{\beta}$ & aligned state representation & A state representation transformed by $\Phi$ into a form processable by receiver $\beta$. Section 9.1  \\
value alignment & three-layer hierarchy: restricted domain & A relatively narrow operation that aligns human evaluation signals with AI behavior, including RLHF, DPO, Constitutional AI, and related methods. Section 9.2  \\
state-representation alignment & three-layer hierarchy: middle layer & A formal operation described by $\Phi$, which transforms state representations into forms processable by another subject. Section 9.2  \\
world-model alignment & three-layer hierarchy: upper layer & The broader goal of constructing processability among heterogeneous world models through collections and chains of $\Phi$ maps. Section 9.2  \\
\bottomrule
\end{longtable}
}

\medskip
\noindent\textbf{Genealogical Correspondence of Major Concepts}
{\footnotesize
\setlength{\tabcolsep}{3pt}
\begin{longtable}{@{}L{0.24\linewidth} L{0.30\linewidth} L{0.40\linewidth}@{}}
\toprule
related simplified notation & formalization in this paper & meaning \\
\midrule
$\theta=(R,E,S,D)$ & $\theta=(r,e,s)$, $\lambda$, $q$, $\mathcal Z_{\mathrm{phase}}$ & Operating tendencies are represented as the foregrounding field $r$ , the exploration field $e$ , and the stabilization field $s$  on the candidate sufficient-statistic space. Plasticity/fixation is handled by $\lambda$ , firing/awareness thresholds by $q$ , and phase-formation conditions by $\mathcal Z_{\mathrm{phase}}$. \\
$R$ (reference) & $r$ and conditioning basis $\psi$ & Reference weights are treated as the foregrounding field $r$ over candidate sufficient statistics $c=(Y,\psi,\rho,T)$ and as the formation of the conditioning basis $\psi$. \\
$E$ (exploration) & $e: \mathcal C_t^\alpha\to\mathbb R_{\ge0}$ & An operating field that explores alternative estimation targets, conditioning bases, resolutions, and state representations. \\
$S$ (stabilization) & $s: \mathcal C_t^\alpha\to\mathbb R_{\ge0}$ & An operating field that maintains and fixes the current candidate sufficient-statistic construction. \\
$D$ (horizon) & $\rho$, $d$, $R(D)$ & Temporal horizon, spatial horizon, and abstraction level are treated as part of the resolution $\rho$  , the distortion function $d$  , and the rate-distortion constraint $R(D)$. \\
SIA & Single Intelligence Assumption & It treats the assumption that the same observations should yield the same conclusion as a composite assumption of centralization, normativization, and commutability. \\
MIA & Multi-Phase Inference Assumption & The assumption that different constructions of approximate sufficient statistics can arise from the same observations; it is the premise of MIM. \\
MIM & Multi-Phase Inference Mechanism & It formalizes the internal mechanism required by MIA as a tuple of the candidate sufficient-statistic space, operating profile, and alignment map. \\
$\theta$level non-identifiability & diversity of operating profiles & The fact that which candidate sufficient statistics are foregrounded, explored, and stabilized for the same observation sequence differs across subjects. \\
$W$level non-identifiability & divergence of world-model states & The divergence of world-model states $W$  into different stable structures across subjects through repeated sufficient-statistic formation and feedback updating. \\
abstract/concrete, externalizability, and order/freedom & $\mathcal Z_{\mathrm{phase}}$, $\psi$, $\rho$, $r,e,s$, $\mu$ & Abstract/concrete is decomposed into the resolution space $\mathcal Z_{\mathrm{res}}(\psi)$  ; externalizability into the formation of conditioning basis $\psi$  ; and order/freedom into the exploration field $e$  and the stabilization field $s$  as well as directional fitness $\mu$. \\
\bottomrule
\end{longtable}
}

\medskip
\noindent\textbf{contact point}
\begin{flushright}
Email: \texttt{toru@mis.doshisha.ac.jp}
\end{flushright}

\section*{Appendix B: Supplementary Notes on the Formal Status of MIM}
\phantomsection
\addcontentsline{toc}{section}{Appendix B: Supplementary Notes on the Formal Status of MIM}

\subsection*{B.1 Descriptive formulation and executable implementation}
\phantomsection
\addcontentsline{toc}{subsection}{B.1 Descriptive formulation and executable implementation}

The equations of this paper do not directly yield an executable estimation algorithm. They are, rather, a descriptive formulation that fixes the positional relations among estimation targets, state representations, prediction errors, foregrounding, plan candidates, and feedback update. This does not mean abandoning implementation. It is the prior step that makes explicit which components are to be implemented.

Hence MIM is not ``an already completed computational model'' but ``a formal apparatus that makes explicit the constitutive components of an implementable research program.'' Candidate-generation distributions, plan-evaluation functions, error-cost estimation, and feedback-update rules are objects to be made concrete in future technical research.

\subsection*{B.2 Processability and agreement are not the same}
\phantomsection
\addcontentsline{toc}{subsection}{B.2 Processability and agreement are not the same}

Understanding in MIM is neither the sharing of an identical representation nor agreement on an identical conclusion. To understand is for a state representation formed in one subject to be reconstructed, without breakdown, in another subject's world model and, where necessary, in a form that can enter exploratory update.

Hence two subjects can make each other's state representations processable without agreeing. Conversely, two subjects who appear to agree on a word may, if each connects it to different estimation targets or prediction errors, fail to achieve processability. The alignment map $\Phi$ is not a map that guarantees agreement; it is a local transformation that constitutes processability.

\subsection*{B.3 Why retain the coarse-grained domains}
\phantomsection
\addcontentsline{toc}{subsection}{B.3 Why retain the coarse-grained domains}

The four coarse-grained domains --- empirical, ideal, structural, and existential --- are not axioms of MIM. They are an intermediate map for explaining the high-dimensional inference-phase-formation space $\Zphase$ and for connecting the framework to the history of philosophy and typology.

If this vocabulary were completely removed, the theory would become formally pure but would lose its connection to human experience and existing intellectual traditions. If, instead, these domains were fixed as basic phases, MIM would degenerate into a four-class theory. Hence the present paper retains them not as axioms but as empirically, philosophically, and explanatorily useful coarse-grained domains.

\subsection*{B.4 Minimal empirical protocol}
\phantomsection
\addcontentsline{toc}{subsection}{B.4 Minimal empirical protocol}

A minimal empirical test of MIM can be set up as follows. First, the same observation --- for example, the same meeting record, news article, social-media post, policy statement, or fragment of dialogue --- is presented to multiple subjects. Next, each subject is asked to describe what estimation target is raised, which state representation is preserved, which prediction error is evaluated as important, and which plan candidates are generated.

Then one measures the extent to which another subject can reconstruct these state representations, whether AI support improves reconstructability, and at which layers processability is lost. This protocol does not test MIM as a whole at once, but it is a minimal route for empirically investigating the central hypothesis that different estimation targets, state representations, and prediction errors arise from the same observation.

As an implementation, a three-stage structure is reasonable as a research program. The first stage is LLM persona simulation. Multiple LLM agents, each encoding a different $\Omega=(\theta,\lambda,q,\zeta,\sigma)$ via system prompts, are presented with the same observation and are asked to output estimation targets, state representations, prediction errors, and plan candidates. This functions as a minimal working demonstration of the formal apparatus of MIM and allows fast iteration. Its limitations are also clear: because LLMs do not directly hold human bodily or existential prediction errors, error detection in the empirical and existential domains remains an approximation. The second stage is human-subject experimentation. The predictions obtained at the LLM stage --- which observations cause which subjects to raise different estimation targets, state representations, and prediction errors --- are verified through free description and structured interviews with human subjects of different occupations, expertise, and cultural backgrounds. Existential and relational prediction errors that LLMs fail to pick up can be detected in the collected data, making the limitations of LLMs themselves an object of research. The third stage is AI-supported evaluation of the alignment map $\Phi$. The AI is made to translate among the state representations of different subjects, and the original subjects evaluate whether ``my own prediction error has been reconstructed in a processable form.'' This constitutes a concrete evaluation pipeline for transformation-loss-displaying AI and multi-error-selection-proposing AI, as discussed in Section 11. The three stages are not sequential but constitute a research program that proceeds by mutual iteration.

\subsection*{B.5 Relation to future technical research}
\phantomsection
\addcontentsline{toc}{subsection}{B.5 Relation to future technical research}

Future technical research can proceed in at least three directions. First, simulation studies using agent populations with different profile states. Second, estimation of state-representation differences using the internal representations of LLMs and multimodal models. Third, the construction of AI-support systems that visualize estimation targets, state representations, prediction errors, and processability against actual dialogue or meeting records.

These studies do not directly complete MIM. They do, however, clarify which symbols are measurable, which transformations are implementable, and which limitations remain. In this sense, MIM is not a completed algorithm but a research program for constituting AI understanding of self and other.

\subsection*{B.6 Methodology of philosophical-history stress test}
\phantomsection
\addcontentsline{toc}{subsection}{B.6 Methodology of philosophical-history stress test}

To read the history of philosophy in the vocabulary of MIM is not to reconstruct each philosopher's system exhaustively. It is, rather, to make explicit the estimation targets, state representations, impermissible prediction errors, coarse-grained objects, and transformation losses that each tradition has preserved. This method is not a replacement for the history of philosophy; it is a reading procedure for extracting the error structures required by an MIA-based theory of AI alignment.

\subsection*{B.7 The relation between humanities/social-science verification and AI implementation}
\phantomsection
\addcontentsline{toc}{subsection}{B.7 The relation between humanities/social-science verification and AI implementation}

Philosophy, social science, and cultural theory are not mere application areas of MIM. They are high-load verification domains showing which errors human society has preserved, which it has rendered invisible, and which transformation losses it has tolerated. AI implementation should not bypass this humanities/social-science verification. Rather, which errors an implemented AI system makes visible and which it renders invisible must be evaluated in connection with humanities/social-science knowledge.

\subsection*{B.8 Limits of AI-assisted error selection}
\phantomsection
\addcontentsline{toc}{subsection}{B.8 Limits of AI-assisted error selection}

AI can detect errors, translate them, present multiple selection proposals, and display transformation losses. If, however, AI ultimately monopolizes which errors are picked up and which are rejected, that strips human society of its intrinsic capacity to update its own world models. Hence the role of AI is not to be the sovereign of error selection but to act as a mediating device that increases the resolution of institutional, political, and cultural human judgment.

\subsection*{B.9 Multi-phasing of design itself}
\phantomsection
\addcontentsline{toc}{subsection}{B.9 Multi-phasing of design itself}

To multi-phase design itself is to open not only the design object but also the design process itself to diverse error detection. Engineers, jurists, frontline practitioners, those directly affected, local communities, traditional knowledge, lay knowledge, and AI each detect different errors. The task of design is not to compress these into a single evaluation function but to render them processable while making transformation losses explicit, so that human society can take responsibility for which errors are picked up.

\medskip
\noindent The aim is not to eliminate inferential diversity, but to build forms of intelligence, governance, and alignment that can preserve and coordinate the errors that each correct inference would otherwise exclude.

\end{document}